\documentclass[10pt,twocolumn,letterpaper]{article}

\usepackage[pagenumbers]{cvpr} 

\usepackage{graphicx}
\usepackage{amsmath}
\usepackage{amssymb}
\usepackage{booktabs}
\usepackage{xcolor}
\usepackage{tabularray}
\usepackage{caption,subcaption}

\usepackage[pagebackref,breaklinks,colorlinks]{hyperref}

\usepackage[capitalize]{cleveref}
\crefname{section}{Sec.}{Secs.}
\Crefname{section}{Section}{Sections}
\Crefname{table}{Table}{Tables}
\crefname{table}{Tab.}{Tabs.}

\def\cvprPaperID{7870} 
\def\confName{CVPR}
\def\confYear{2023}

\begin{document}

\title{An Image Quality Assessment Dataset for Portraits}

\author{Nicolas Chahine$^{1,2}$\hspace{10px}Ana-Stefania Calarasanu$^{1}$\hspace{10px}Davide Garcia-Civiero$^{1}$ \\ Théo Cayla$^{1}$\hspace{10px}Sira Ferradans$^{1}$\hspace{10px}Jean Ponce$^{2,3}$ \\
\normalsize{$^{1}$DXOMARK\hspace{10px}$^{2}$Département d'informatique de l'Ecole normale supérieure (ENS-PSL, CNRS, Inria)}\\
\normalsize{$^{3}$Institute of Mathematical Sciences and Center for Data Science, New York University}
}

\maketitle

\begin{abstract}
Year after year, the demand for ever-better smartphone photos continues to grow, in particular in the domain of portrait photography. Manufacturers thus use perceptual quality criteria throughout the development of smartphone cameras. This costly procedure can be partially replaced by automated learning-based methods for image quality assessment (IQA). Due to its subjective nature, it is necessary to estimate and guarantee the consistency of the IQA process, a characteristic lacking in the mean opinion scores (MOS) widely used for crowdsourcing IQA. In addition, existing blind IQA (BIQA) datasets pay little attention to the difficulty of cross-content assessment, which may degrade the quality of annotations. This paper introduces PIQ23, a portrait-specific IQA dataset of 5116 images of 50 predefined scenarios acquired by 100 smartphones, covering a high variety of brands, models, and use cases. The dataset includes individuals of various genders and ethnicities who have given explicit and informed consent for their photographs to be used in public research. It is annotated by pairwise comparisons (PWC) collected from over 30 image quality experts for three image attributes: face detail preservation, face target exposure, and overall image quality. An in-depth statistical analysis of these annotations allows us to evaluate their consistency over PIQ23. Finally, we show through an extensive comparison with existing baselines that semantic information (image context) can be used to improve IQA predictions. The dataset along with the proposed statistical analysis and BIQA algorithms are available: \url{https://github.com/DXOMARK-Research/PIQ2023} 
\end{abstract}

\SetTblrInner{rowsep=0pt, colsep=2pt}
\begin{figure*}[!h]
\centering

\begin{tblr}{ccccc@{\hskip 20pt}cc}
    \includegraphics[width = 0.11 \textwidth]{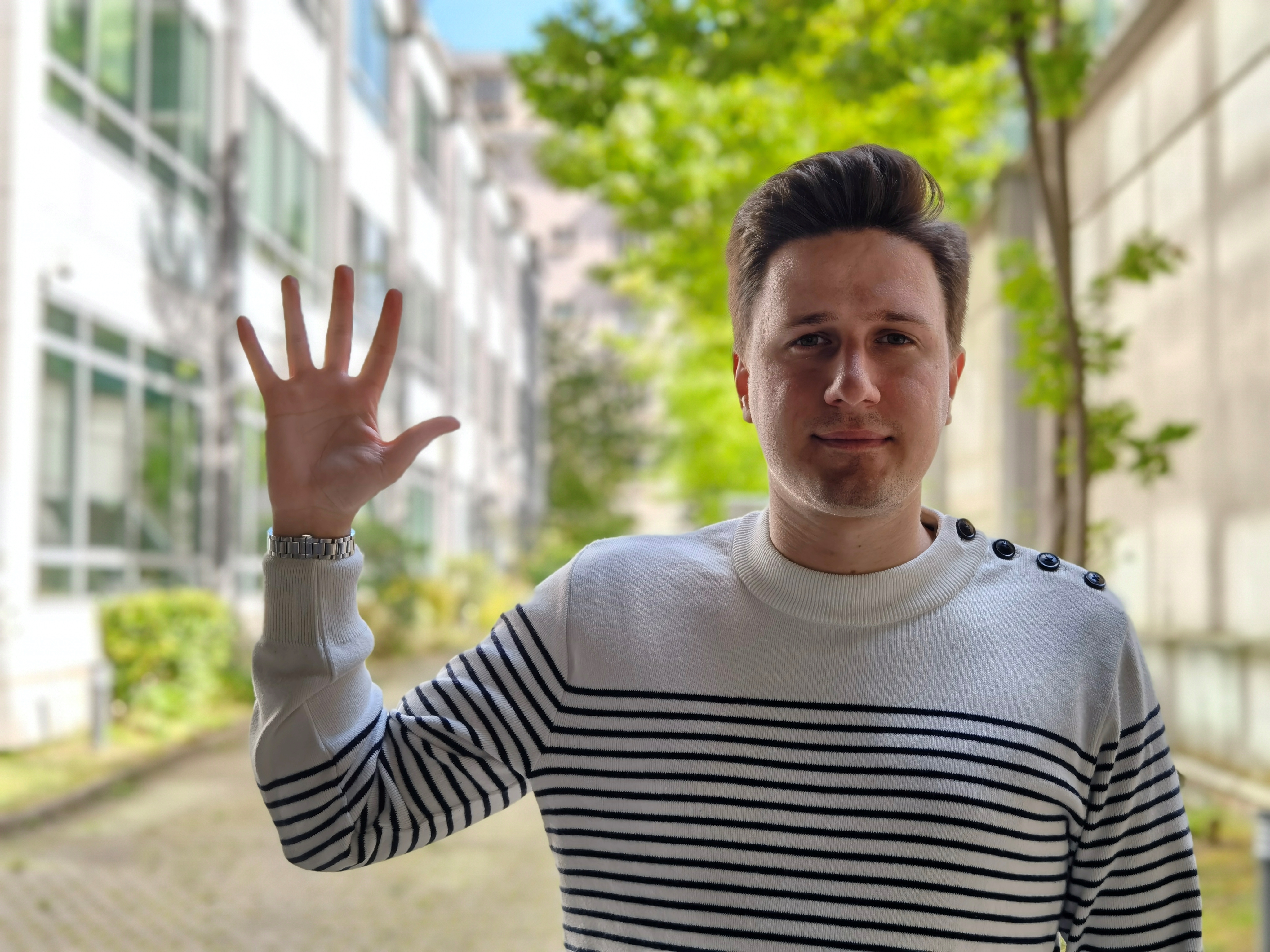} & 
    \includegraphics[width = 0.11 \textwidth]{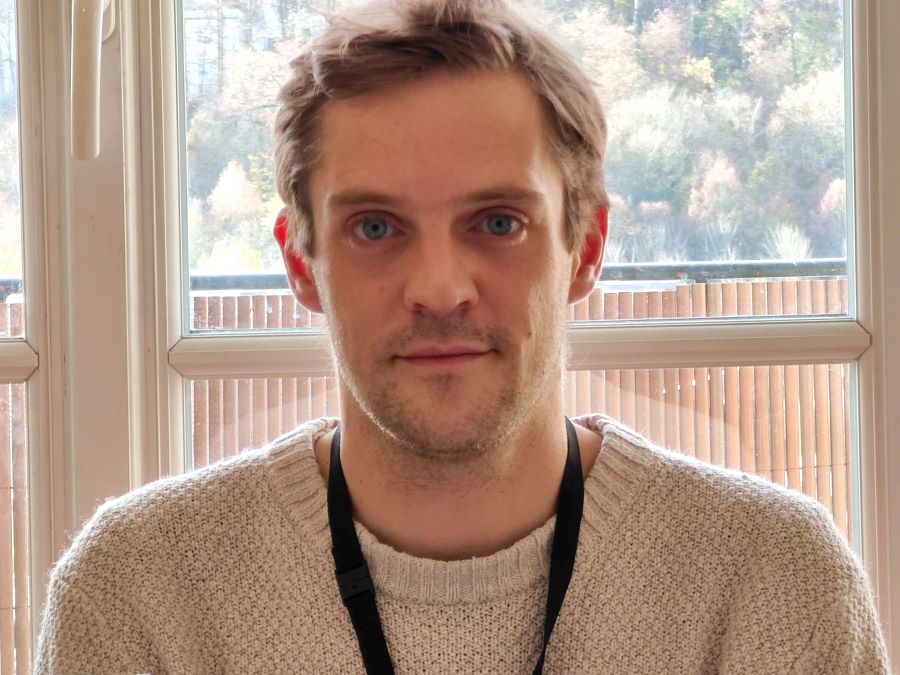} & 
    \includegraphics[width = 0.11 \textwidth]{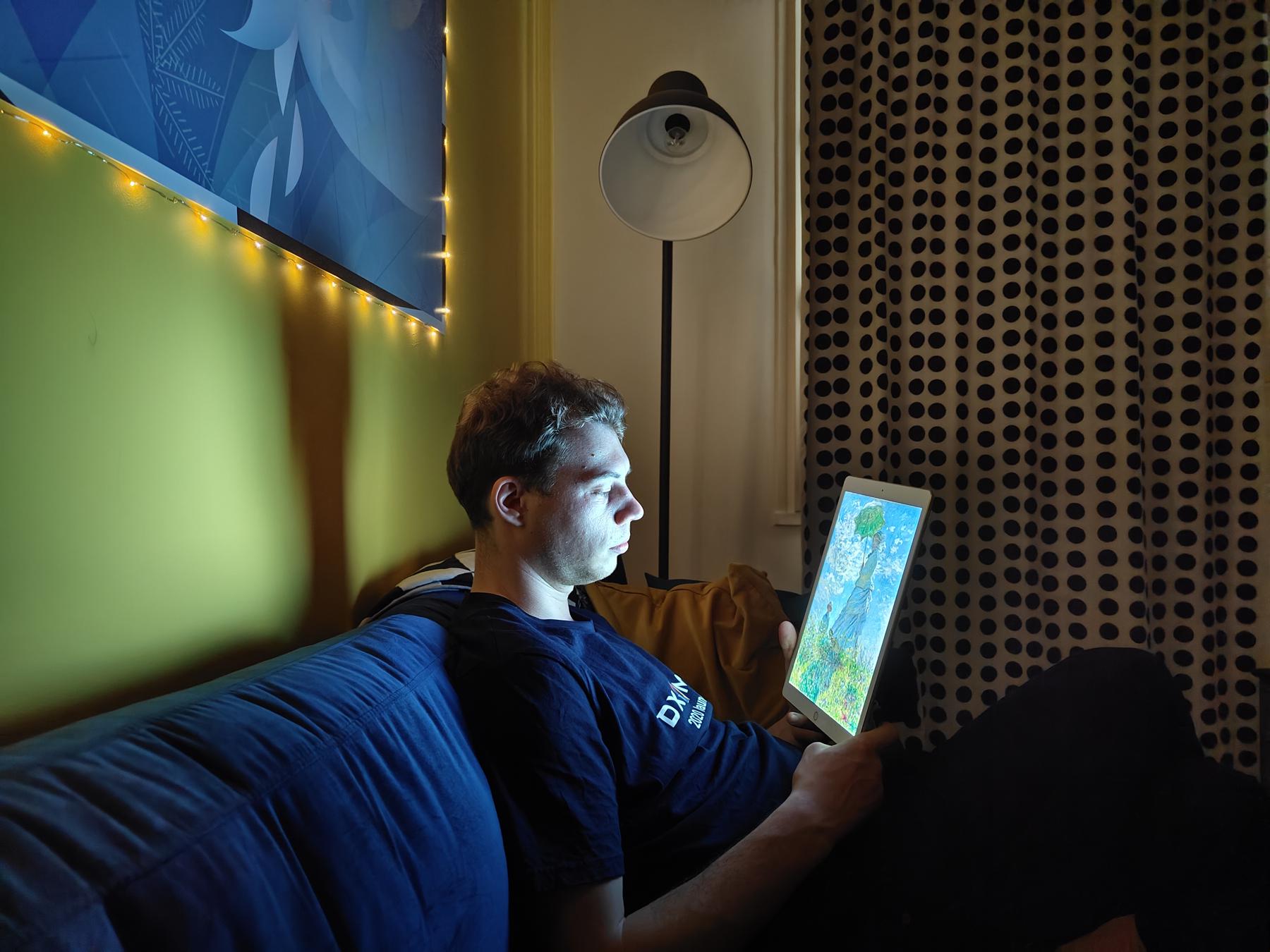} &
    \includegraphics[width = 0.11 \textwidth]{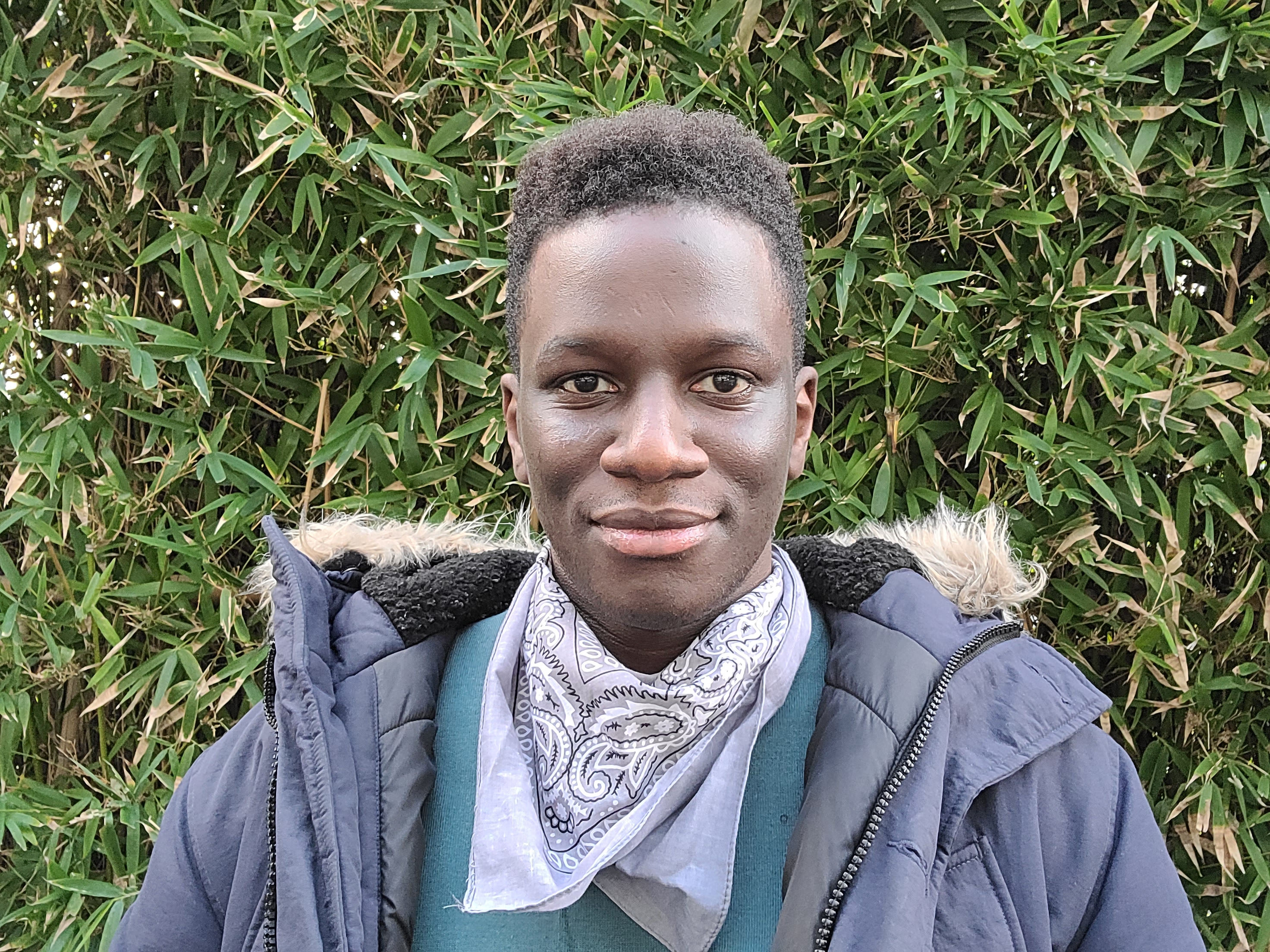} &
    \includegraphics[width = 0.11 \textwidth]{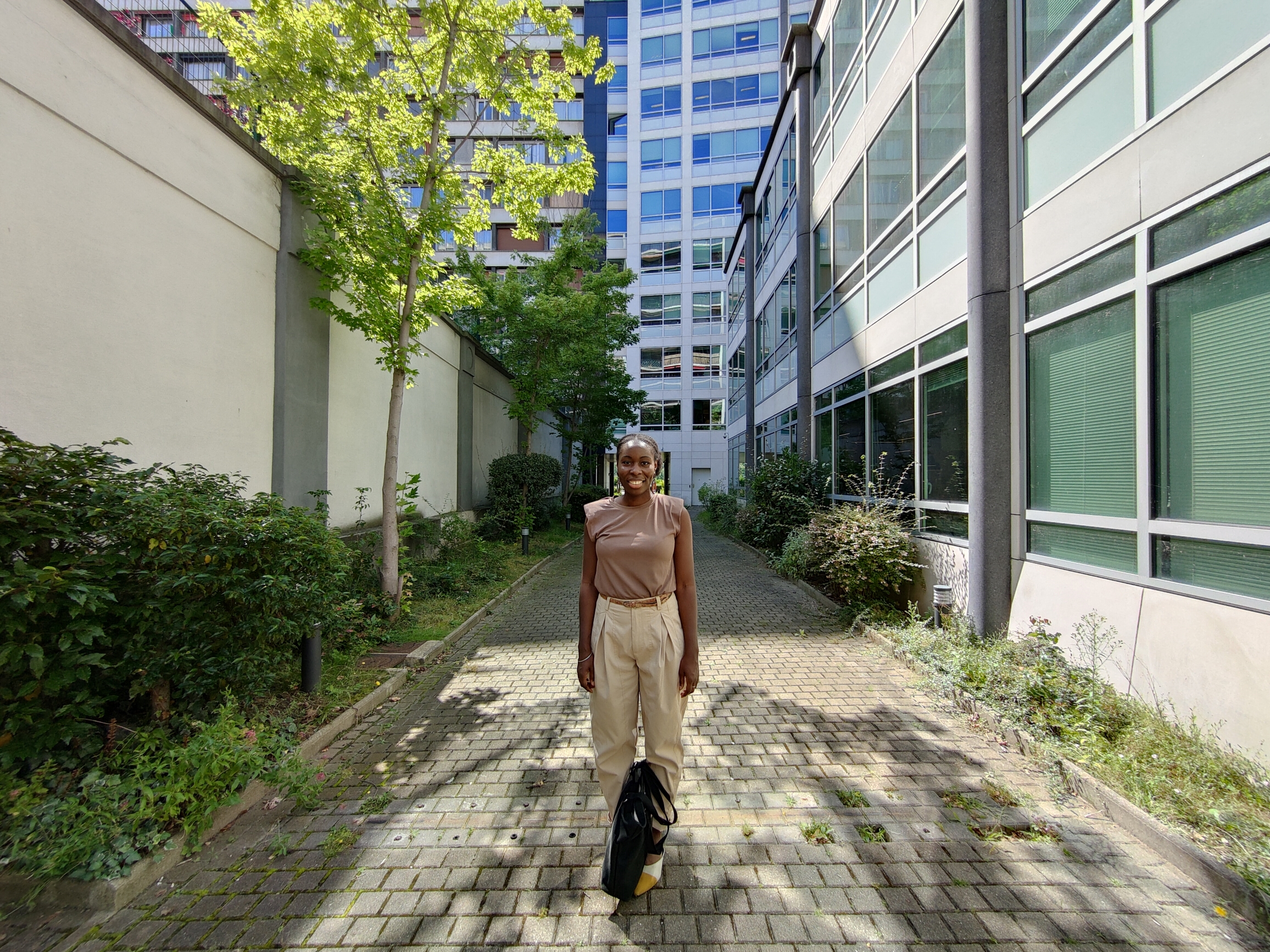} &
    \includegraphics[width = 0.115 \textwidth]{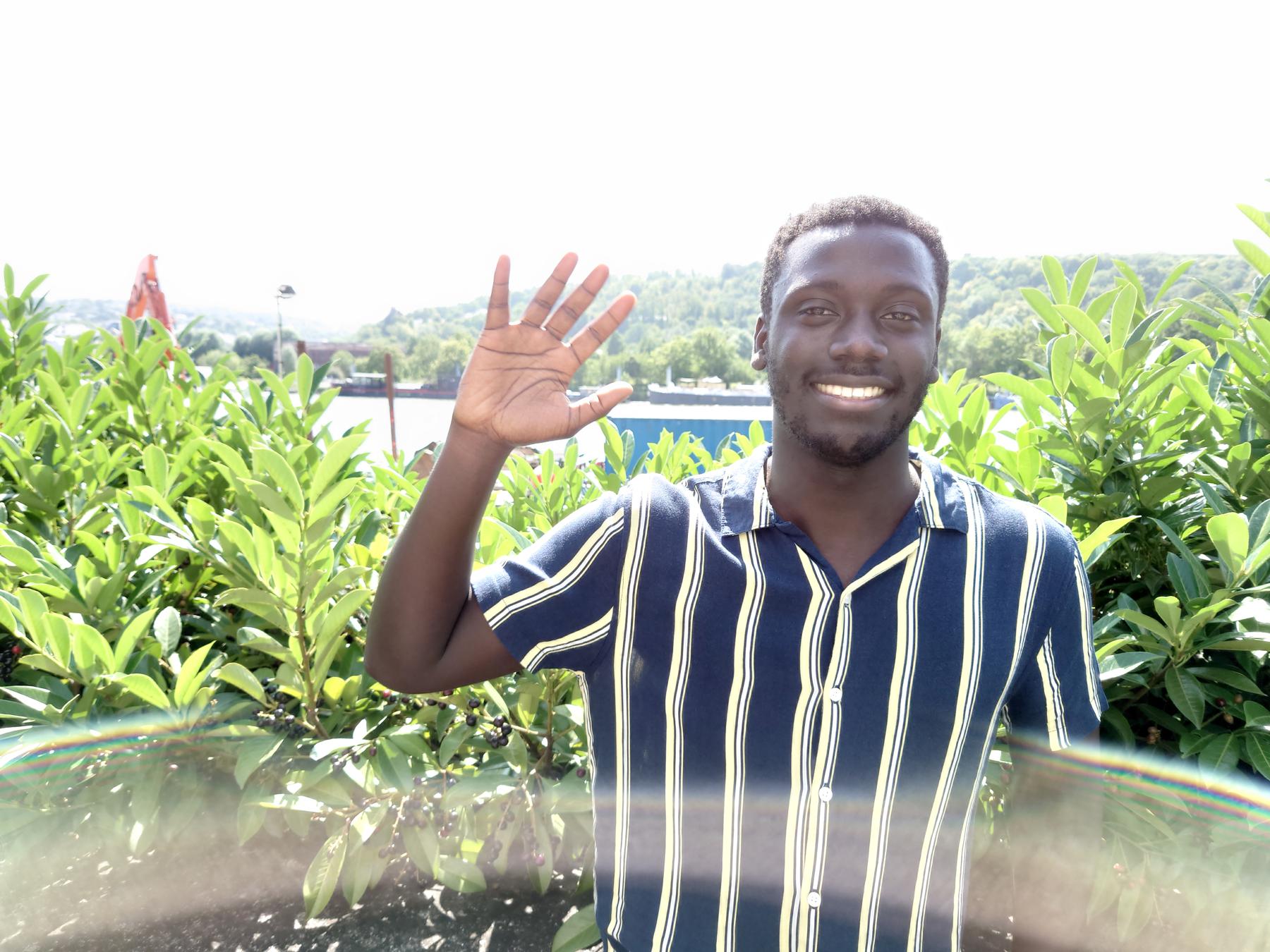} & 
    \includegraphics[width = 0.115 \textwidth]{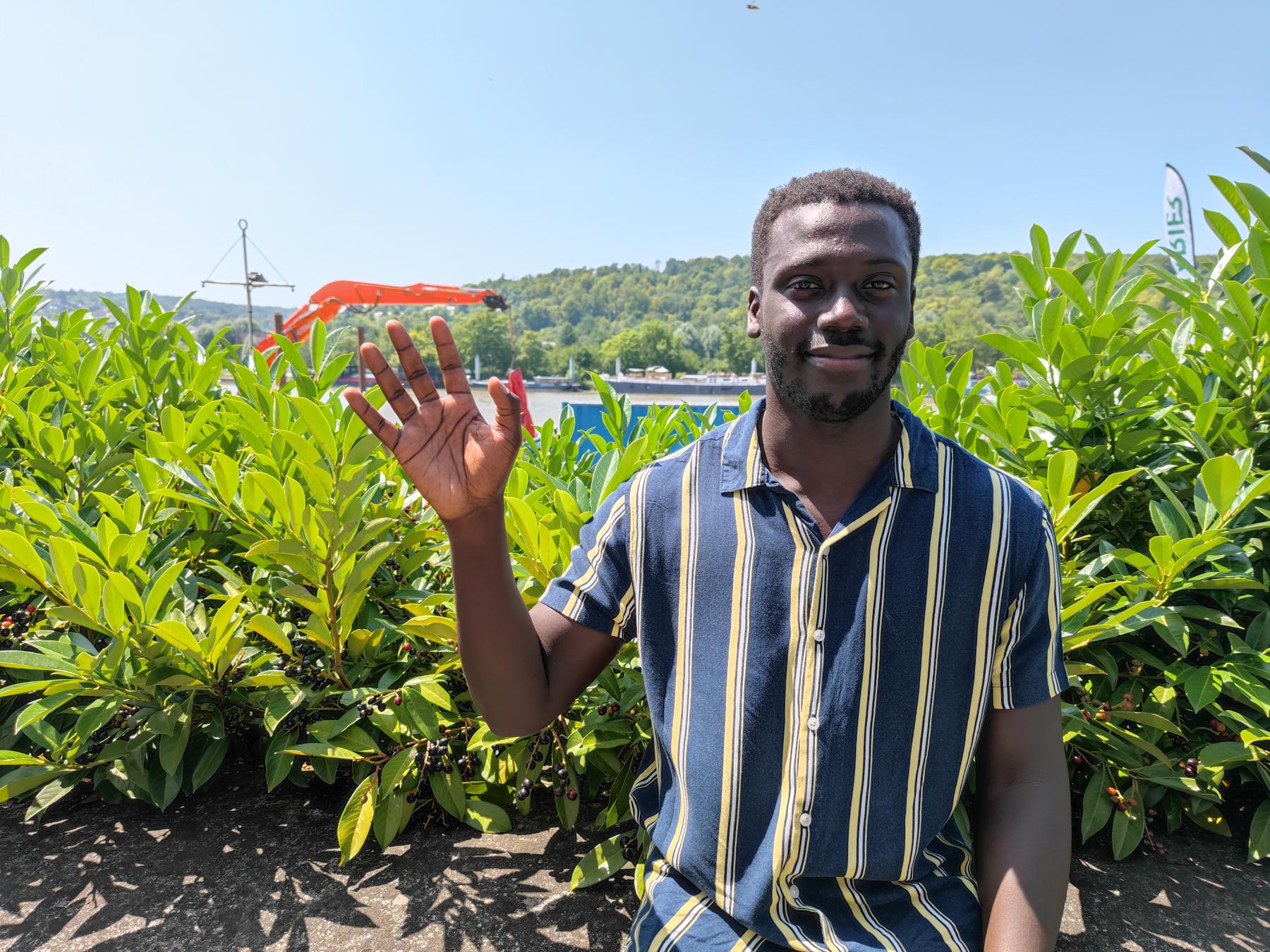} 
    \\
    \includegraphics[width = 0.11 \textwidth]{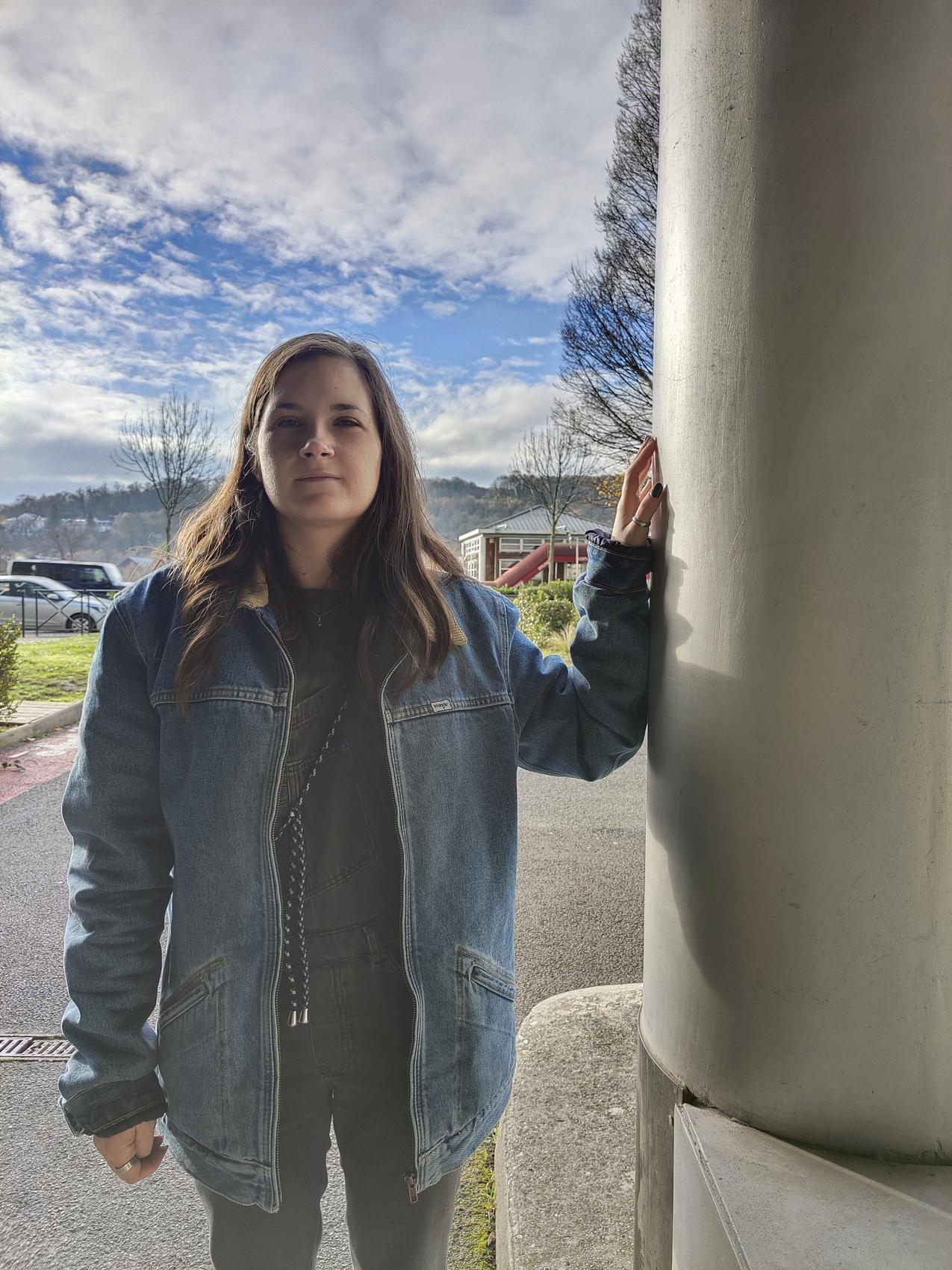} &
   \includegraphics[width = 0.11 \textwidth]{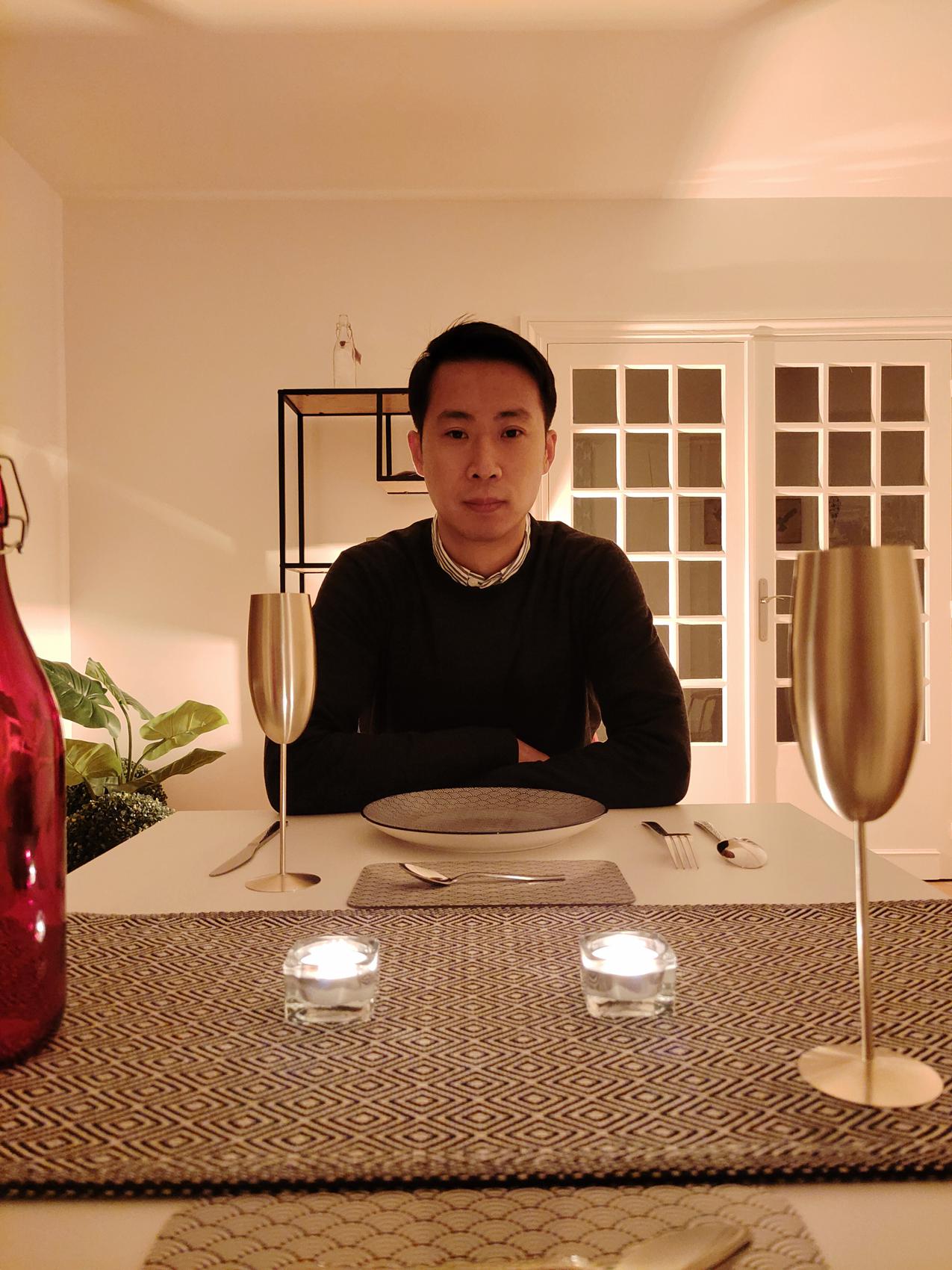} &
    \includegraphics[width = 0.11 \textwidth]{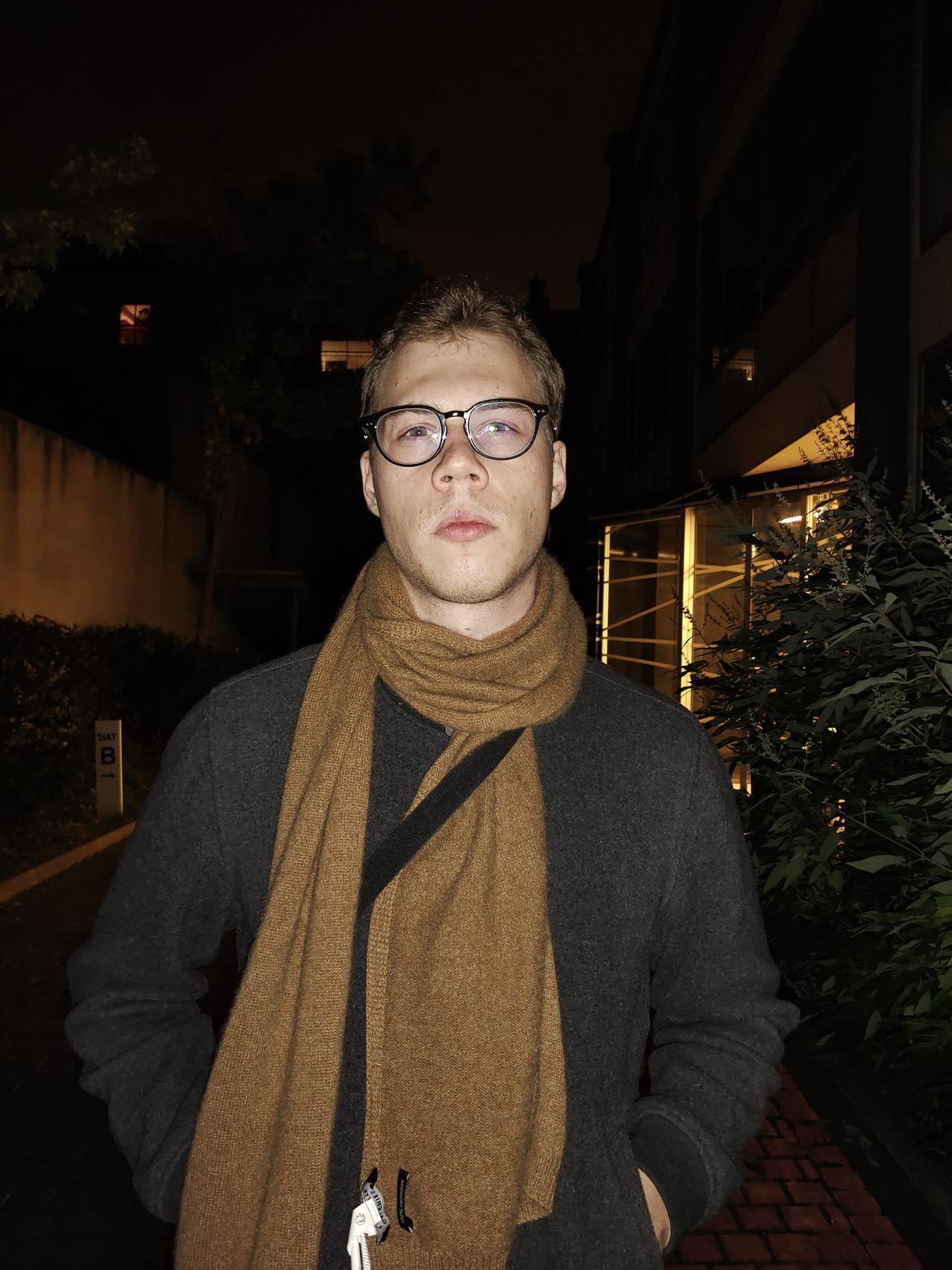} &
    \includegraphics[width = 0.11 \textwidth]{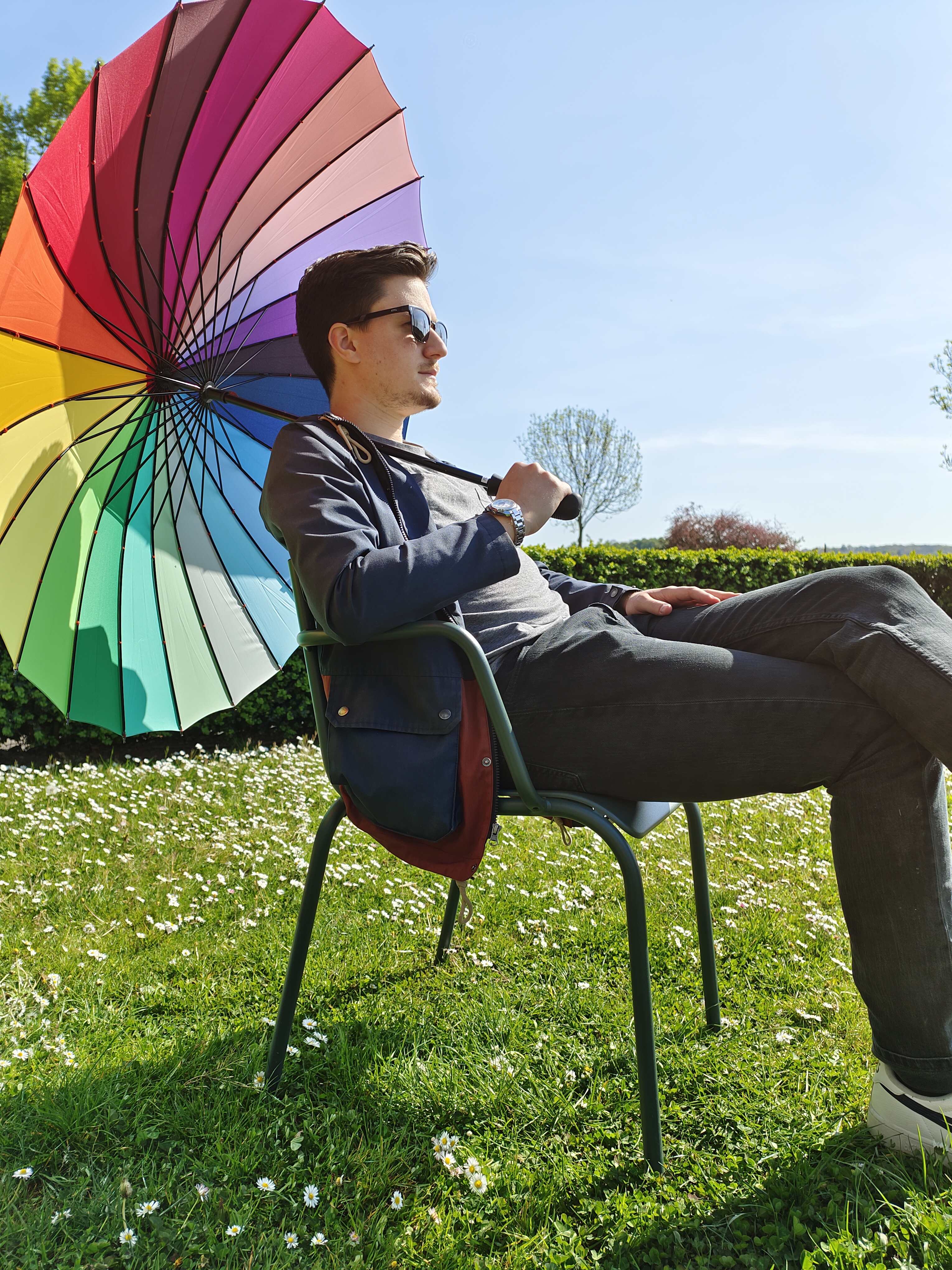} &
\includegraphics[ width = 0.11 \textwidth]{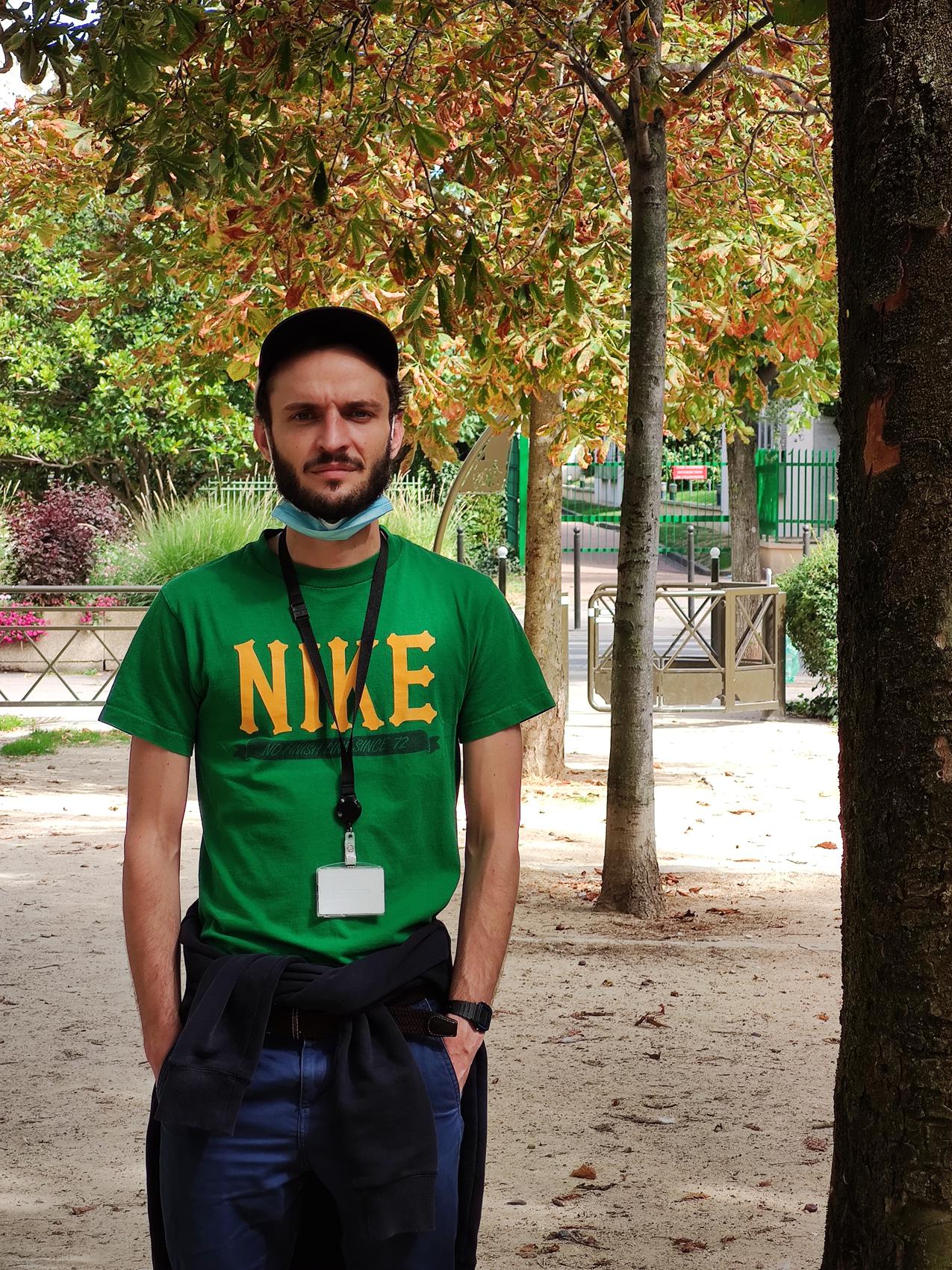} &
\includegraphics[width = 0.115 \textwidth]{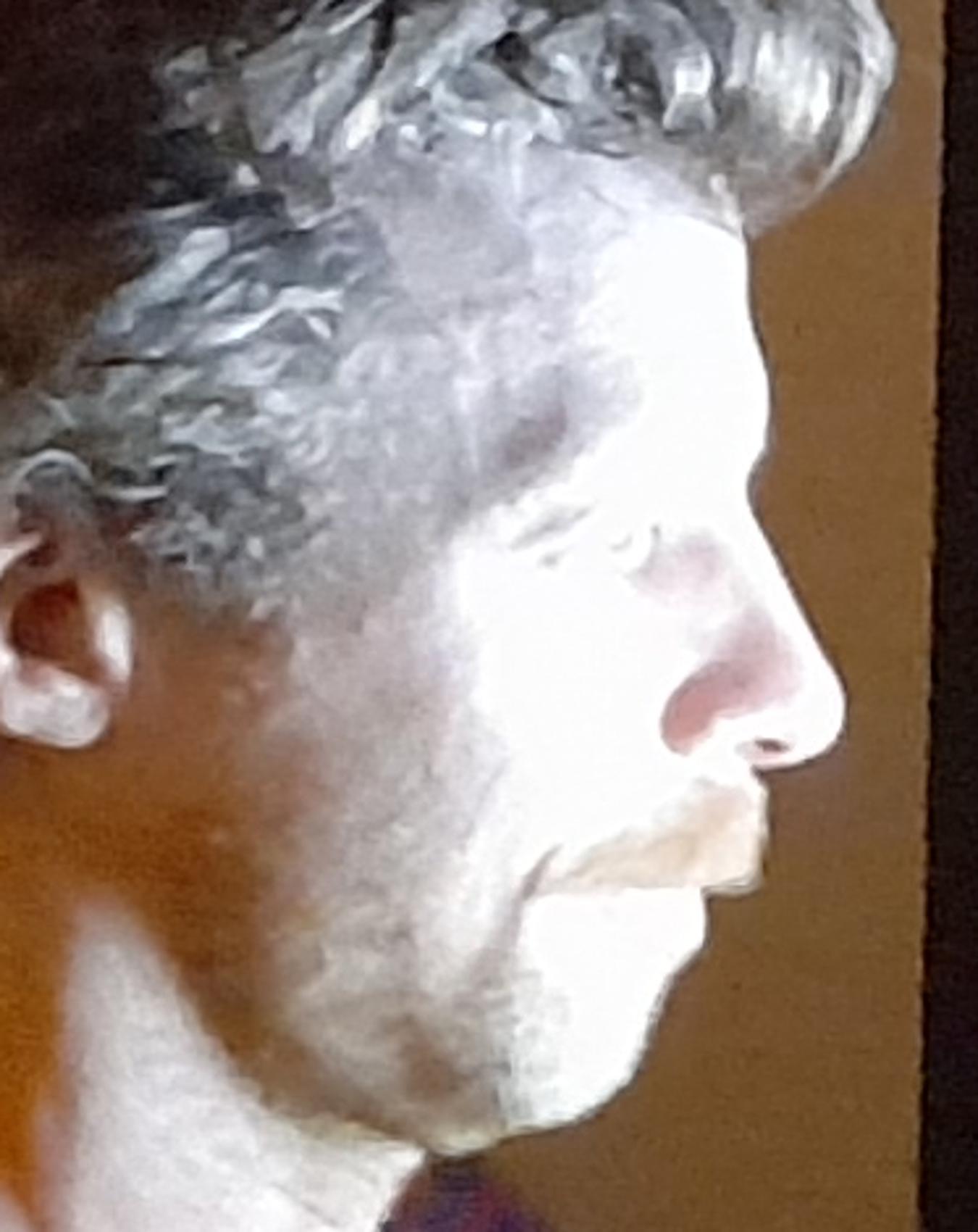} &
   \includegraphics[width = 0.115 \textwidth]{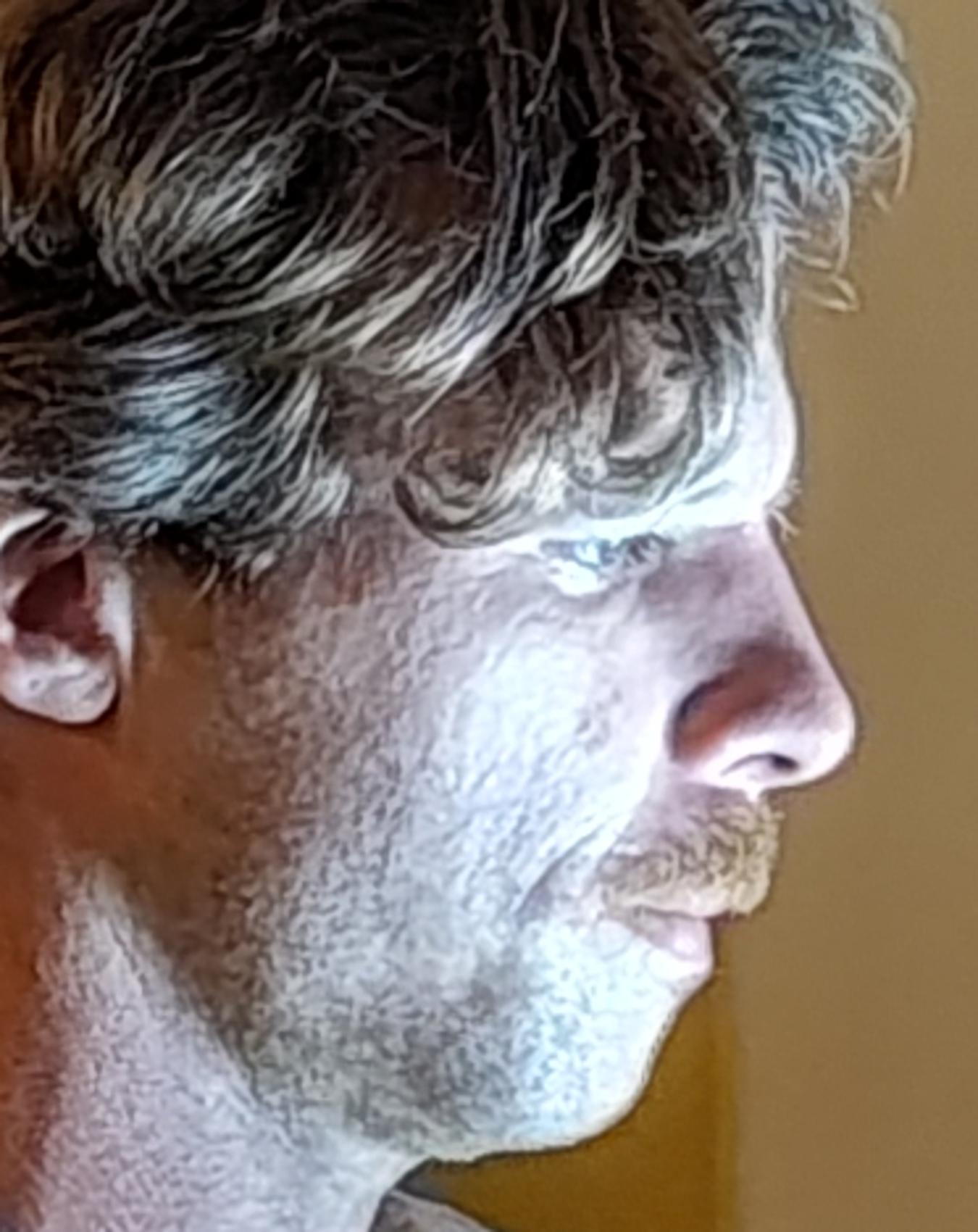} 
   \\

    \SetCell[c=5]{c} \footnotesize{(a)} 
    & & & &
    & \SetCell[c=2]{c} \footnotesize{(b)}
\end{tblr}

\caption{(a) Scenes from the PIQ23 dataset.
(b) Examples of the region of interest (ROI) used for different attribute comparisons. \textit{Top:} overall quality; we use a resized version of the full image. \textit{Bottom:} details \& target exposure; we use an upscaled face area.}

\label{fig:dataset_examples}
\end{figure*}
\section{Introduction}
\label{sec:intro}

Social media has made smartphones a vital tool for connecting with people worldwide. Visual media, particularly portrait photography, has become a crucial aspect of sharing content on these platforms.

Portrait photography serves numerous applications (\eg, advertisements, social media) and use cases (\eg, anniversaries, weddings). Capturing a high-quality portrait is a complex exercise that demands careful consideration of multiple factors, such as scene semantics, compositional rules, image quality, and other subjective properties \cite{redi2015beauty}.

Smartphone manufacturers strive to deliver the best visual quality while minimizing production costs to rival professional photography. Achieving this requires implementing complex tuning and optimization protocols to calibrate image quality in smartphone cameras.
These cameras introduce sophisticated non-linear processing techniques such as multi-image fusion or deep learning-based image 
enhancement \cite{van2019edge}, resulting in a combination of authentic (realistic) camera distortions.

This makes traditional objective quality assessment \cite{loebich2007digital, gousseau2007modeling, cao2009measuring, nicolas2021portrait} that models digital cameras as linear systems  unreliable \cite{fang2020perceptual}.
Therefore, in addition to objective measurements, the tuning process also includes perceptual evaluations where cameras are assessed by image quality experts. This procedure requires shooting and evaluating thousands of use cases, which can be costly, time-consuming, and challenging to reproduce. Automatic image quality assessment (IQA) methods that try to mimic human perception of quality have been around for many years, in order to help in the tuning process \cite{moorthy2011blind, saad2012blind, mittal2012no, ye2012unsupervised, ghadiyaram2017perceptual, xue2013learning, mittal2012making, zhang2015feature}. Blind IQA (BIQA), in particular, is a branch of IQA where image quality is evaluated without the need for undistorted reference images. Learning-based BIQA methods \cite{kang2014convolutional, kim2016fully, zhang2018blind, zhang2021uncertainty, su2020blindly, you2021transformer, ke2021musiq, golestaneh2022no, yang2022maniqa} have shown good performance on authentic camera distortion datasets \cite{virtanen2014cid2013, ghadiyaram2015massive, zhu2020multiple, hosu2020koniq, ying2020patches, fang2020perceptual}, annotated by subjective assessment of image quality. 
Annotating these datasets is considered an ill-posed problem, as the subjective opinions are not deterministic, making it challenging to use BIQA methods as accurate quality measures. Therefore, there is a need to develop a quantitative and formal framework to evaluate and compare subjective judgments in an objective manner. In this paper, we rely on pairwise comparisons performed by image quality experts along a fixed and relevant set of attributes.
Multiple attributes, including target exposure, dynamic range, color, sharpness, noise, and artifacts, define image quality \cite{vcadik2006image}. Portrait images require additional considerations, such as skin tone, bokeh effect, face detail rendering, and target exposure on the face, which fall under the scope of portrait quality assessment (PQA) \cite{nicolas2021portrait}.

To the best of our knowledge, the problem of assessing the quality of a portrait image has received limited attention. Most of the work on face IQA \cite{schlett2022face} has been directed towards improving face recognition systems and not as an independent topic. As far as we know, our paper introduces the first-of-its-kind, smartphone portrait quality dataset. We hope to create a new domain of application for IQA and to push forward smartphone portrait photography. Our contributions are the following:
\begin{itemize}
    \item A new dataset, PIQ23, consisting of 5116 single portrait images, taken using 100 smartphone devices from 14 brands, and distributed across 50 different natural scenes (\textit{scene = fixed visual content}). We have addressed the ethical challenges involved in creating such a dataset, by obtaining from each individual depicted in the dataset a signed and informed agreement,
    making it the only IQA dataset with such legal and ethical characteristics, as far as we know.
    
    \item A large IQA experiment controlled in a laboratory environment with fixed viewing conditions. Using pairwise comparisons (PWC) and following carefully designed guidelines, we gather opinions for each scene, from over 30 image quality experts (professional photographers and image quality experts) on three attributes related to portrait quality: face detail preservation, face target exposure, and overall portrait image quality.
    
    \item An in-depth statistical analysis method that allows us to evaluate the precision and consistency of the labels as well as the difficulty of the IQA task. This is particularly important given the fact that image quality labels are heavily affected by subjectivity, disagreement between observers, and the number of annotations.
    \item An extensive comparison between multiple BIQA models and a simple new method combining scene semantic information with quality features to strengthen image quality prediction on PIQ23.
\end{itemize}
\section{Related work}
\label{sec:related_work}
\subsection{BIQA}
The main goal of blind IQA (BIQA) is to predict image quality without requiring a pristine reference image. We review the datasets that already exist in this domain as well as the BIQA computational algorithms.

\paragraph{BIQA datasets.}
Early datasets like LIVE \cite{sheikh2006statistical}, CSIQ \cite{larson2010most}, TID \cite{ponomarenko2009tid2008, ponomarenko2015image} and BAPPS \cite{zhang2018unreasonable} consist of noise-free images processed with several artificial distortions. These distortions aim to describe image compression or transmission scenarios and most of them fail to capture the complexity of modern smartphone camera systems, with non-linear processing pipelines. Recent ``in-the-wild” datasets such as CLIVE \cite{ghadiyaram2015massive}, KonIQ10k \cite{hosu2020koniq} and PaQ-2-PiQ \cite{ying2020patches} consist of media-gathered images with more complex mixtures of distortions closer to real-world images.
However, due to their wild nature and uncontrolled labeling environment, they do not form a strong background to evaluate the quality of digital cameras, which we are most interested in. As an early effort on this topic, Virtanen \etal \cite{virtanen2014cid2013} have developed a database (CID2013) that spans 8 visual scenes with 79 digital cameras. In recent work, Zhu \etal  \cite{zhu2020multiple} provide a smartphone IQA dataset (SCPQD2020) of 1800 images shot with 15 devices on 120 visual scenes. They annotate the database in a well-controlled laboratory, by three image quality experts. Fang \etal published SPAQ \cite{fang2020perceptual}, a smartphone IQA dataset with 11125 images shot with 66 devices. Both datasets provide multiple attribute evaluations and scene categories. They include generic visual content and do not deal with PQA. While SCPQD2020 lacks in the number of observers, SPAQ relies on resized images which heavily degrades the quality. All previously mentioned datasets, except TID2013 \cite{ponomarenko2015image} and BAPPS \cite{zhang2018unreasonable}, rely on rating systems (MOS), and do not pay close attention to the difficulty of cross-content observations.
In PIQ23, we provide 50 scenes, each annotated independently. We collect opinions from over 30 image quality experts by pairwise comparisons, which has been shown to be more consistent in IQA experiments \cite{mantiuk2012comparison, perez2019pairwise}. We also analyze the uncertainty and consistency of our annotations through a new statistical analysis method.
\paragraph{BIQA methods.}
BIQA can be separated into classical and deep learning approaches. Early learning-based approaches \cite{moorthy2011blind, saad2012blind, mittal2012no, ye2012unsupervised, ghadiyaram2017perceptual} use a combination of hand-crafted statistical features (natural scene statistics) to train shallow regressors (\eg SVR). Other approaches try to estimate the quality without the need for training \cite{xue2013learning, mittal2012making, zhang2015feature}. These methods perform relatively poorly on modern IQA datasets, as they do not fully reflect the human perception of realistic distortions \cite{ghadiyaram2015massive, zhang2018blind}. 
Consequently, deep BIQA models have been surging in the last decade. Multiple convolutional neural networks (CNN) based methods \cite{kang2014convolutional, kim2016fully, zhang2018blind} have demonstrated solid performance on modern datasets. Zhang \etal \cite{zhang2021uncertainty} address the problem of uncertainty in IQA and present a method to simultaneously train on multiple datasets using image pairs as training samples. Su \etal \cite{su2020blindly} try to separate semantic features from image quality features by training an adaptive hyper network that captures semantic information. Recent works that adopt transformer architectures \cite{you2021transformer, ke2021musiq, golestaneh2022no, yang2022maniqa} to extract global quality information, have shown impressive performances on IQA datasets. Because of the per-scene annotation structure of our dataset, we adopt a semantics-aware multitasking method to adapt the scale and features to the input scene. 

\subsection{PQA}
Despite the lack of portrait quality assessment (PQA) research, solely focusing on evaluating portrait quality,
we still recognize the importance of face IQA (FIQA). FIQA aims to assess the quality of face images to boost the performance of face recognition algorithms
\cite{best2018learning, rose2019deep, hernandez2019faceqnet, lijun2019multi, yang2019dfqa, ou2021sdd, grother2020ongoing, schlett2022face}. The closest FIQA work to PIQ23 is that of Zhang \etal \cite{zhang2017illumination}, where they have developed a dataset to objectively evaluate the illumination quality of a face image. Redi \etal \cite{redi2015beauty} define a set of attributes to evaluate the ``beauty" of the portrait. Kanafusa \etal \cite{kanafusa2000standard} propose a method to define a standard portrait image, which can be later used to evaluate color rendering and other attributes between cross-media. In this work, we can see a first attempt to use a standard portrait as a subject for IQA. Chahine \etal \cite{nicolas2021portrait} proposed the first approach to evaluate specific face attributes as a separate metric for PQA on realistic mannequins. Finally, Liang \etal \cite{liang2021ppr10k} have developed a large-scale portrait photo retouching dataset, with multiple use cases and cameras. To the best of our knowledge, PIQ23 is the first smartphone PQA dataset, with a variety of visual scenes, legal validation, and expert annotations.

\subsection{Domain shift}
\label{subsub:domainshift}
The annotation strategies and image content can vary widely between different IQA datasets. Hence, their respective quality scales are usually relative and independent. With this characteristic, we encounter a problem known as domain shift \cite{zerman2017extensive, zhang2020learning, zhang2021uncertainty, sun2021blind}. Since quality scales are relative, similar scores may not indicate the same level of perceptual quality across different datasets. This can lead to confusion when attempting to learn from multiple sources. As a result, understanding image semantics is necessary. Current BIQA models implicitly try to learn semantics and quality simultaneously. However, it is extremely difficult to merge these two problems, as they seem to be contradictory \cite{fang2020perceptual, kendall2018multi}. Some papers try to solve this problem using multitask learning \cite{huang2020multi, fang2020perceptual, zerman2017extensive, sun2021blind}.
Explicitly separating semantic information from quality is not well represented in previous works. Su \etal \cite{su2020blindly} propose HyperIQA, a self-adaptive hyper network that implicitly extracts semantic information and adapts the quality prediction accordingly. The hyper network, however, is not trained to predict categories explicitly. Since PIQ23 consists of multiple relative content-dependent scales, we propose to combine multitasking with HyperIQA in order to adapt the quality scale of each scene based on semantic understanding. 

\section{PIQ23}
\label{sec:PRIMARQ22}

\subsection{Dataset details}
\label{sub:Challenges}

\paragraph{Legal aspects.}

We believe that unrestricted access to PIQ23 for public research is crucial. Accordingly, we have taken steps to address any potential legal obstacles that may obstruct this access.
All individuals in the photos have given explicit permission for image rights via signed transfer and received a privacy notice detailing how their images will be processed.
Also, to ensure the effectiveness of people's rights, we have tagged each photo with a unique identifier assigned to each person by using a face clustering algorithm. This pseudonymization technique prevents access to individuals' names by dataset users. 
Finally, we contractually require all dataset users to comply with relevant data protection laws, including the GDPR.
\paragraph{Dataset properties.}
We have constructed PIQ23 with the intent of reducing annotation biases and covering a variety of common real-life scenarios. To achieve this goal, we have broken down the factors affecting the quality of a portrait image. We consider lighting to be 
a primary element influencing the quality. Hence, we have separated lighting conditions into four groups: outdoor, indoor, low light, and night. Also, we have paid attention to lighting homogeneity, which describes the reflection of the light on the subject (\eg front light, side light, backlight). The characteristics of the subject, such as age, skin tone, gender, subject position, framing, face orientation, movement, and subject-to-lens distance play an equally important role.
Our skin tone ruler is based on the Fitzpatrick skin type (FST) \cite{fitzpatrick1988validity}. 
We have tried to cover a sizeable chunk of smartphone devices and brands utilized over the past decade. Additionally, we have included diverse smartphone camera lens focal lengths such as zoom, wide, and selfie along with distinct camera modes such as night and bokeh. Furthermore, we have contemplated the possibility of augmenting our dataset with high-quality images sourced from DSLR cameras. Nonetheless, we have found through our experimentation that artificially distorting DSLR images to ensure comparability with photos taken by smartphones is a challenging task. As a result, we have excluded DSLR cameras from our dataset. 

To comply with the previous description, we have designed a collection of 50 distinct portrait scenarios (referred to as ``scenes"), captured in predetermined locations that encompass a diverse range of factors (see \cref{fig:dataset_examples} (a)). The dataset images were taken with about 100 smartphones (2014- 2022) from 14 brands and different price segments. We have collected around 5116 images, averaging 100 images per scene. We note that PIQ23 was subsampled from a larger dataset that was collected over a long period of time (a couple of years) and is a result of cumulative efforts in engineering and photography. We, therefore, believe in its capacity to cover a broad spectrum of smartphone photography. More information about the PIQ23 characteristics can be found in the supplementary material.

\subsection{Portrait quality assessment}

\paragraph{Portrait quality attributes.}
In a portrait, most attention is given to the person depicted, which is known as the human region priority (HRP) \cite{liang2021ppr10k}.

Portrait quality may vary significantly depending on the application. For example, Redi \etal
\cite{redi2015beauty} try to define all the characteristics to capture the 'beauty' of a portrait. Quality in this case is strongly correlated with beauty and aesthetics. In other applications, such as FIQA \cite{schlett2022face}, quality assessment is a measure of utility, to filter out poor quality faces from face recognition systems. Neither application totally aligns with PQA \cite{nicolas2021portrait}. 
Thus, we intend to broaden the research on PQA by studying a preliminary group of three attributes: face detail preservation, face target exposure,
and overall image quality. Additionally, we have conducted a study concerning a fourth attribute, namely global color quality. However, due to the difficulty in annotating this attribute through pairwise comparisons on different content, 
we have decided to exclude it from our dataset (\cref{fig:StatsGraph}). The annotation guidelines can be found in the supplementary material. 
\paragraph{Annotation strategy.}
Perception-based IQA experiments present a high degree of difficulty and are usually subjective. Opinions can vary widely depending on multiple factors: viewing conditions, the observer's cultural and professional backgrounds, image content, etc.
The objective of PIQ23 is to deliver image quality annotations obtained (as much as possible) from impartial and unbiased observations. To maximize objectivity and consistency, we propose two elementary steps:  
\begin{itemize}
    \item First, we have chosen to annotate each scene separately using a forced-choice pairwise comparison approach (PWC). Combined with the active sampling technique proposed in \cite{mikhailiuk2021active}, we have been able to reach good annotation consistency with a minimal number of comparisons (see \cref{sec:stat_analysis}).

    \item Second, we have fixed the region of interest (ROI) for each attribute, as shown in \cref{fig:dataset_examples} (b). For details preservation and target exposure, we have extracted the face area using RetinaFace \cite{deng2020retinaface}. We have then upscaled it using standard bicubic upsampling to a reference resolution of about 4.5 megapixels with a fixed aspect ratio. 
    For color and overall attributes, we have resized the images to an approximate Full HD resolution (about 2.5 megapixels) while keeping the original aspect ratio (i.e. portrait or landscape).
    
\end{itemize}

\paragraph{Experiment details.}
We have reached out to professional photographers and experts with a solid background in photography and image quality to help us annotate the dataset.
The opinions of more than 30 experts were gathered using an internal PWC tool. Observers were asked to select the best out of two images, following the guidelines described in the supplementary material.
We have adapted our settings so that the viewing conditions are aligned with that of a human eye, with a cutoff frequency $\nu_{cut}$ = 30cpd. Hence, we have used a BenQ 32" 4k monitor with a pixel pitch of 0.185, and we have fixed the eye-to-screen distance at 65cm. We have calibrated the display to standard sRGB settings (D65 white point with luminance $\geq 75 cd/m^2$ with no direct illumination of the screen and a background illumination with a lighting panel set to D65 / 15\% for reducing eye stress). We have also converted all images in DCI-P3 color space to sRGB. We have kept the sessions short, around five minutes per attribute, in order to reduce fatigue and stress on the observers.
The annotation procedure took around eight months. 
 For each scene and each attribute, we have collected around 4k pairwise comparisons, a total of 600k data points. 

 Though, for a limited number of comparisons, given the subjectivity of the task, noise and outliers are commonly encountered. In \cref{sec:stat_analysis}, we present a new statistical analysis method to rectify this noise.

\begin{figure*}[!h]
    \centering    \includegraphics[width = 0.965\textwidth]{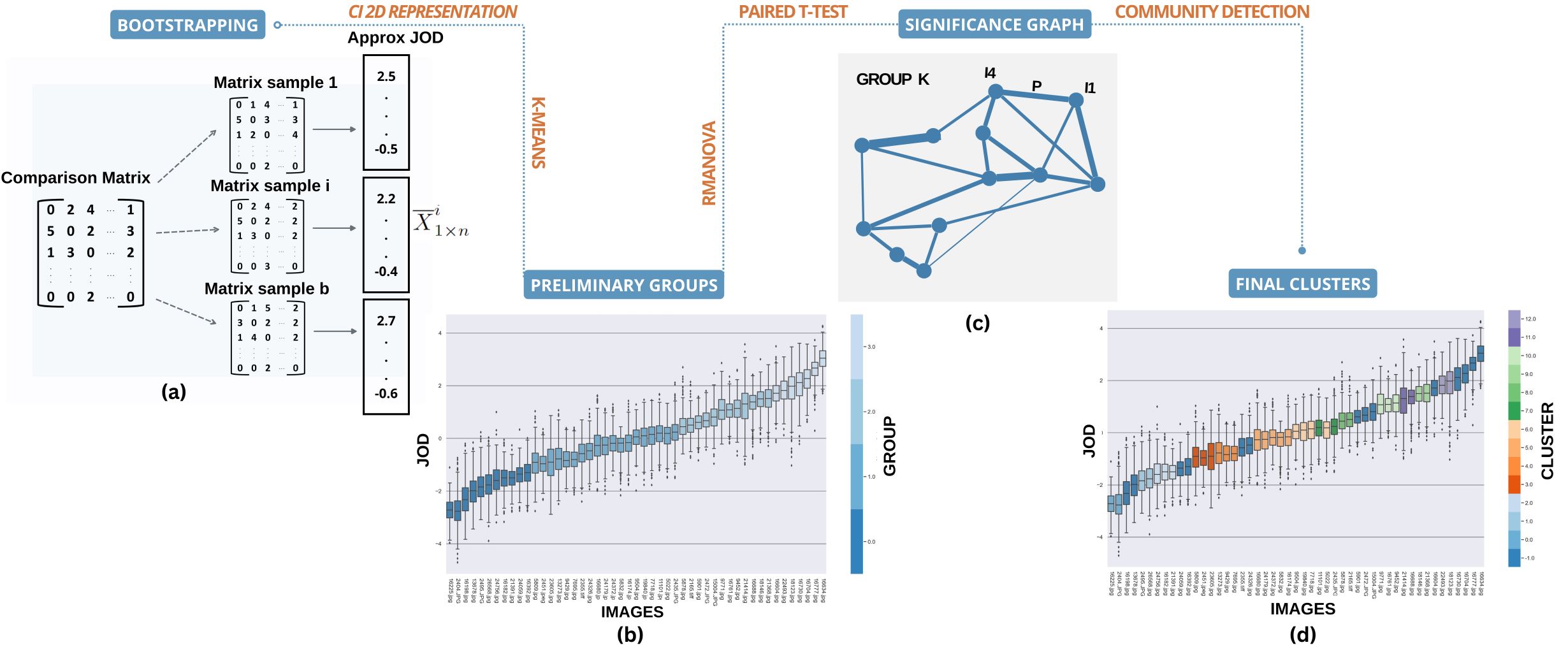}
 \caption{ {Diagram of the statistical analysis strategy used to estimate the uncertainty in a PWC experiment. (a) Given a PWC matrix, we generate confidence intervals (CIs) using percentile bootstrapping. (a-b) We then apply the K-means algorithm to the ``2d representation" of the CIs to cluster the images into preliminary quality groups (b). (b-c) In each group, we apply RMANOVA to detect significant differences between the JOD scores. (c) For groups with such differences, we construct a weighted undirected graph, where the weights consist of the p-value of paired t-tests between the image pairs. (c-d) Finally, we apply Louvain community detection to extract sub-clusters of similar quality inside each group (d). Figures (b) and (d) represent the boxplots of the bootstrapped image scores generated in (a).}}
    \label{fig:StatAna}
\end{figure*}

\section{Statistical analysis}
\label{sec:stat_analysis}
We present a new approach to quantifying uncertainty in IQA experiments. We recall in \cref{subsub:uncertainty} how quality scores are extracted from a PWC experiment and how uncertainty can be estimated using bootstrapped confidence intervals. We then introduce in \cref{sub:Beyond_CI} a new statistical analysis strategy to go beyond the calculation of confidence intervals. The complete pipeline is illustrated in Fig. \ref{fig:StatAna}

\subsection{Psychometric scaling and confidence intervals}
\label{subsub:uncertainty}
\paragraph{Psychometric scaling.}
\label{subsub:Limitations}
Designing a PWC experiment requires modeling the statistical distribution of the image quality. Commonly, the quality of an image is described by the Thurstone Case V observer model \cite{thurstone1994law, davidson1976bibliography} as a Gaussian distribution $\mathcal{N}(\mu,\sigma)$. The average $\mu$ represents the actual quality and $\sigma^2$ is its ``perceptual" variance across observations. The latter encompasses the intra-variance and inter-variance of the perceptual quality. The intra-variance represents the uncertainty of one observer when the observation is repeated multiple times. The inter-variance represents the uncertainty across multiple observers. Based on this formulation, psychometric scaling methods \cite{perez2019pairwise, perez2017practical} transform the comparison matrix $M$ constructed from a PWC experiment, into a continuous scale of image scores representing the average opinions across multiple observers.
The results are typically expressed in Just-Objectionable-Difference (JOD) units \cite{perez2019pairwise}. Two images are 1 JOD apart if 75\% of observers choose one as better than the other. In our work, we have adopted the psychometric scaling method proposed by Mikhailiu \etal \cite{mikhailiuk2021active}. The authors propose an efficient active pair selection technique via approximate message passing and information gain maximization, combined with the TrueSkill scaling algorithm \cite{herbrich2006trueskill} to minimize the PWC experiment cost. Thus, $M$ is typically a very sparse matrix (so-called incomplete design) with a limited number $c$ of non-zero elements.
\paragraph{IQA limitations.}
The choice of image and observer samples plays a critical role in the accuracy of the JOD scores. From a statistical point of view, 
these samples are taken from infinitely large populations of images and observers respectively.  
When sampling images of similar quality, for example, the comparison becomes harder, 
requiring a correspondingly larger number of comparisons than a sample of images with distinguishable quality differences. Similarly, a sample of inexperienced observers generally leads to noisier annotations, requiring more observers than a sample of experts. In addition, psychometric scaling algorithms introduce an estimation error that is inversely proportional to the size of the data.
In conclusion, estimating the difficulty of the comparison task (linked to image sampling), the quality of the experiment (linked to observer sampling), and the precision of the psychometric scaling algorithm, contribute to what we call the experiment error, which is the image JOD score estimation error.

\paragraph{Estimating the uncertainty in IQA.}
\label{subsub:estimate_uncertainty}
One way of quantifying the experiment error is by calculating the confidence interval (CI) of the JOD scores. 
The original formulation of the CI does not directly apply to PWC experiments, since it does not take into account the error introduced by the image sampling process \cite{montag2003louis}.
A practical alternative for computing CIs is bootstrapping \cite{efron1992bootstrap}. We follow the approach proposed in \cite{perez2017practical} and resort to the percentile method of bootstrapping \cite{davison1997bootstrap}. We repeatedly generate JOD scores by sampling, with replacement, the observer comparison matrices, each of which is a unique opinion on all images. The CI boundaries for each image are then defined as the 2.5th and 97.5th percentiles of the JOD scores (\cref{fig:StatAna} (a)).
\subsection{JOD clustering via confidence intervals}
\label{sub:Beyond_CI}
\paragraph{CI limitations.}
Confidence intervals represent well the sample mean error for independent variables and samples, 
but in PWC these conditions are not achieved \cite{perez2017practical}.
The psychometric scaling algorithm 
calculates all the JOD scores at the same time.
This means that every change in the comparison matrix will simultaneously affect the scores of all images. This behavior makes the image JOD scores somewhat interdependent, which is not apparent in the CIs.
To identify which images have a significant difference in quality, we need to analyze the overlap of their confidence intervals. But how to quantify this overlap? Do we consider the overlap of 20\%, 30\%, or 60\% to be significant? What to do in case of multiple overlaps? We address these issues by combining two techniques. We first cluster the images using their CIs, then, we use variance analysis to identify which images have significant quality differences.

\paragraph{Preliminary clustering.}
Overlapping CIs may indicate possible quality similarity, so we can use this information to group the images. To define the distance between intervals, we consider each CI as a point $C(x,y)$ in the subspace $\{(x,y) \in \mathbb{R}^2$ $|$ $y-x>=0\}$ where $x$ is the lower bound of the CI, and $y$ is the upper bound of the CI. In this way, we can calculate the Euclidian distance between $C_1$ and $C_2$. We then resort to the K-means algorithm \cite{macqueen1967classification}
to define our preliminary quality groups (\cref{fig:StatAna} (a-b)). 
We estimate the number of preliminary quality groups by dividing the total JOD range by the median size of the CIs (\cref{fig:StatAna} (b)).

\subsection{JOD clustering via variance analysis}
\paragraph{Variance analysis.}
To estimate the significance of the CI overlaps, we turn to the analysis of variance (ANOVA) \cite{fisher1919xv, fisher1921014} and particularly repeated measures ANOVA (RMANOVA) \cite{gueorguieva2004move}. RMANOVA is a statistical significance test used to investigate the differences in mean scores of a given continuous variable (called dependent variable), that has been ``repeatedly tested", on the same group of subjects, under three or more different conditions taken from a categorical variable (called the within-subject factor or the independent variable).
In a PWC experiment, we interpret the set of images as the independent variable and consider each bootstrapped matrix (\cref{subsub:estimate_uncertainty}) as a subject that was tested on different conditions (one matrix $\equiv$ one subject). Finally, since the JOD scores are estimated from the same matrix, we consider them as measurements of the dependent variable, that is, the image quality. 
\paragraph{Statistical hypothesis.}
Let $M$ be the sparse comparison matrix defined in \cref{subsub:Limitations}.  
Let $\mathbb{X}=\{\overline{X}_{1 \times n}^i, i=1, \dots,  b\}$, where $\overline{X}_{1 \times n}^i=\left[\begin{array}{c c c c}\overline{x}^i_{1}& \overline{x}^i_{2}& \hdots & \overline{x}^i_{n}\end{array}\right]$, be the set of score vectors inferred from $b$ comparison matrices bootstrapped from $M$, and $n$ the number of images.  
We define the two hypotheses: 
\begin{equation}
    \begin{cases}H_{0}: & \bar{x}_{1}=\bar{x}_{2}=\cdots=\bar{x}_{n}; \\ H_{1}: & \text {At least two means are different.}\end{cases}
    \label{eq:hypothesis_test}
\end{equation}
where $\overline{x}_k$ represents the inferred average score of the image $I_k$ from the PWC experiment. Refusing the null hypothesis $H_0$ only guarantees that at least two image scores are different ($H_1$). Accepting $H_0$ means that all images in the test have indistinguishable quality.
We apply the previous hypothesis testing on each of the preliminary groups and deduce whether there is a significant difference between the images or not (\cref{fig:StatAna} (b-c)). We can identify two cases:
\begin{enumerate}
    \item \textbf{No significant difference has been found:} we consider in this case that all the images of the group have the same average score and variance.
    \item \textbf{A significant difference exists:} we don't know how many images are significantly different. In this case, we conduct a post hoc analysis, using paired t-tests \cite{student1908probable}, at a confidence level of 0.95, on all the possible pairs in the given cluster.
\end{enumerate}

\paragraph{Significance graph.}
For the groups where the significant difference exists, we create a weighted undirected graph by weighting the connections between pairs with the $p$-value of their corresponding paired $t$-tests  
 (\cref{fig:StatAna} (c)). Then, we apply the Louvain community detection algorithm \cite{blondel2008fast} to group dense regions of nodes into the same ``community" or cluster (\cref{fig:StatAna} (c-d)). With this method, we separate the graph into sub-clusters, then assign their average score and variance to the corresponding images (\cref{fig:StatAna} (d)).
\subsection{Results and discussion}
\label{parag:discussonStatsAna}
\begin{figure}[!h]
    \centering    
    \includegraphics[width = 0.45\textwidth]{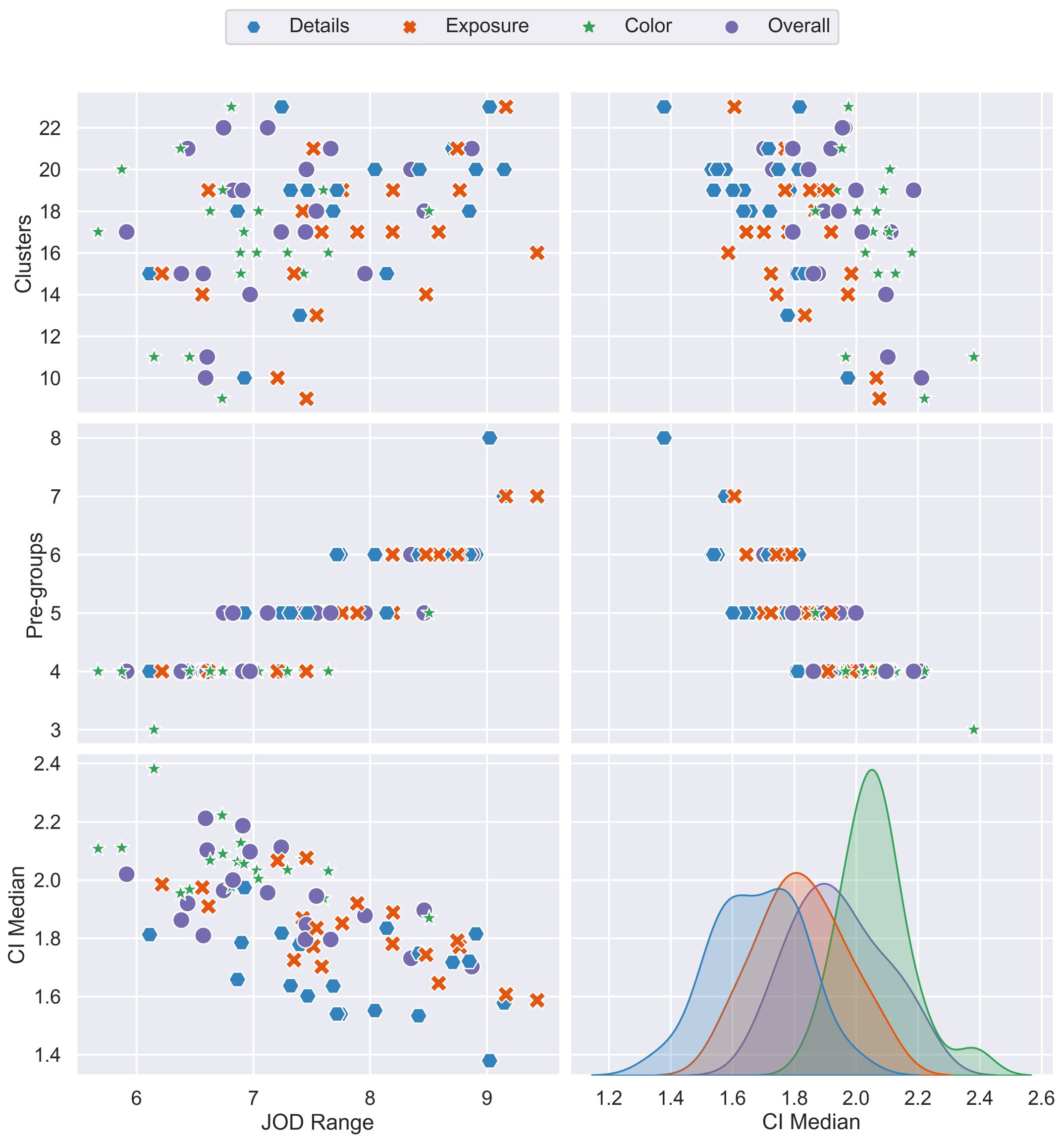}
 \caption{Statistical analysis on 20 scenes for the four attributes. \textit{From top to bottom row (shared x-axis):} distribution of the number of clusters (row 1) and preliminary groups (row 2) in terms of the JOD range (left) and the median CI size (right). Row 3 displays the distribution of the median CI in terms of the JOD range (left), as well as the median CI distribution per attribute (right).}
    \label{fig:StatsGraph}
\end{figure}

 We show the results of our statistical analysis on 20 scenes for the four attributes in \cref{fig:StatsGraph}, from which we can make several interesting observations. First, we note that the number of clusters and groups is correlated with the JOD range and the median CI size (rows 1, 2). A wider JOD interval indicates a wider quality coverage, which implies a higher number of quality levels. Similarly, a smaller confidence interval indicates that the images can be more easily separated, which in turn implies a higher number of quality levels. Second, we observe that the median CI size decreases with the JOD range (row 3, left). This confirms our hypothesis in \cref{subsub:Limitations} that sampling images of close quality makes the task more difficult.
 Finally, we note that detail preservation and exposure have smaller CIs, while color has the largest, indicating a greater difficulty in annotating this attribute (row 3, right), thus justifying its omission.
 
\section{Blind image quality assessment with a tweak}
Based on the PIQ23 dataset, we introduce a deep BIQA method (SEM-HyperIQA) that adapts to the specific structure of the dataset, where each scene has a separate quality scale. We retrain several existing BIQA methods from the literature and compare them to our proposed approach.

\subsection{Semantics aware IQA}
\begin{figure*}[!h]
    \centering    \includegraphics[width = 0.88\textwidth]{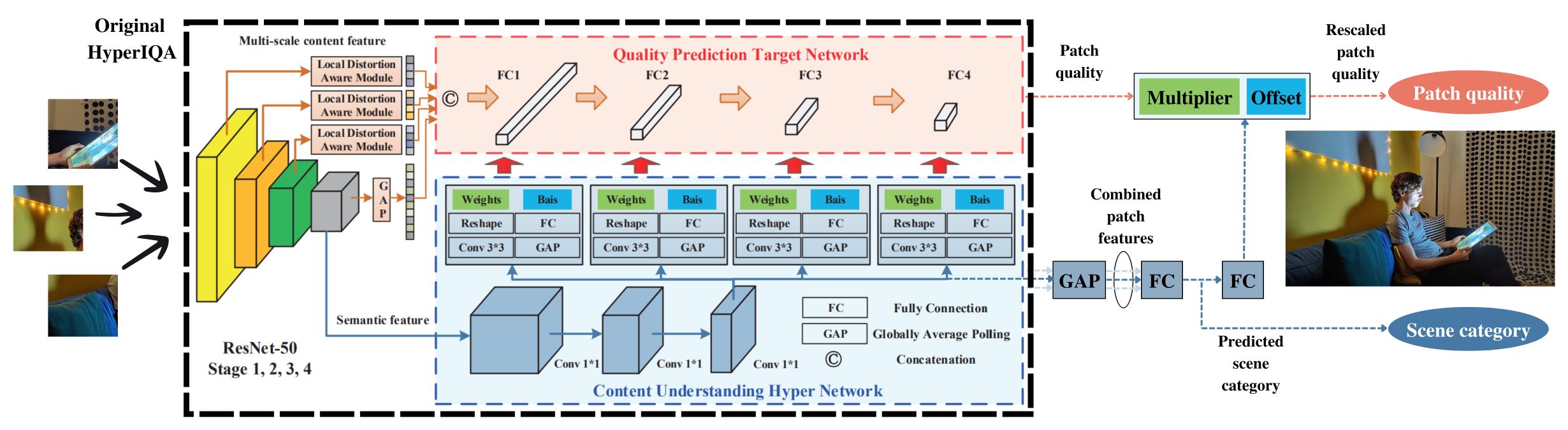}
 \caption{The SEM-HyperIQA architecture. We combine the semantic representation acquired by HyperIQA \cite{su2020blindly} for multiple patches to predict the scene category. We then use the predicted category to rescale the patch quality. The image score is averaged across patches.}
    \label{fig:SEMHyperIQA}
\end{figure*}
The PIQ23 dataset contains individually annotated scenes, each with its own quality scale and unique content. This characteristic introduces a problem known as domain shift (\cref{subsub:domainshift}), involving both content-dependent and annotation-dependent factors. This emphasizes the need to understand the scene's semantics and align the predicted quality with its corresponding scale.
In order to address the challenges of domain shift in PIQ23, we propose SEM-HyperIQA, a solution that involves combining the HyperIQA architecture, which integrates semantic information, with multitasking, which allows scene-specific rescaling. Based on the HyperIQA architecture, we concatenate the semantic features of multiple random crops and feed them to a multi-layer perceptron (MLP) that predicts the scene category for the respective image. We then feed the predicted category to a smaller MLP that predicts a multiplier $a_i$ and offset $b_i$ to adapt the predicted quality score of each patch to its respective scene scale, such as $\hat{q}_i = a_i q_i + b_i$, where $q_i$ is the predicted quality score of patch $i$ (\cref{fig:SEMHyperIQA}). The loss is the sum of the $\ell_1$-norm loss and the cross entropy.

We also propose two other variants, SEM-HyperIQA-SO and SEM-HyperIQA-CO. In the first variant, we omit the scene category prediction and instead feed the scene information directly to the MLP that rescales the predicted score. In the second variant, we omit the rescaling part and only keep the scene prediction. 
The two variants will help us explore the individual importance of scene-specific rescaling and semantic prediction, respectively.

\subsection{Performance evaluation}
\label{sub:baseline}
\paragraph{Training strategy.}
We test different training configurations for all the proposed methods and report the best results. Specifically, we randomly sample 70\% of the images in PIQ23 for training and leave the rest for testing. We randomly crop the images to patches of one of the three following sizes: 672, 448, and 224. 
We use Adam stochastic optimization \cite{kingma2014adam} with different learning rates between $10^{-6}$ and $10^{-4}$. We fix the training for 300 epochs and adopt a learning rate decay factor of 0.05 for every 10 epochs. The final image quality score is computed by averaging the individual patch scores. To evaluate the performance, we compute Spearman’s rank correlation coefficient (SRCC) between the model outputs and the JOD scores. 
Since each scene is annotated separately, we compute the correlation over the scores for each individual scene and evaluate the performance as $ \overline{C} = \frac{1}{s} \sum_{i=1}^{s}C_i
$, where $s$ = number of scenes, $C_i$ = correlation for scene $i$.

\begin{table}
  \centering
  \resizebox{0.96\columnwidth}{!}{
  \begin{tabular}{clllll} 
  
    \toprule
     \# & Method & Details & Exposure  & Overall \\
    \midrule
   1 & BRISQUE \cite{mittal2012no} & 0.323 & 0.307  & 0.192\\ 
   2 & NIQE \cite{mittal2012making} & 0.378 & 0.265 & 0.298\\
   3 & ILNIQE \cite{zhang2015feature} & 0.353 & 0.312 & 0.214\\
   4 & DB-CNN \cite{zhang2018blind} & 0.628 $\pm 0.07$ & 0.635 $\pm 0.06$ & 0.555 $\pm 0.07$ \\ 
   5 & HyperIQA \cite{su2020blindly} & 0.649 $\pm 0.08$ & 0.706 $\pm 0.04$ & {0.611} $\pm 0.06$\\
   6 & MUSIQ \cite{ke2021musiq} & \underline{0.671} $\pm 0.07$ & \textbf{0.725} $\pm 0.04$ & 0.589 $\pm 0.07$ \\
    7 & SEM-HyperIQA & \underline{0.671} $\pm 0.07$ & {0.71} $\pm 0.04$ & \underline{0.621} $\pm 0.06$ \\
    8 & SEM-HyperIQA-SO & \textbf{0.722} $\pm 0.06$ & \underline{0.721} $\pm 0.06$ & \textbf{0.642} $\pm 0.08$ \\
    9 & SEM-HyperIQA-CO & 0.664 $\pm 0.07$ & 0.71 $\pm 0.06$ & 0.621 $\pm 0.07$ \\
    \bottomrule
  \end{tabular}
  }
  \caption{Comparison of the baselines according to their average scene Spearman's rank correlation coefficient with the measured JOD scores and their error margin across the scenes. As shown by the table, the deep learning methods tested perform significantly better than their classical counterparts on PIQ23.}
  \label{tab:benchmarks}
\end{table}
\paragraph{Baseline methods.}
We compare SEM-HyperIQA with existing
BIQA models, including BRISQUE \cite{mittal2012no}, NIQE \cite{mittal2012making}, ILNIQE \cite{zhang2015feature}, 
DB-CNN \cite{zhang2018blind}, HyperIQA \cite{su2020blindly} and MUSIQ \cite{ke2021musiq}. We train these models on PIQ23 using their official implementations. 
NIQE and ILNIQE do not require any training. DB-CNN and MUSIQ are pre-trained on LIVE Challenge and PaQ-2-PiQ, respectively. HyperIQA is pre-trained on ImageNet. Results are shown in \Cref{tab:benchmarks}.

\paragraph{Discussion.}
From \Cref{tab:benchmarks} we can make the following observations. First, the deep learning methods tested (4-9) perform better than their classical counterparts (1-3), indicating a difficulty to adapt to high-resolution images, scene-specific scales, and attribute-specific annotations. Zhu \etal \cite{zhu2020multiple} have demonstrated the ineffectiveness of such methods when the annotations do not represent an overall subjective evaluation of the quality,  
Second, the proposed SEM-HyperIQA method improves upon the original HyperIQA, which indicates the effectiveness of scene semantics and multitasking in quality prediction, especially for separate scene scales. Third, SEM-HyperIQA-SO with scene-specific rescaling achieves the best performance. It notably enhances the detail preservation attribute, possibly due to the limited information available in face crops for scene analysis. Therefore, semantic information cannot be fully utilized and we are better off using scene-specific rescaling only. Fourth, we note that deep BIQA models perform significantly better for detail preservation and exposure than overall, which directly reflects this task's difficulty and the uncertainty of the annotations, as discussed in \cref{parag:discussonStatsAna}. 

\section{Conclusion}
We have presented PIQ23, a new dataset for portrait quality assessment with a wide variety of smartphone cameras and use cases, which has been annotated by image quality experts using pairwise comparisons. We have shown the importance of identifying the uncertainty in the annotations by providing a new statistical analysis method to cluster the quality scale into consistent levels of quality. Finally, we adopt a training strategy and a deep neural network architecture that adapts to the high-resolution images of PIQ23 and profits from semantic information and multitasking, in order to adjust to the per-scene quality scaling of the dataset. Our results have shown the necessity and effectiveness of quality scale quantification and clustering of similar quality images to contain annotation uncertainty, as well as the importance of semantic information in training IQA models. We believe that this work will be the foundation for a new area of application of IQA for portrait images, as well as for a higher caliber of annotations in IQA datasets.

\section*{Acknowledgments}
This work was funded in part by the French government under the management of Agence Nationale de la Recherche as part of the “Investissements d’avenir” program, reference ANR-19-P3IA-0001 (PRAIRIE 3IA Institute), the Louis Vuitton/ENS chair in artificial intelligence and the Inria/NYU collaboration. NC was supported in part by a DXOMARK/PRAIRIE CIFRE Fellowship. We thank DXOMARK engineers and photographs for their time investment and fruitful discussions about image quality.
{\small
\bibliographystyle{ieee_fullname}
\bibliography{egbib}
}

\end{document}


\maketitle
\section{Portrait quality attributes}
We ask more than 30 experts to annotate PIQ23 following carefully designed guidelines, on four attributes: face detail preservation, face target exposure, global color
quality, and overall image quality. In this section, we elaborate on the guidelines for each attribute and give the remarks that were taken into consideration when developing the dataset.
\subsection*{Face detail preservation} 
We study the fidelity of face rendering in terms of detail preservation and skin smoothness. We tend to appreciate natural facial renderings over unnatural ones. For example, some over-sharpened faces can be considered worse than slightly blurred faces if the latter has a more natural and smooth texture rendering. Additionally, we have found that it is necessary to explicitly define the priorities in penalizing noise when the texture quality is similar. Hence, high-frequency noise is preferable to low-frequency noise, true random noise is better than patterned noise, and luminance noise is better than chromatic noise. Finally, a more general comparison of the skin texture and the facial details were considered, in particular beard, eyebrows, hair, etc.

\subsection*{Face target exposure} 
This attribute is used to evaluate the quality of the light rendering on the face. We have asked to find the balance between target exposure, contrast, and dynamic range on the face. This evaluation is sometimes hard since some of the attributes can be contradictory and finding a sweet spot between all the criteria is not always straightforward. In case of ambiguous comparisons (when it is not clear which image is better), we have left the choice to the observer and his preferences.

\subsection*{Global color quality \footnote[1]{This attribute was omitted from the final version of the dataset because of its annotation challenges.}} 
Using the full image, we have asked to find a trade-off between the overall white balance and color rendering of the portrait image. The focus is usually on the subject since it constitutes the main element in the image, but we have not forced more specific guidelines for this attribute. 
We have also noted the following points:
\begin{itemize}
    \item \textbf{Note on heavily under-exposed images:} In the case of a heavily under-exposed image, color was penalized, since we cannot really see any color.
    \item \textbf{Note on HDR scenes:} In case there was a failure in the dynamic range of the image, color also was penalized.
    \item \textbf{Note on skin tones:} Evaluating the skin tone rendering was not explicitly taken into consideration in the color attribute, since we need a reference image per person to evaluate it correctly. Defining good skin tone rendering is complicated and prone to subjectivity. The quality of skin tone rendering is hard to judge objectively without having a previous idea about the skin tone of the person. Also, skin reflectance (how much light the skin reflects) should be taken into consideration when analyzing skin tone renderings and when evaluating the target exposure on the face. Another type of analysis should be taken into consideration, such as comparing the quality of the skin tone rendering to a reference image of the person in multiple controlled lighting conditions.
\end{itemize}

\subsection*{Overall image quality}    
The overall attribute is considered a trade-off of all the main attributes in a portrait image. We have evaluated the overall quality of the image while prioritizing some aspects over others. We have defined a list of prioritized attributes that were only taken into consideration when the quality difference is ambiguous. Also, important failures were directly penalized. We have left the choice to the observer to estimate the severity of the quality difference in case multiple aspects came into play. We have tried to estimate the role of each attribute on three levels: Essential, important, and subjective. Essential means the attribute always plays a role in the overall quality of the image. Important represents the attributes that occasionally play a role in the overall quality. Finally, subjective attributes rarely play a role in the overall quality and their impact varies accordingly to the observer's opinion. Let us now list the attributes judged to play one of the three aforementioned roles in the overall quality analysis: 

    \begin{itemize}
        \item \textbf{Face target exposure (essential):} we usually prioritize naturally well-exposed faces over underexposed or overexposed ones.
        \item \textbf{Dynamic range, contrast and global target exposure (essential):} Some essential qualities of an image are how well it preserves the details in dark and bright areas, as well as the general exposure and contrast of the scene. Therefore, the dynamic range of the image, the target exposure, and the contrast are three of the first things to look at in the image. This is mostly interesting in portraits since maintaining a high dynamic of the subject and background is not straightforward.
        \item \textbf{White balance, color rendering, and skin tone (essential):} The quality of the color rendering between the portrait and the background can play a big role in image quality. A bit similar to the separate attribute, this one compromises the overall quality of a portrait image.
        \item \textbf{Sharpness and focus (important):} These attributes can be analyzed on single and group photos, with single and multi-planes. We try to analyze the capacity of the camera to focus on the main subject and the ability to separate planes.
        \item \textbf{Artifacts (subjective):} This attribute was be analyzed in case of an important artifact failure or when two images have very close quality levels. Several artifacts can play a subjective role in the overall quality of a portrait. We note some of these such as ghosting, halos, hue shift, ringing, flare, etc.
    \end{itemize}

\section{Examples of attribute quality}

\subsection*{Face detail preservation}
\SetTblrInner{rowsep=0pt, colsep=2pt}
\begin{figure}[H]
\centering
\begin{tblr}{cccccc}
    \hline[0.5pt,black]
    Level 0 & Level 1 & Level 2 & Level 3 & Level 4 & Level 5 \\
    \hline[0.5pt,black] \\
    \includegraphics[width = 0.15 \textwidth]{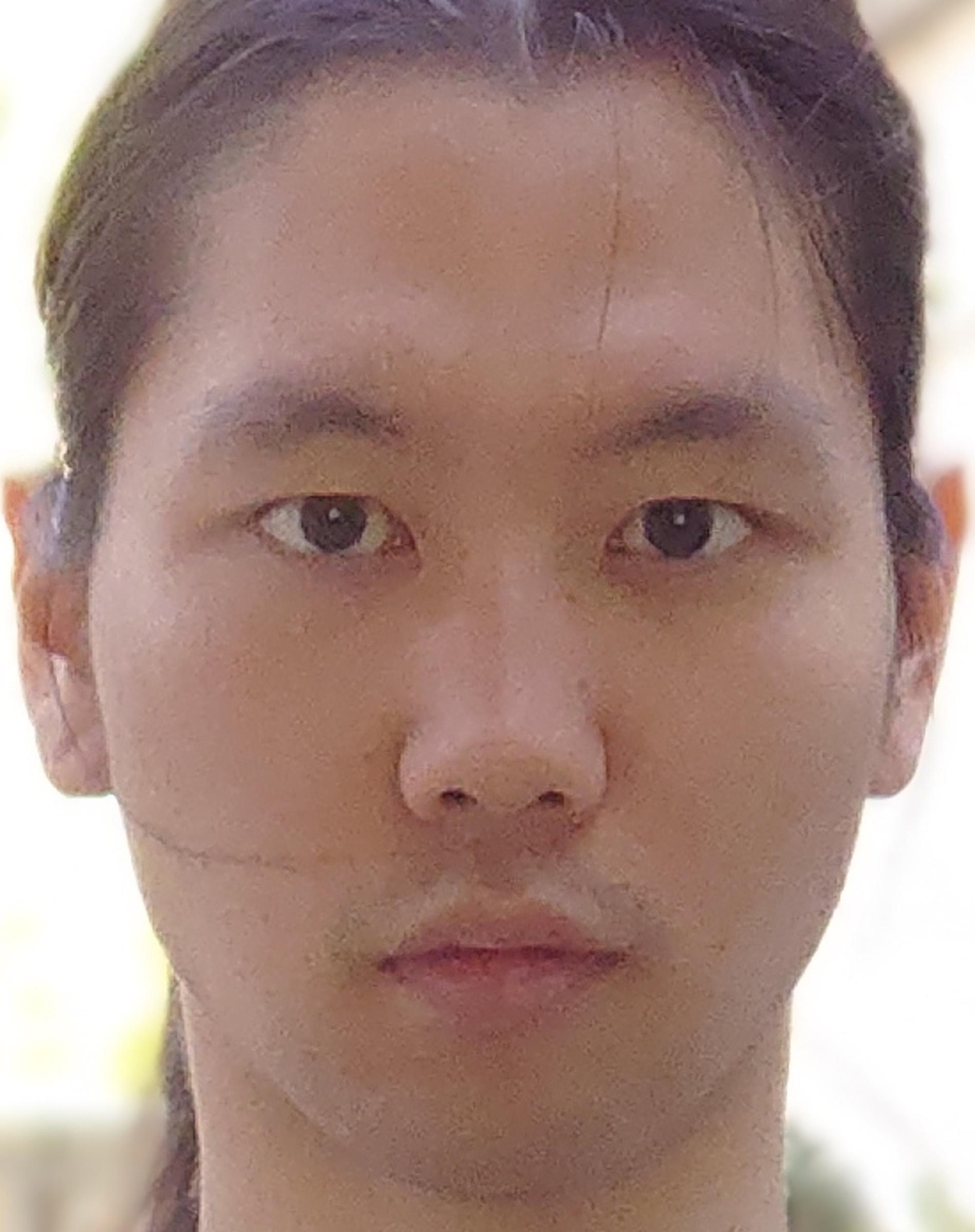} & 
    \includegraphics[width = 0.15 \textwidth]{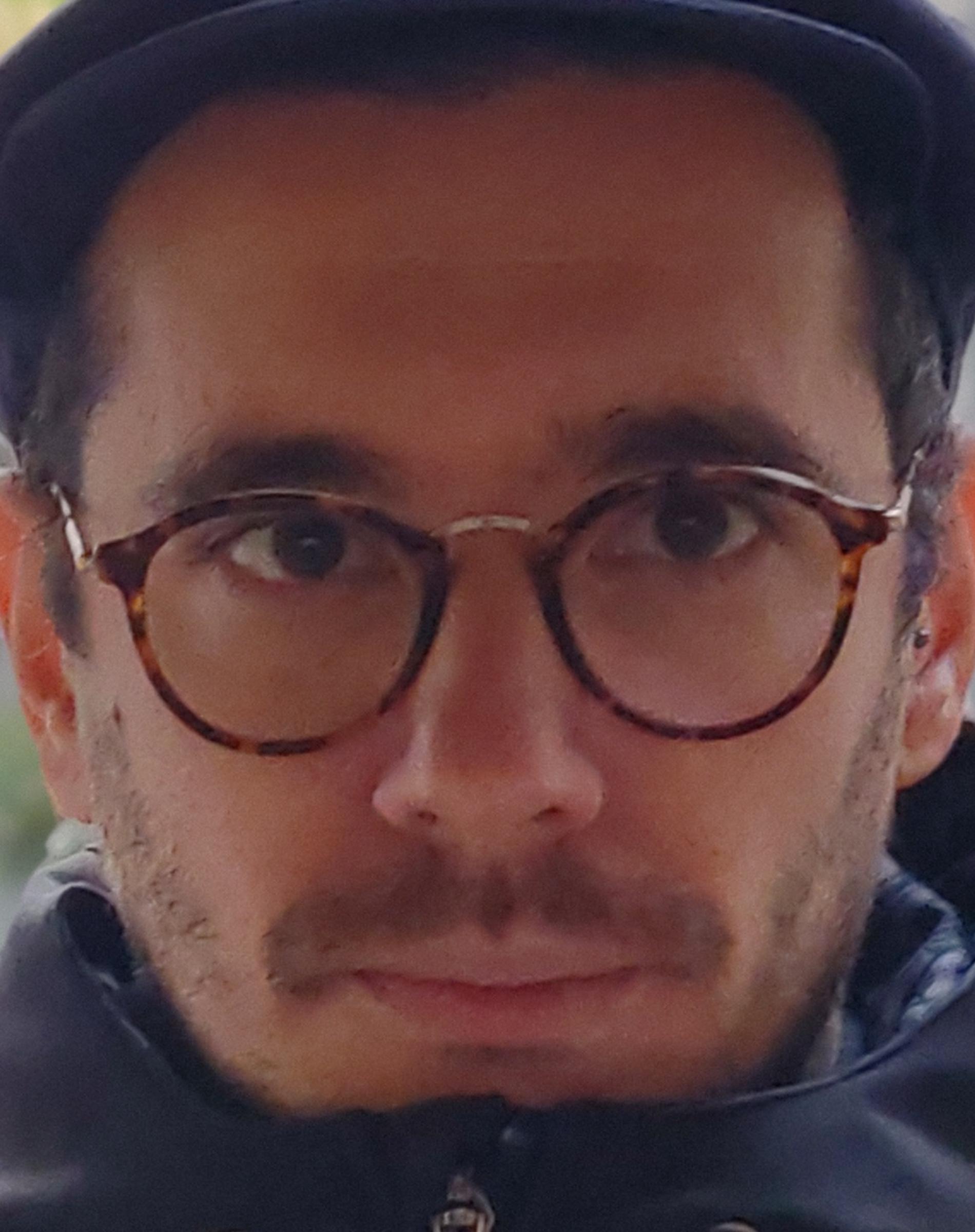} & 
    \includegraphics[width = 0.15 \textwidth]{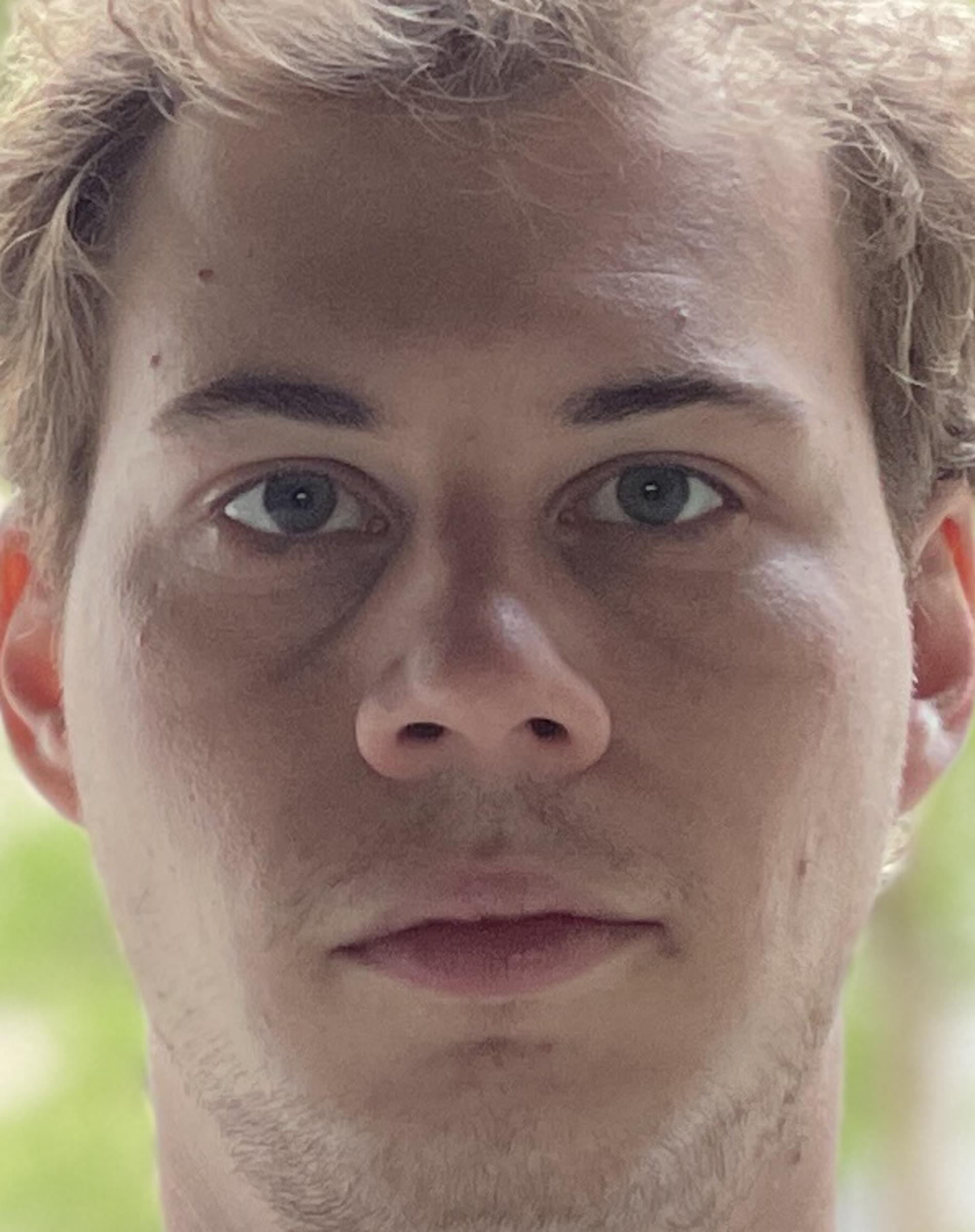} & 
    \includegraphics[width = 0.15 \textwidth]{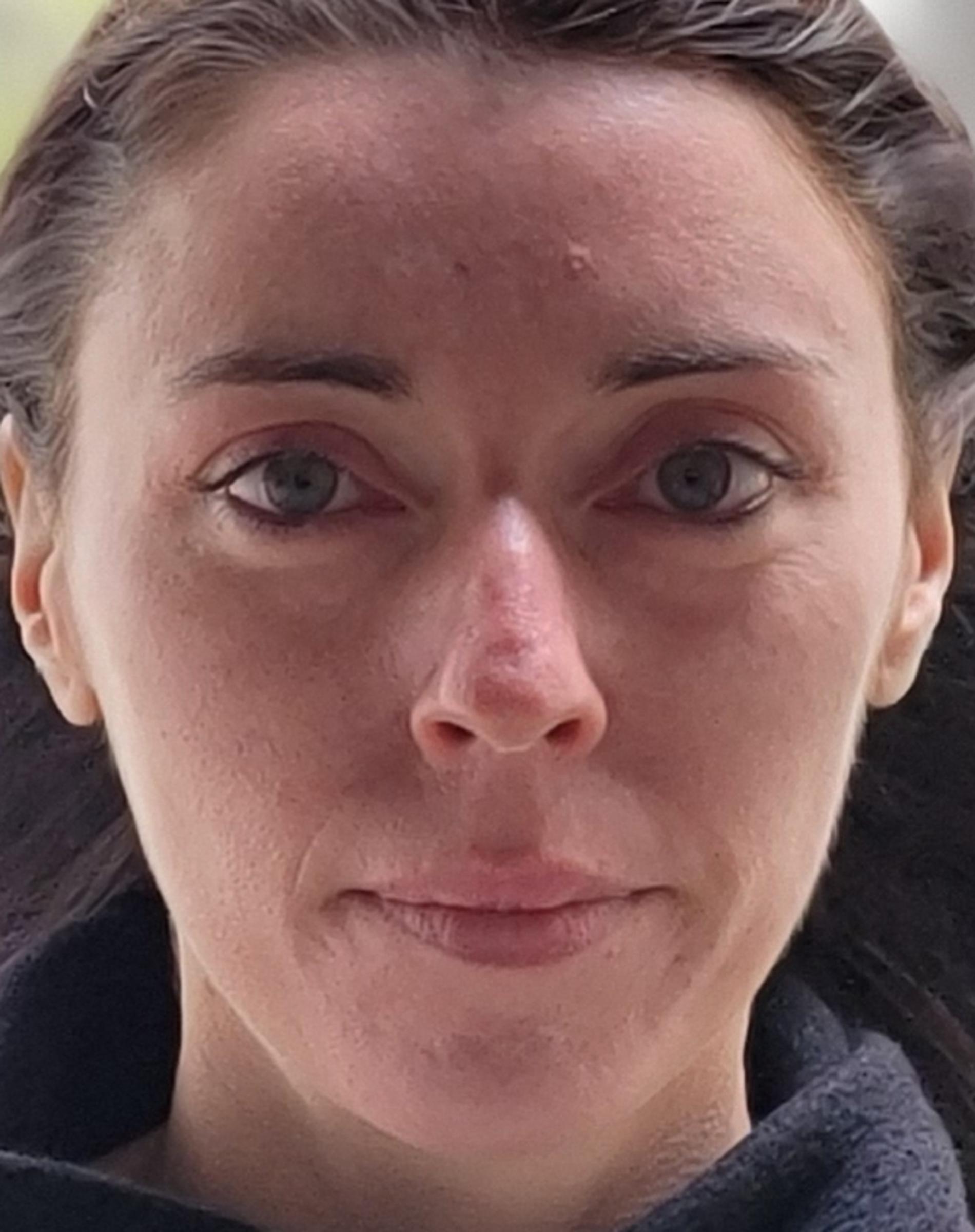} &
    \includegraphics[width = 0.15 \textwidth]{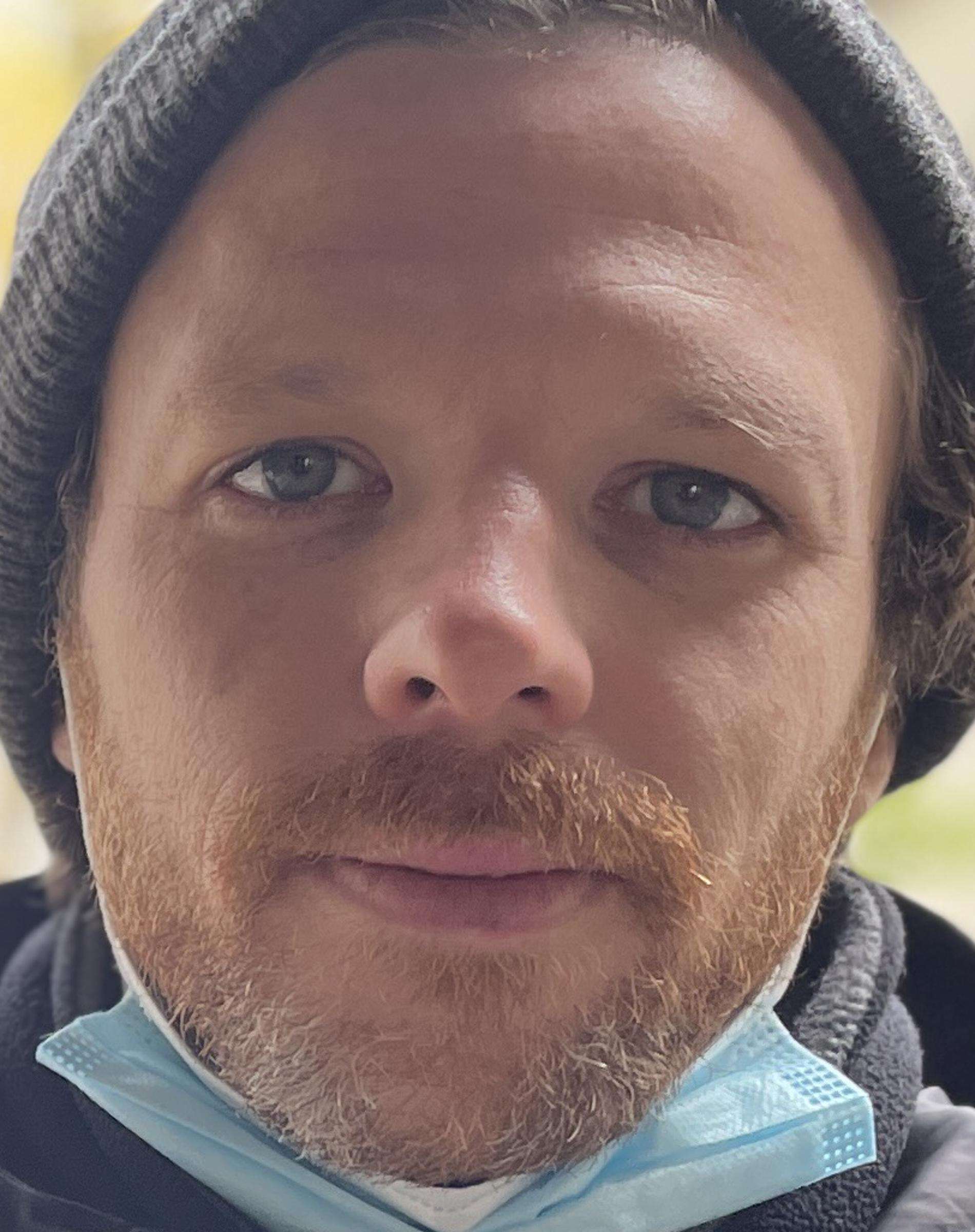} &
    \includegraphics[width = 0.15 \textwidth]{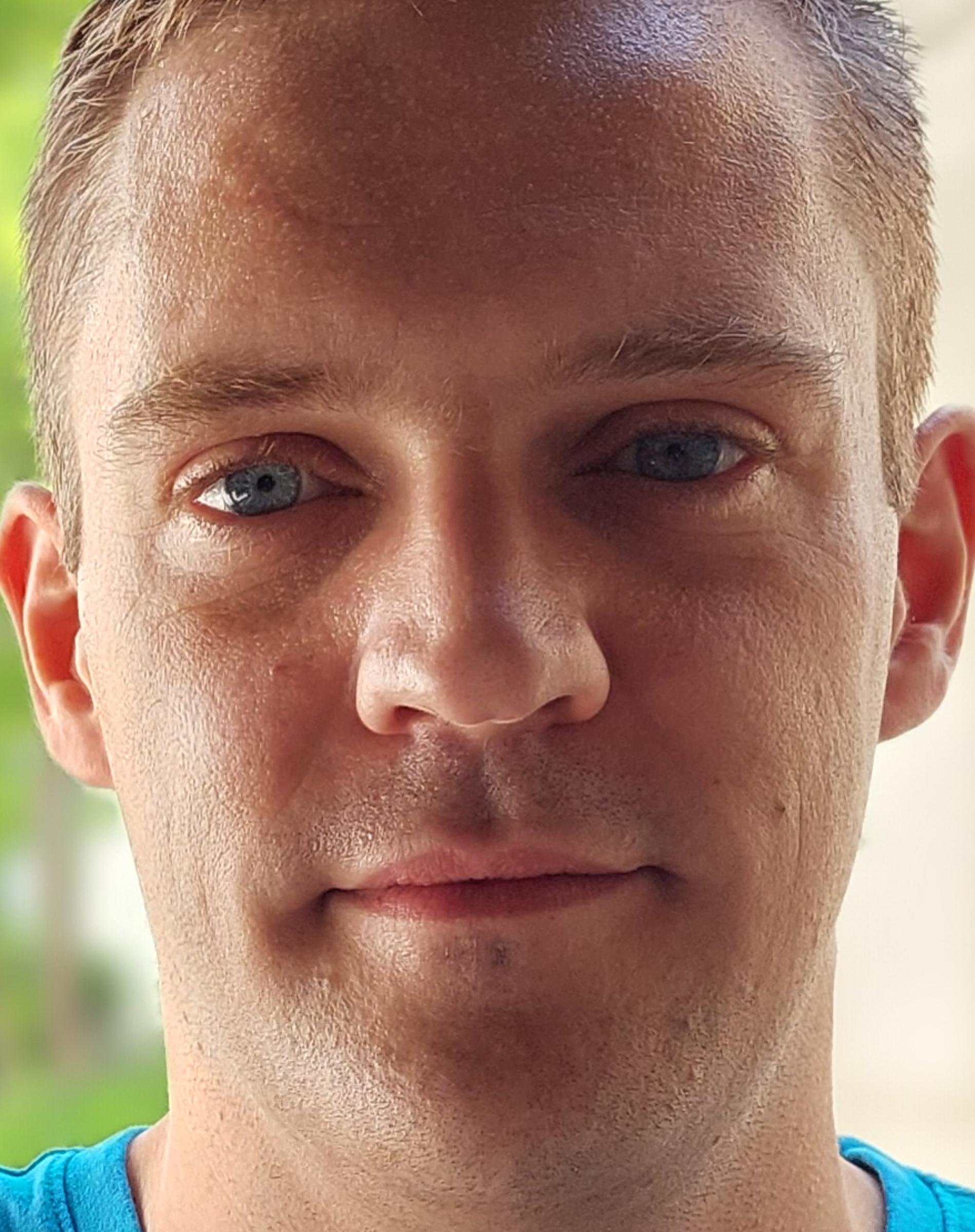}
    \\
    \includegraphics[width = 0.15 \textwidth]{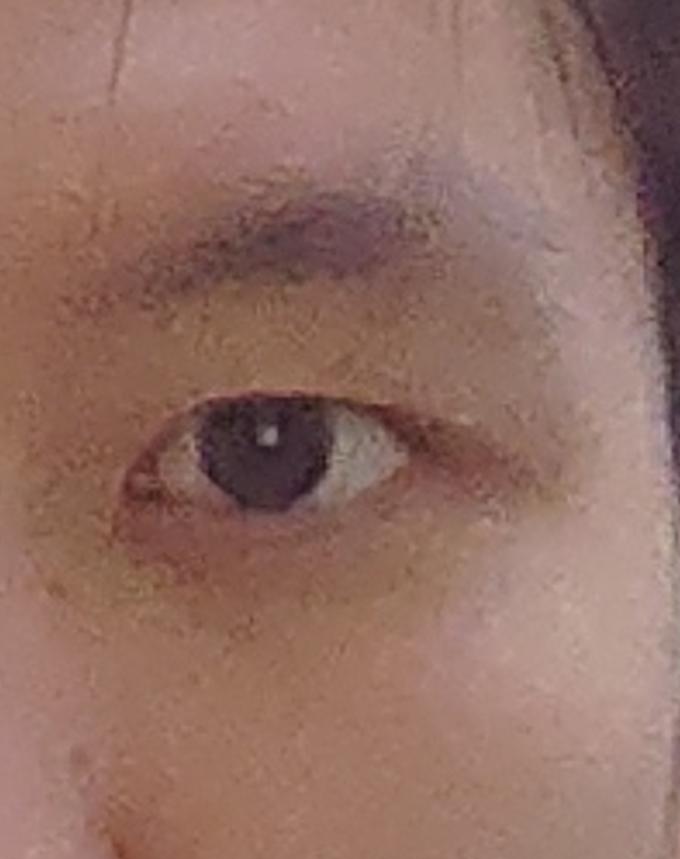} & 
    \includegraphics[width = 0.15 \textwidth]{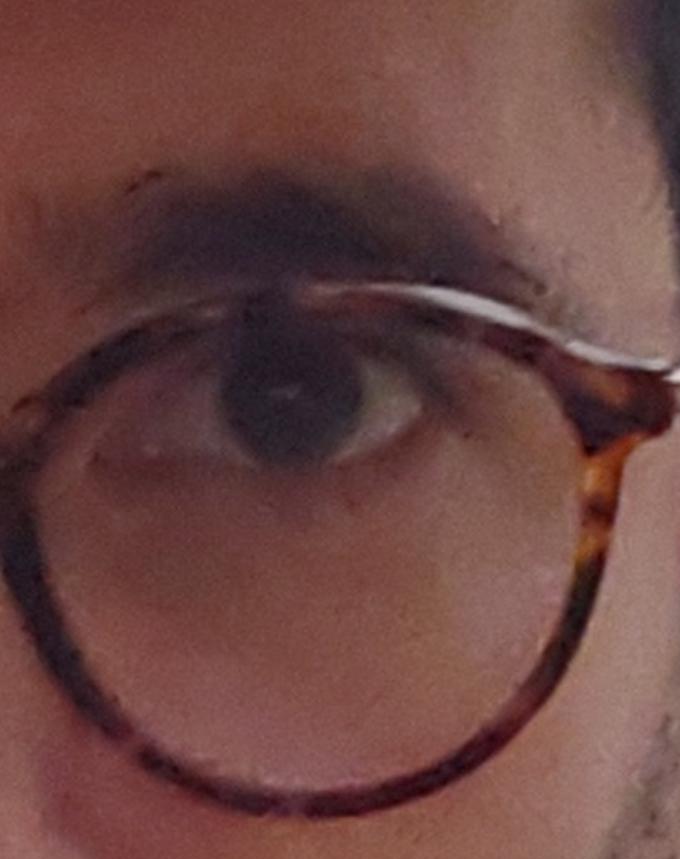} & 
    \includegraphics[width = 0.15 \textwidth]{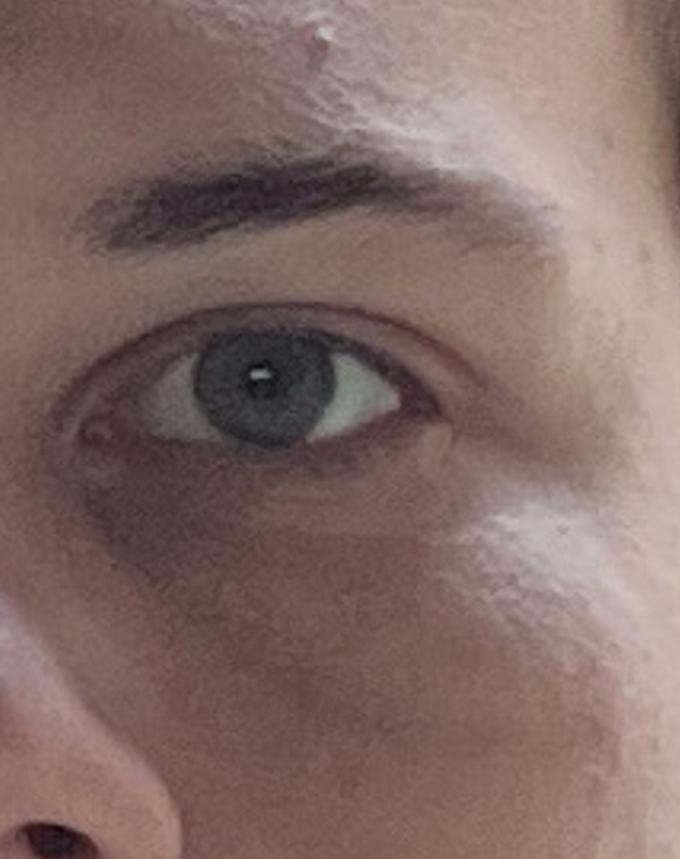} & 
    \includegraphics[width = 0.15 \textwidth]{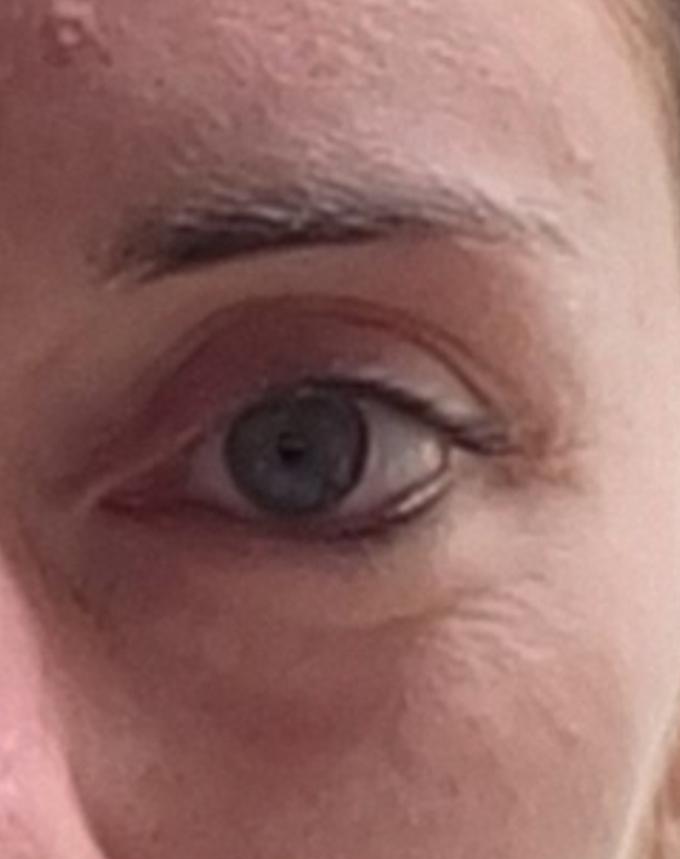} &
    \includegraphics[width = 0.15 \textwidth]{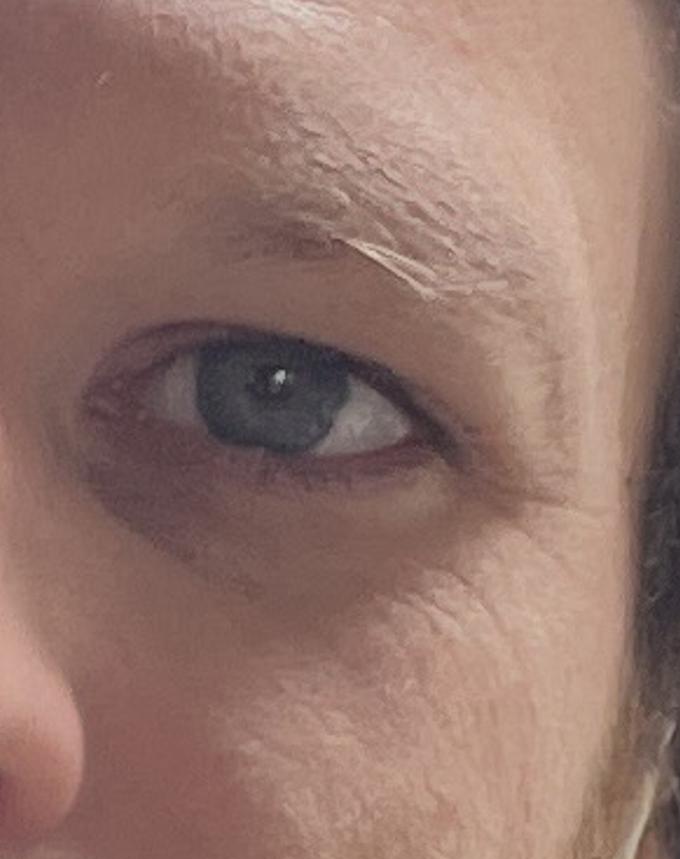} &
    \includegraphics[width = 0.15 \textwidth]{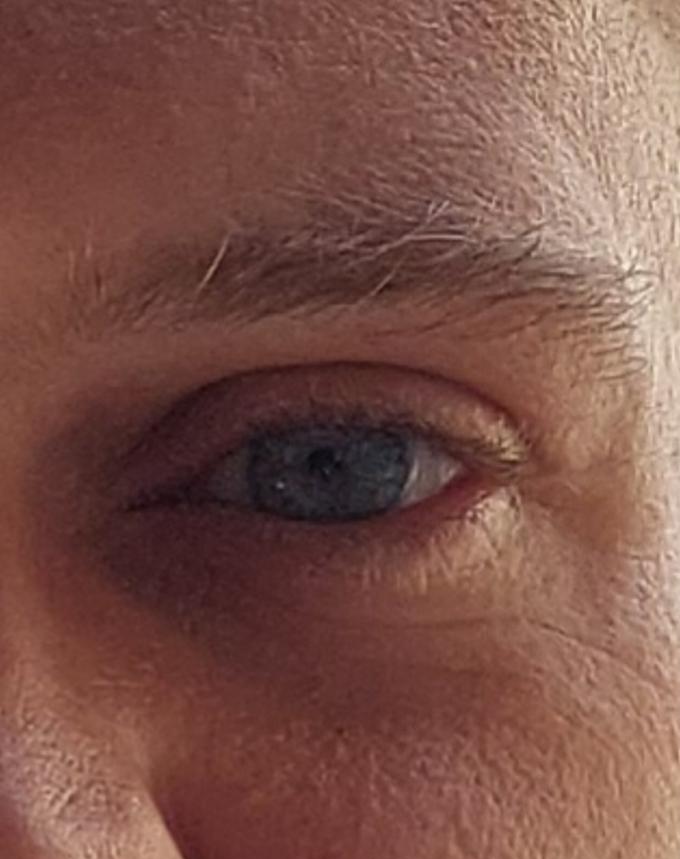}\\
    \hline[0.5pt,black]
    
    -2.8 & -0.81 & -0.1 & 0.75 & 1.57 & 4.14
    \\
     \hline[0.5pt,black]
\end{tblr}
\caption{Examples of different levels of \textbf{face detail preservation quality} and their corresponding \textit{JOD} scores illustrated on the face regions and on crops around the eyes.}
\end{figure}

\subsection*{Face target exposure}
\SetTblrInner{rowsep=0pt, colsep=2pt}
\begin{figure}[H]
\centering

\begin{tblr}{cccccc}
    \hline[0.5pt,black]
    Level 0 & Level 1 & Level 2 & Level 3 & Level 4 & Level 5 \\
    \hline[0.5pt,black] \\
    \includegraphics[width = 0.15 \textwidth]{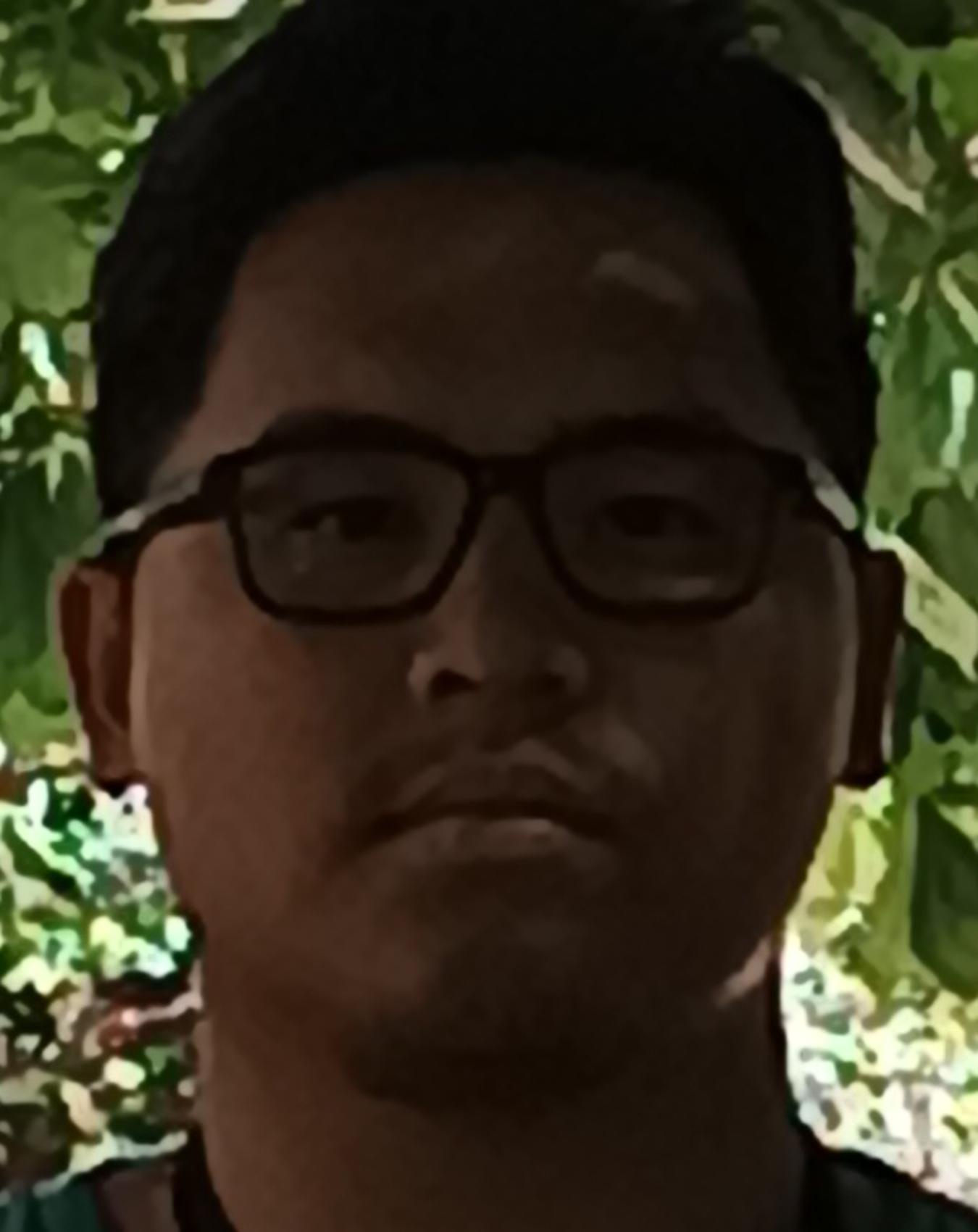} & 
    \includegraphics[width = 0.15 \textwidth]{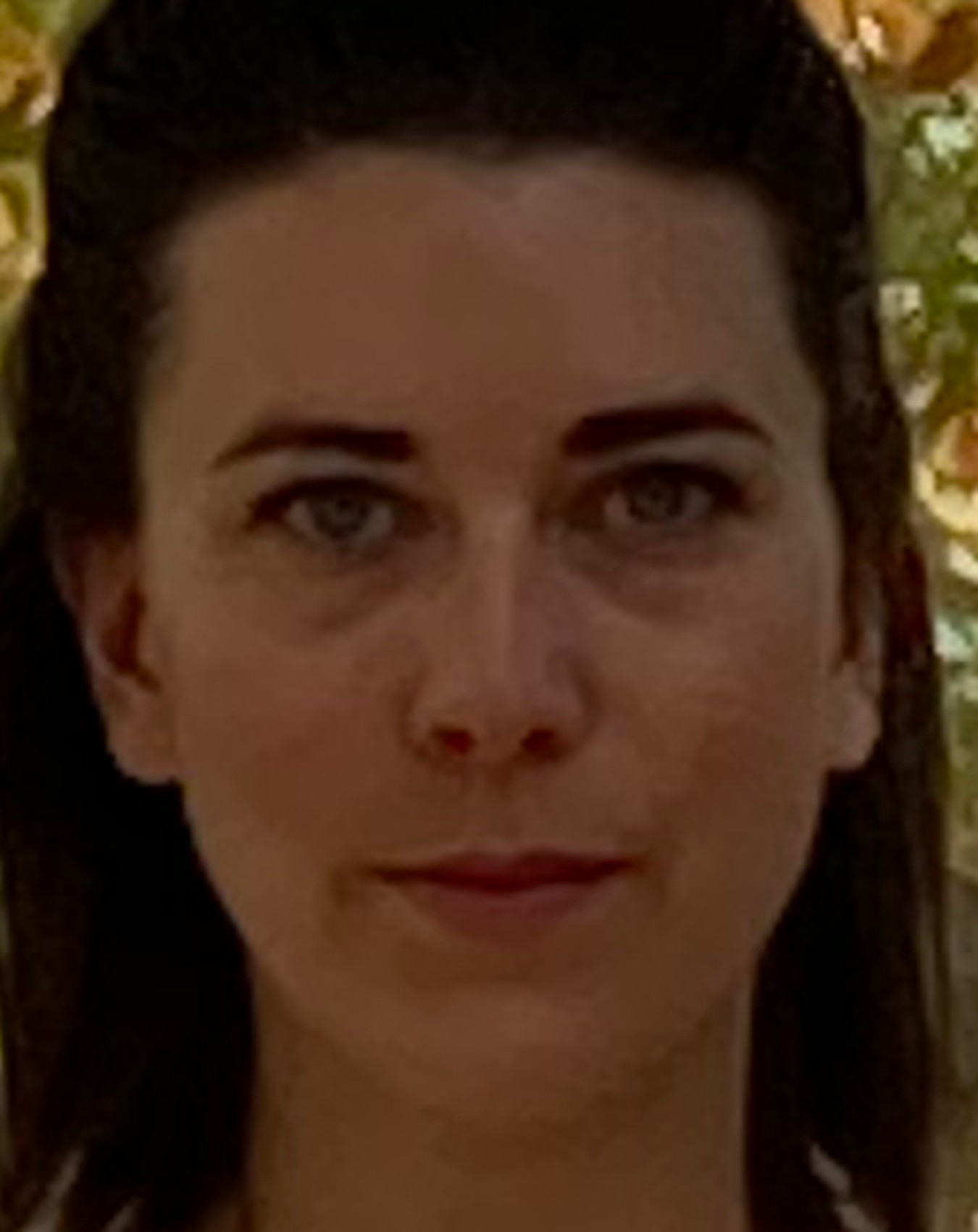} & 
    \includegraphics[width = 0.15 \textwidth]{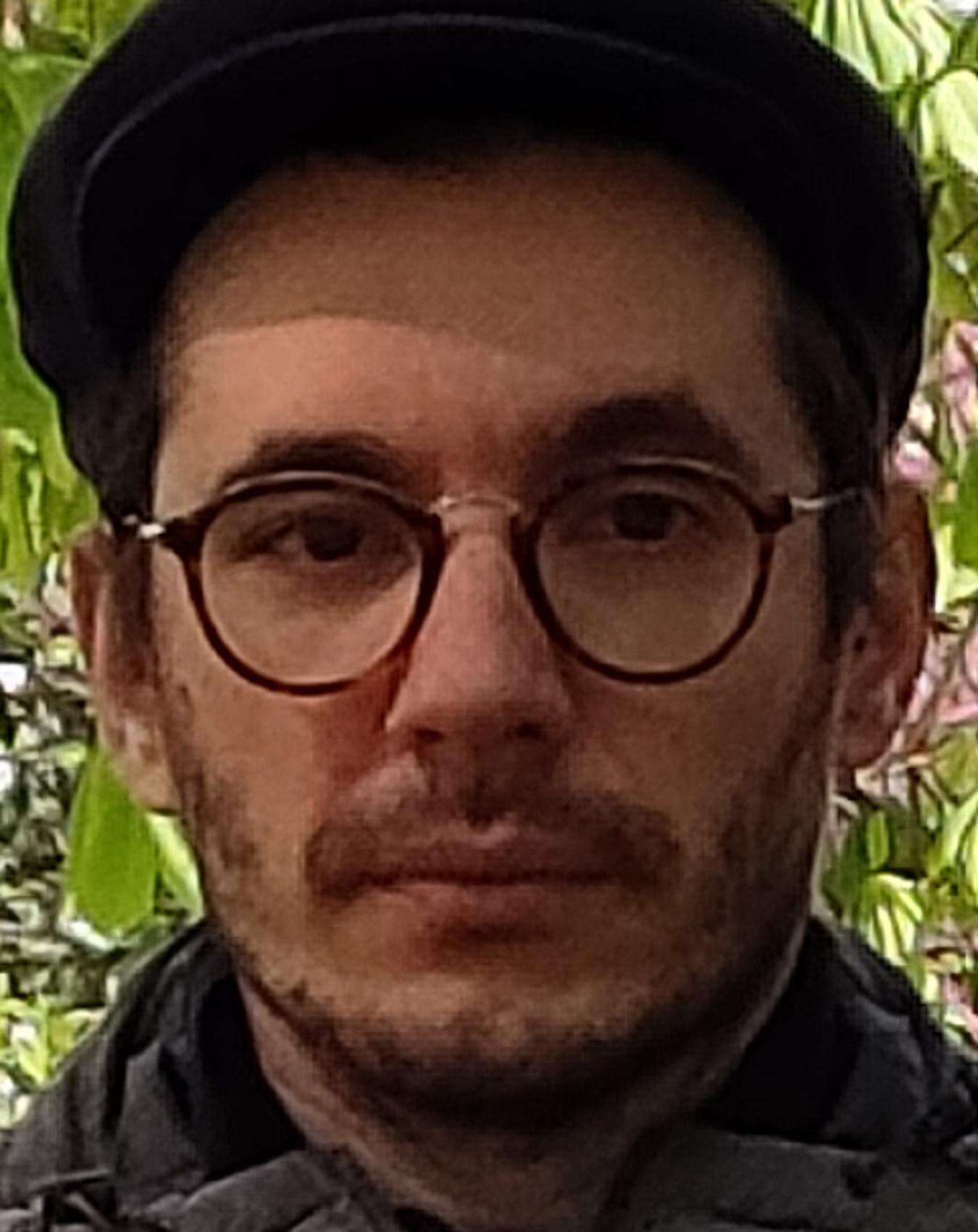} & 
    \includegraphics[width = 0.15 \textwidth]{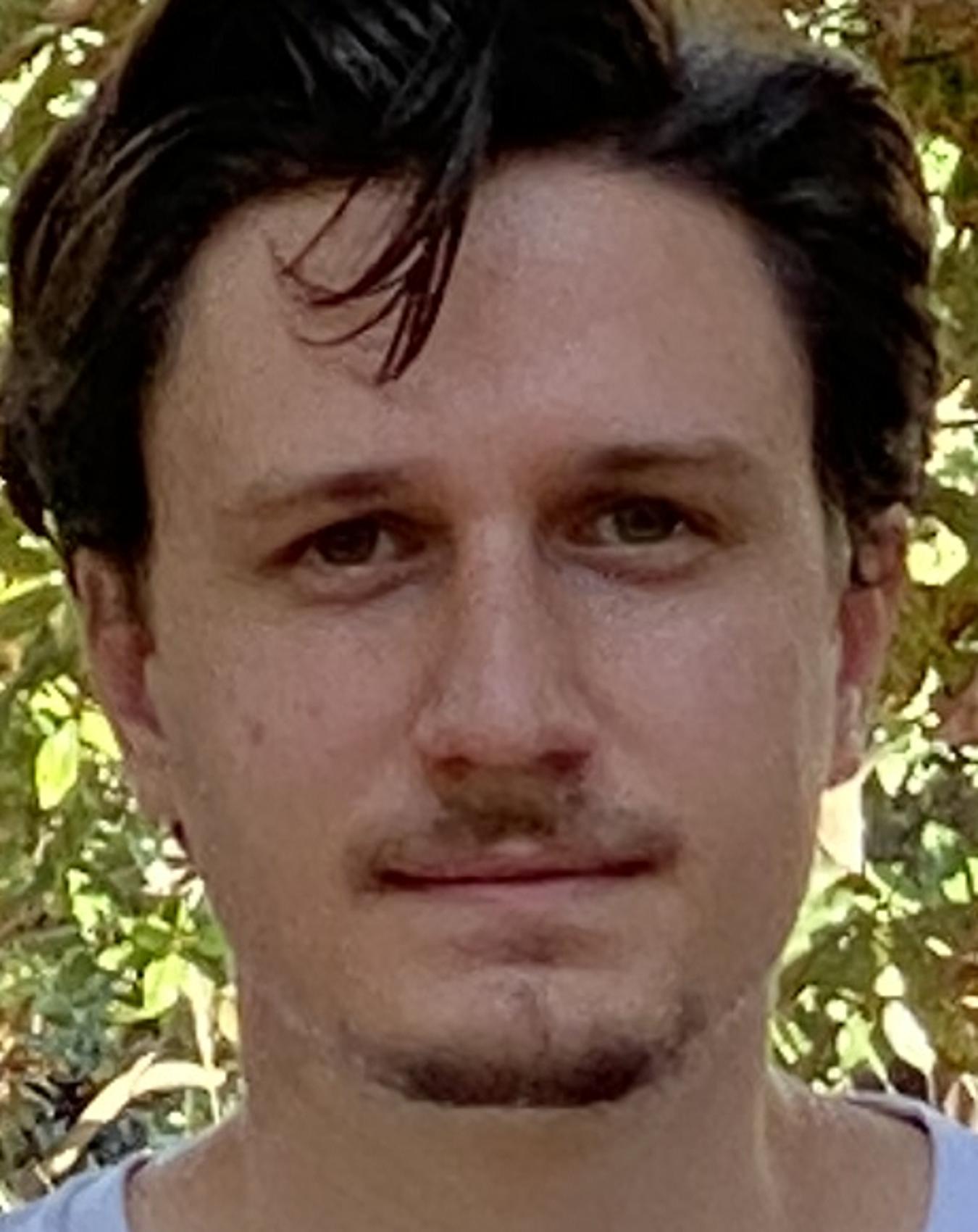} &
    \includegraphics[width = 0.15 \textwidth]{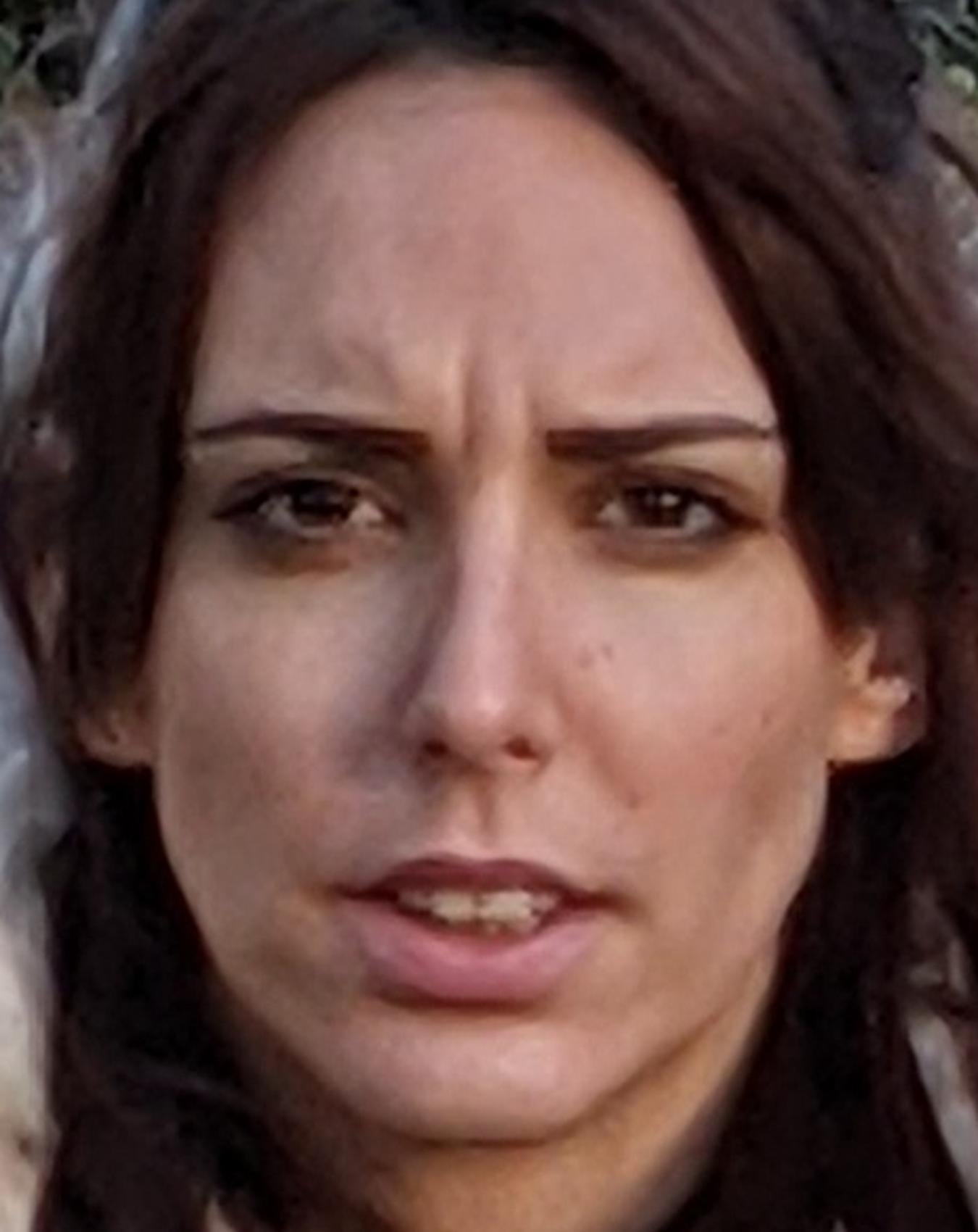} &
    \includegraphics[width = 0.15 \textwidth]{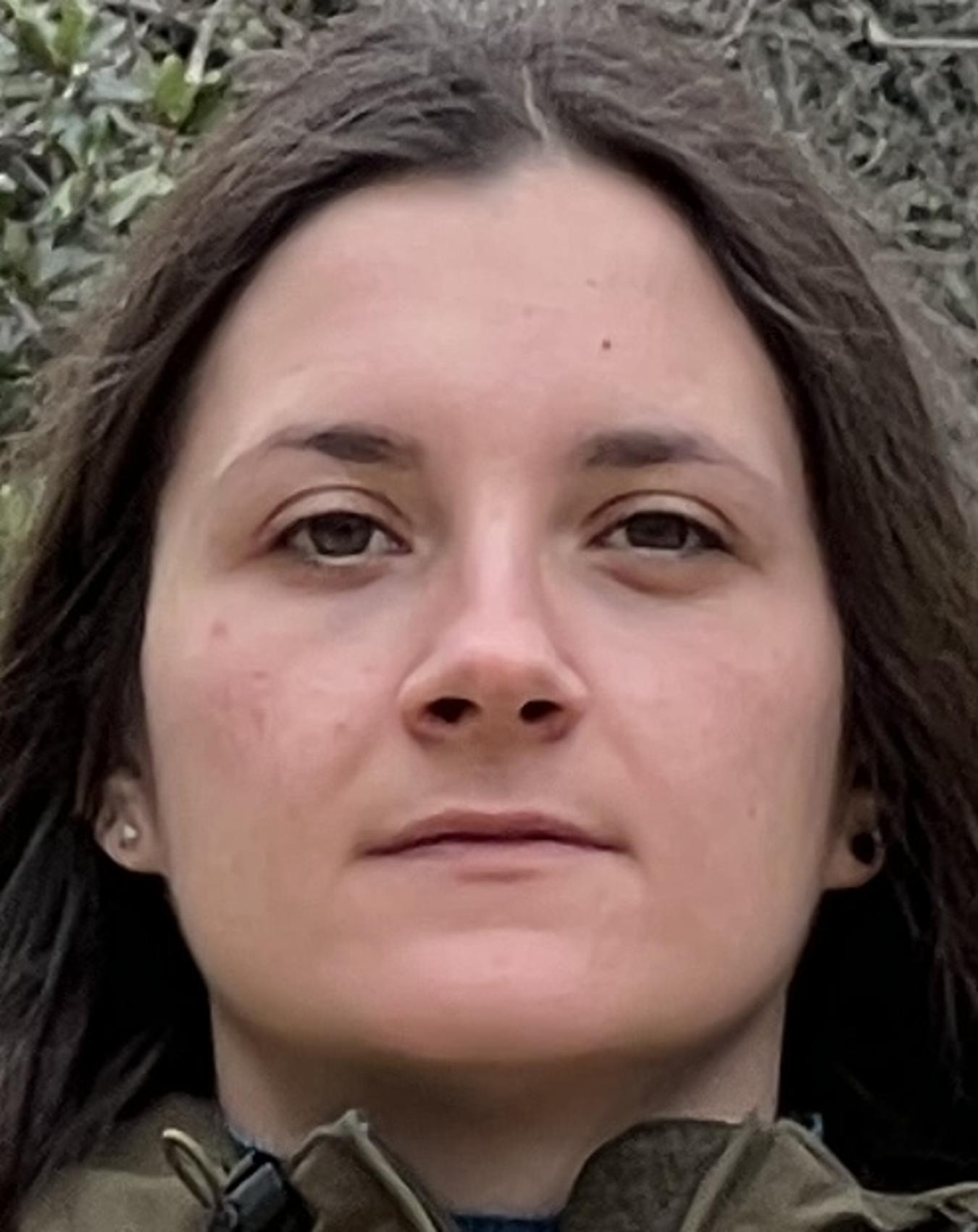}
    \\
    \hline[0.5pt,black]
    -4.91 & -2.21 & -1.39 & 0.02 & 0.9 & 1.96
    \\
     \hline[0.5pt,black]
\end{tblr}
\caption{Examples of different levels of \textbf{face target exposure quality} and their corresponding \textit{JOD} scores.}
\end{figure}

\subsection*{Global color quality}
\SetTblrInner{rowsep=0pt, colsep=2pt}
\begin{figure}[H]
\centering
\begin{tblr}{cccc}
    \hline[0.5pt,black]
    Level 0 & Level 1 & Level 2 & Level 3 \\
    \hline[0.5pt,black] \\
    \includegraphics[width = 0.2 \textwidth]{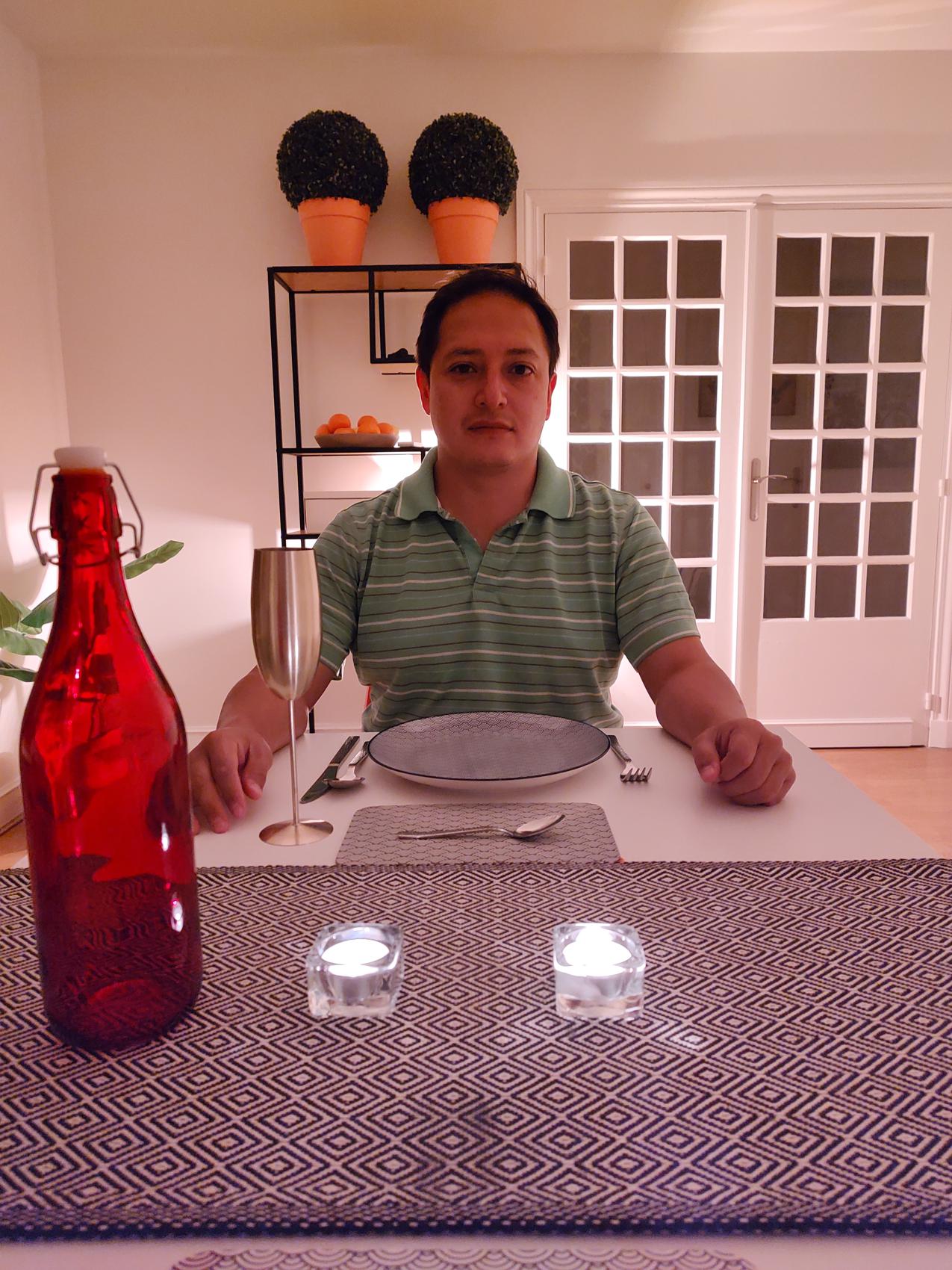} & 
    \includegraphics[width = 0.2 \textwidth]{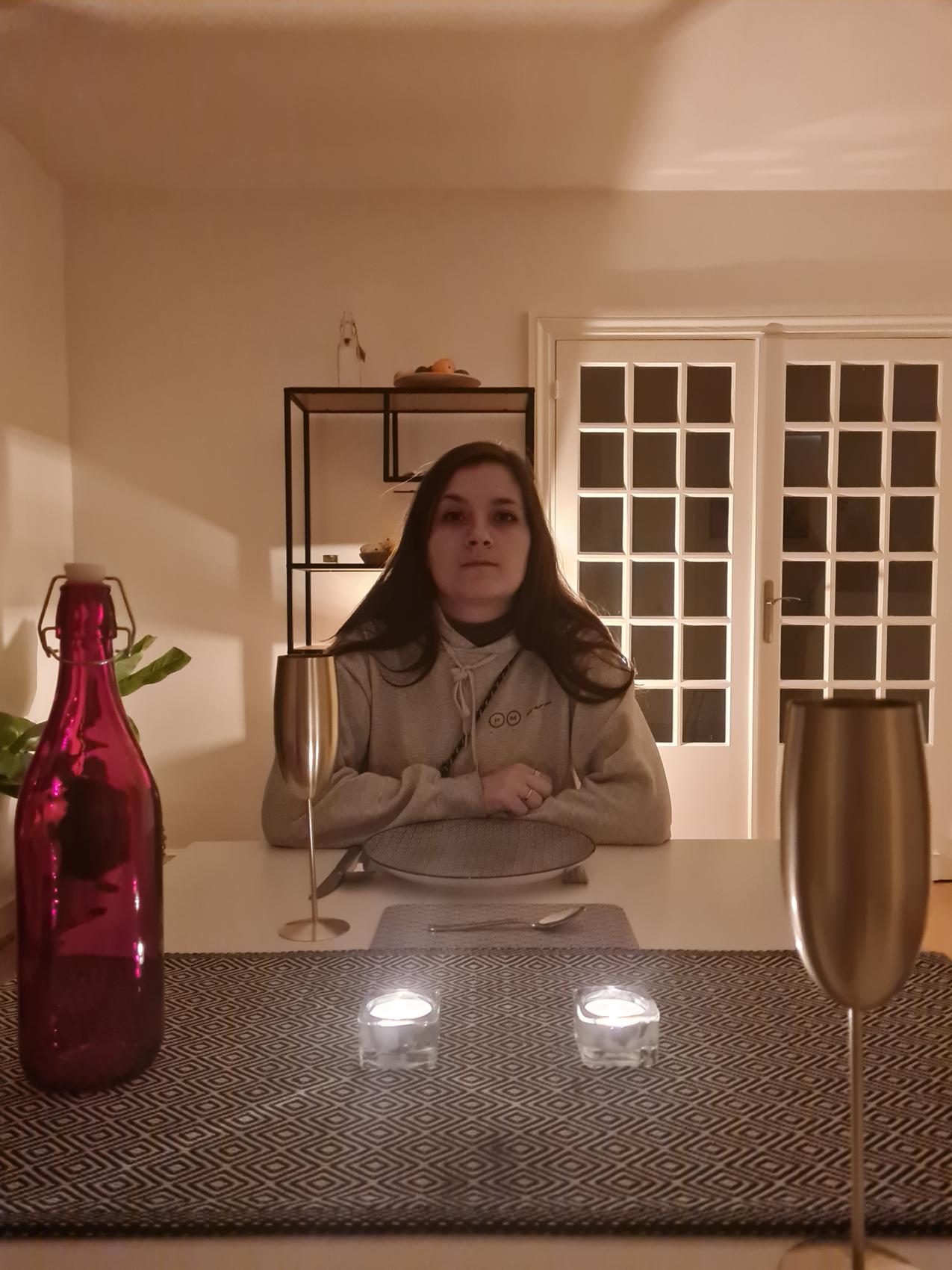} & 
    \includegraphics[width = 0.2 \textwidth]{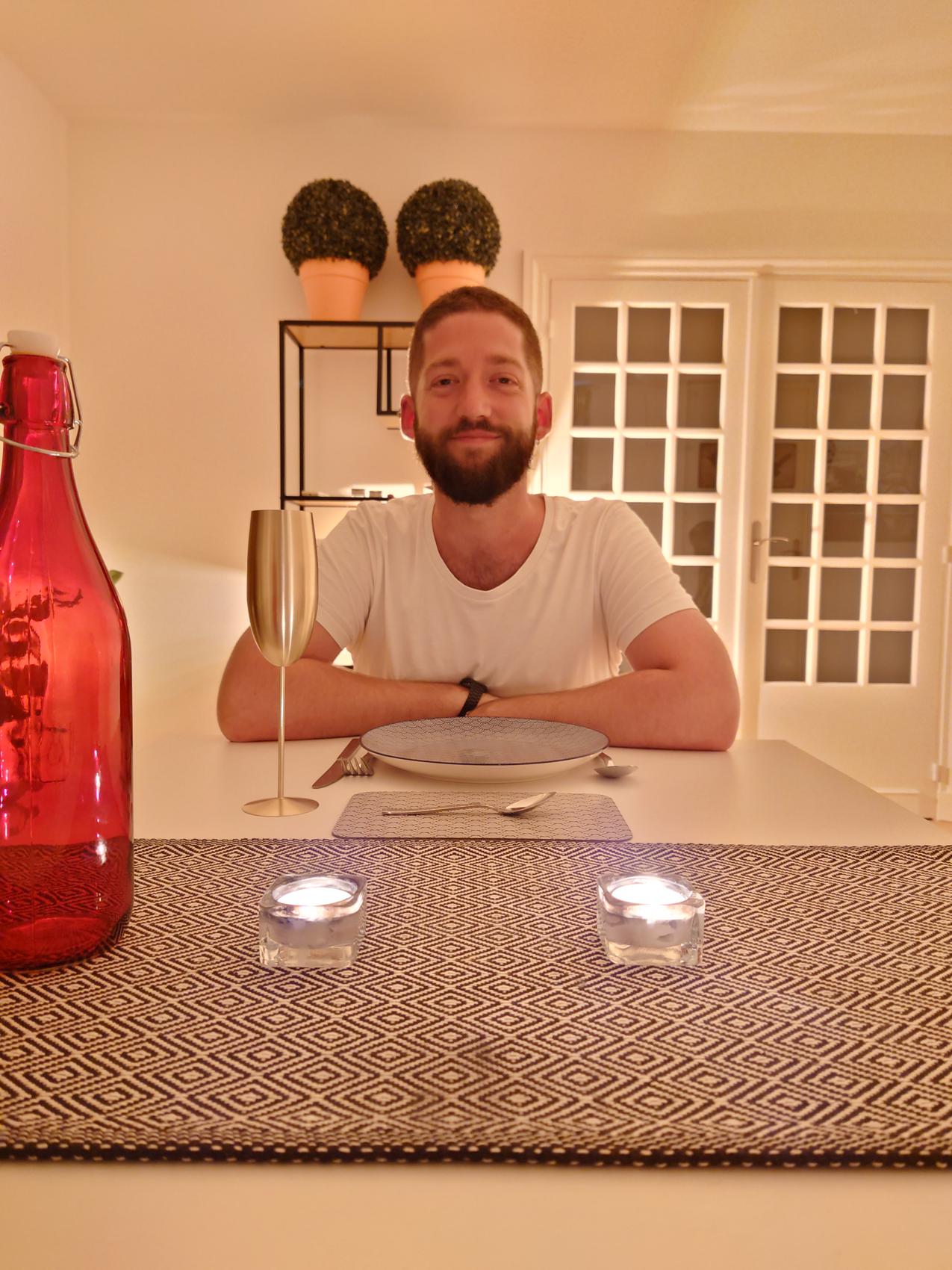} & 
    \includegraphics[width = 0.2 \textwidth]{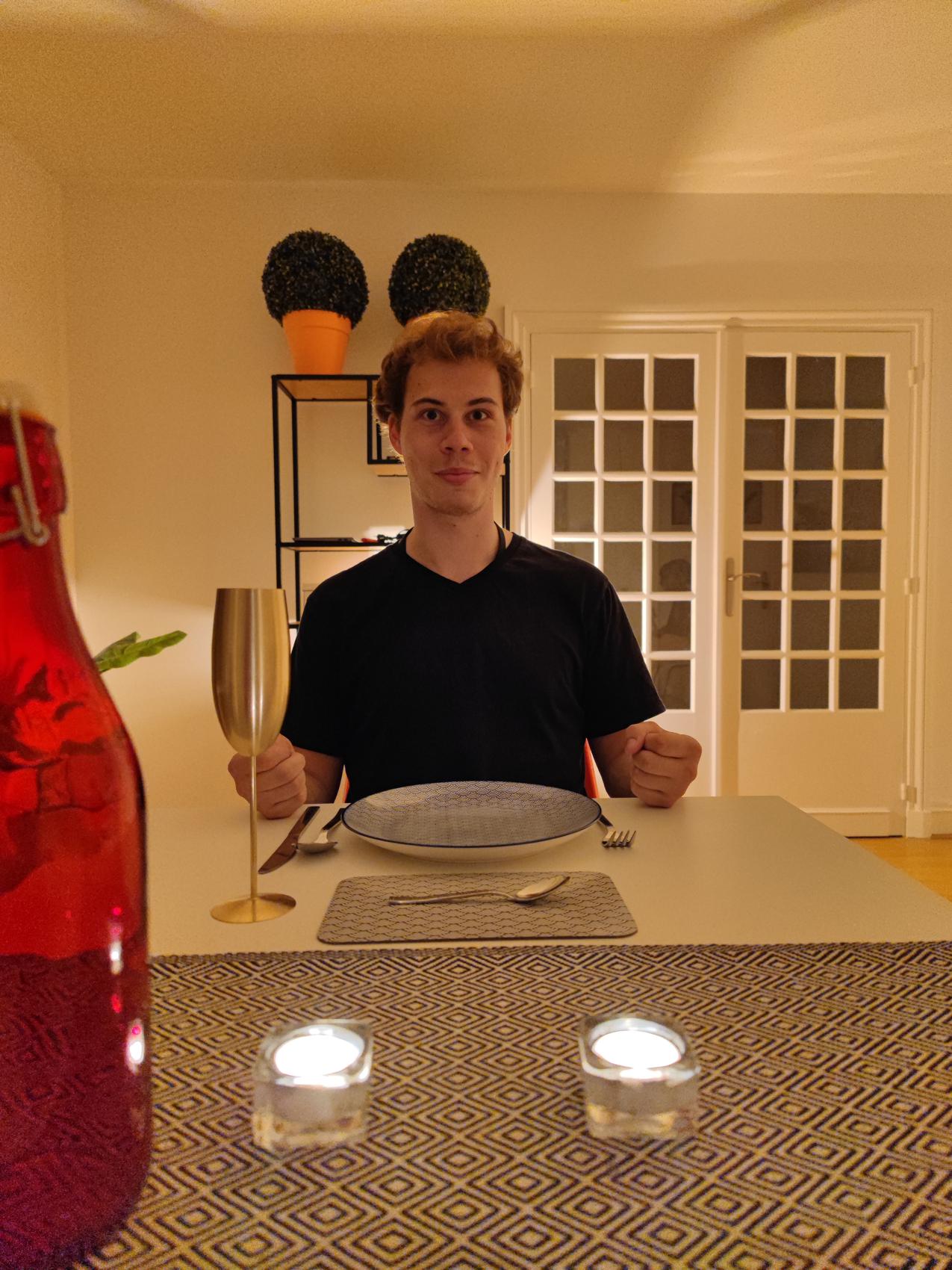} 
    \\
    
    \hline[0.5pt,black]
    
    -3.97 & -0.61 & -0.09 & 1.96
    \\
     \hline[0.5pt,black]
\end{tblr}
\caption{Examples of different levels of \textbf{global color quality} and their corresponding \textit{JOD} scores.}
\end{figure}

\subsection*{Overall image quality}
\SetTblrInner{rowsep=0pt, colsep=2pt}
\begin{figure}[H]
\centering
\begin{tblr}{ccccc}
    \hline[0.5pt,black]
     Level 0 & Level 1 & Level 2 & Level 3 & Level 4 \\
    \hline[0.5pt,black] \\
    \includegraphics[width = 0.18 \textwidth]{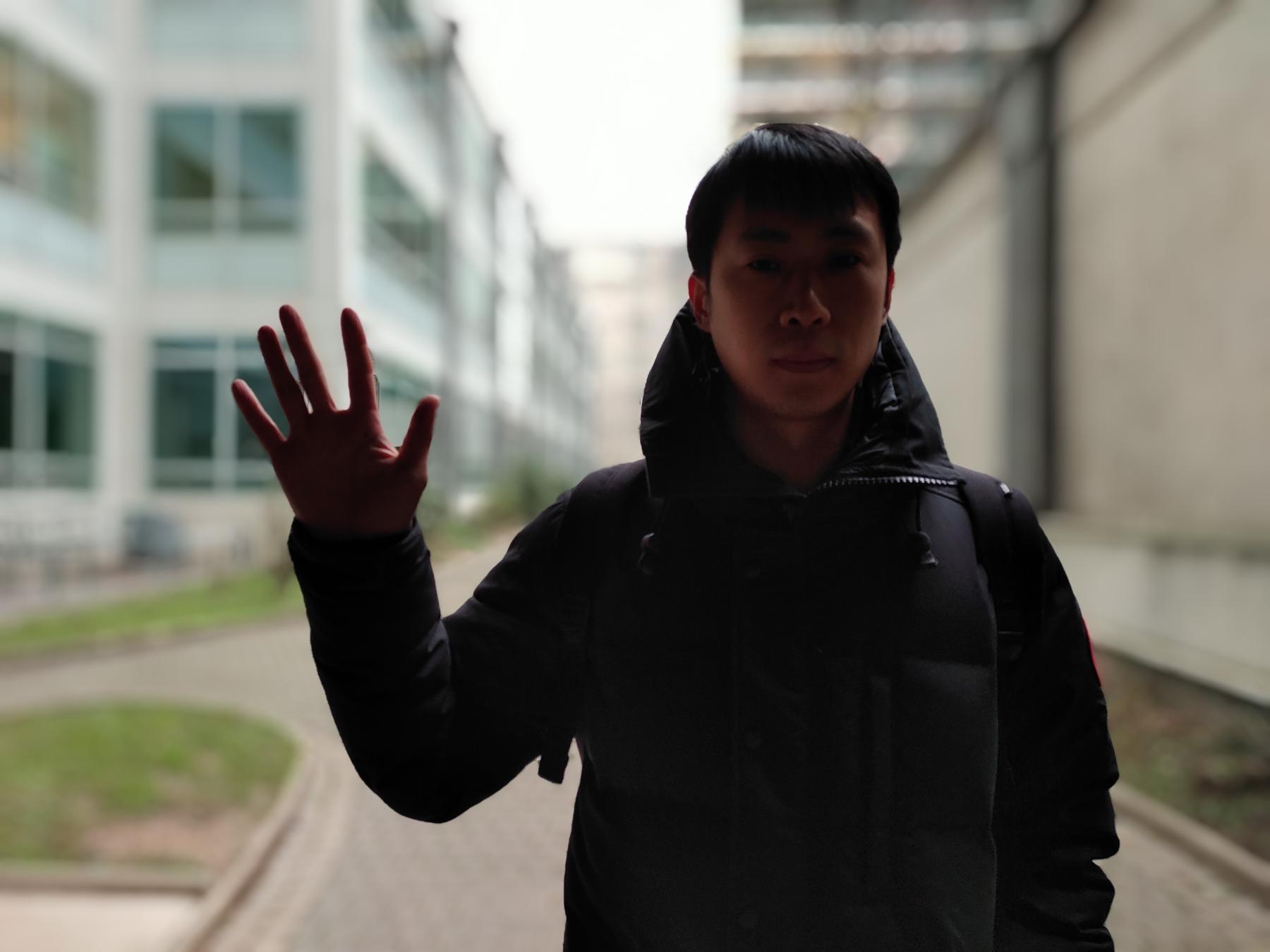} & 
    \includegraphics[width = 0.18 \textwidth]{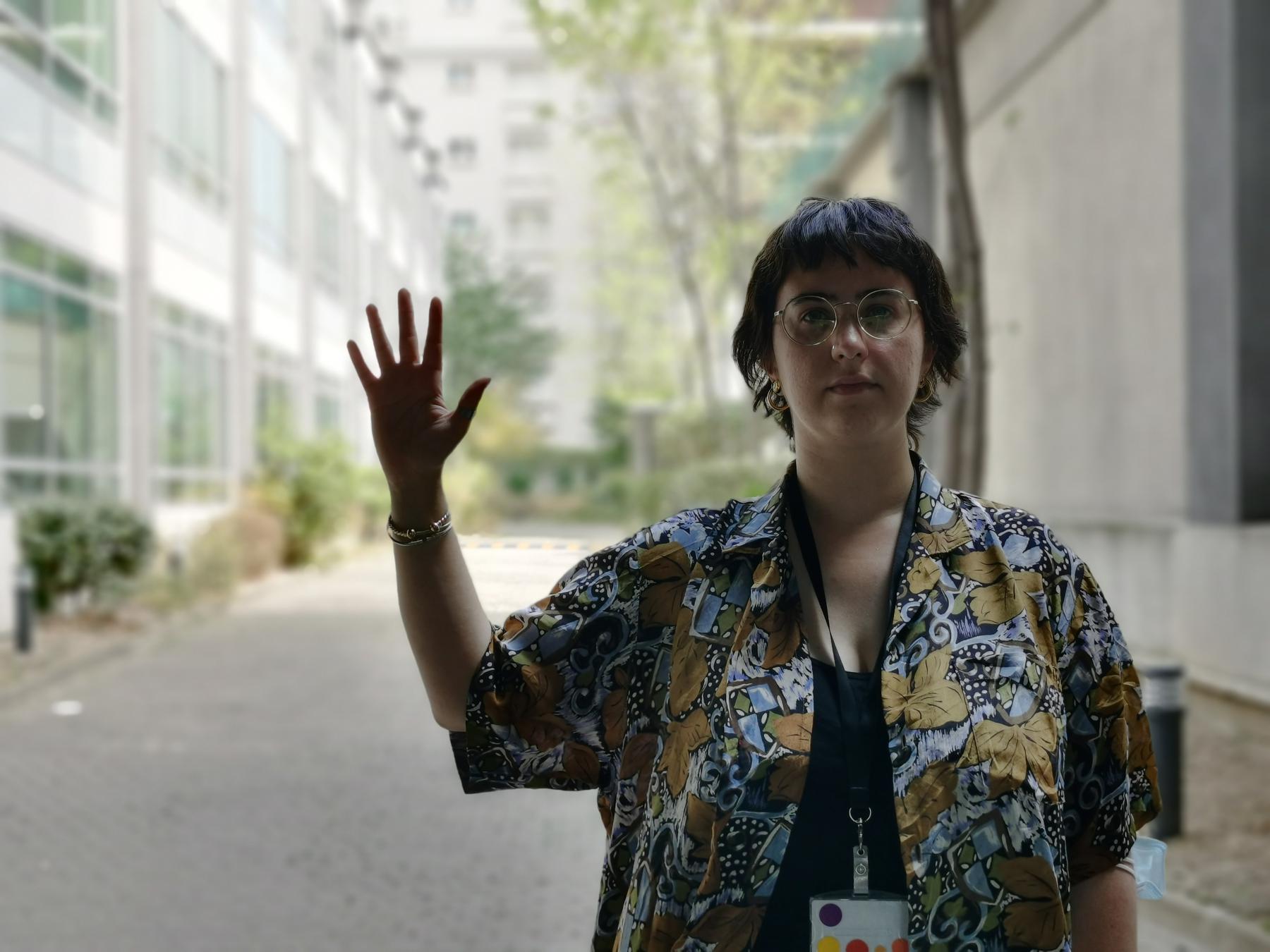} & 
    \includegraphics[width = 0.18 \textwidth]{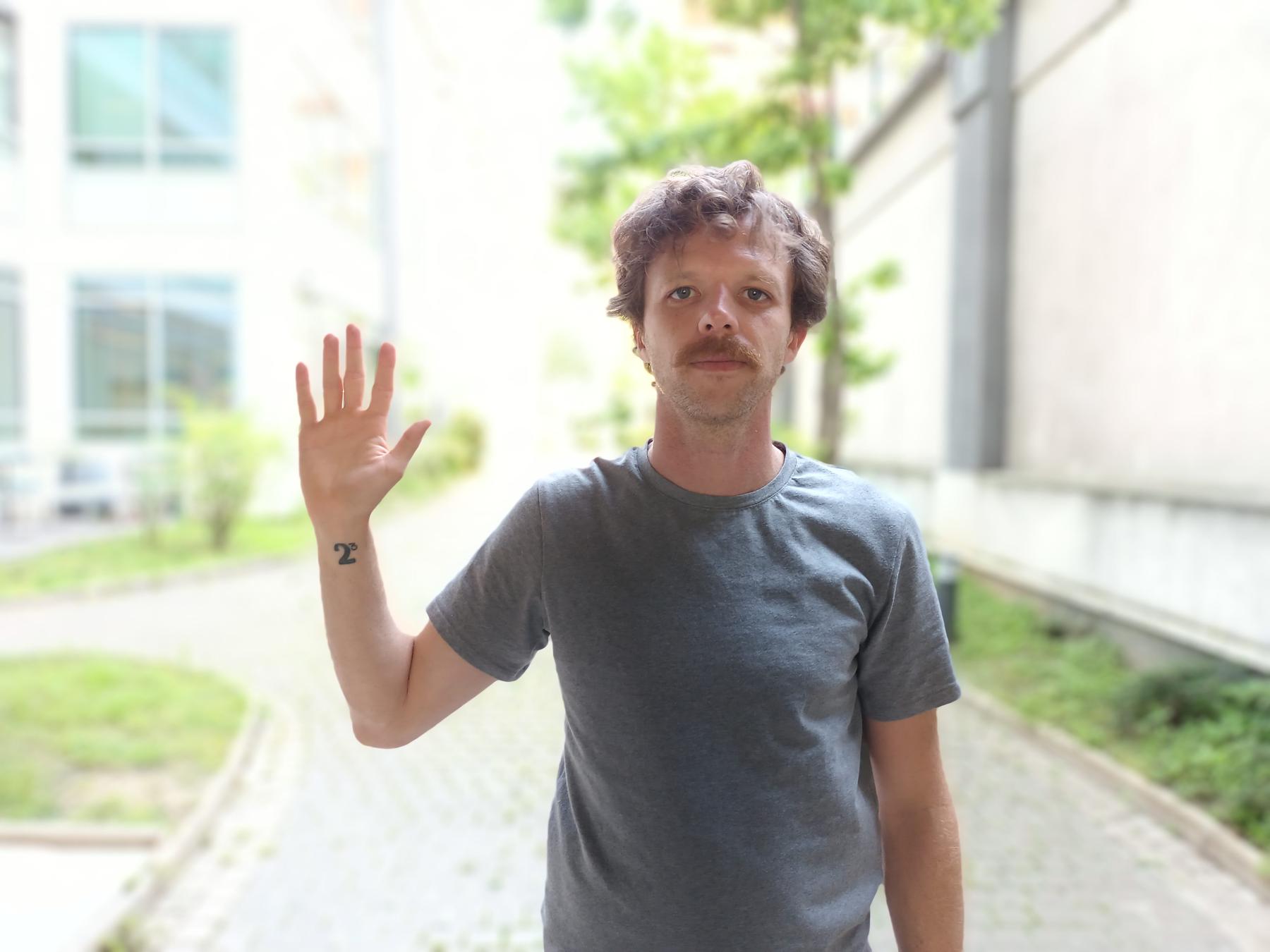} & 
    \includegraphics[width = 0.18 \textwidth]{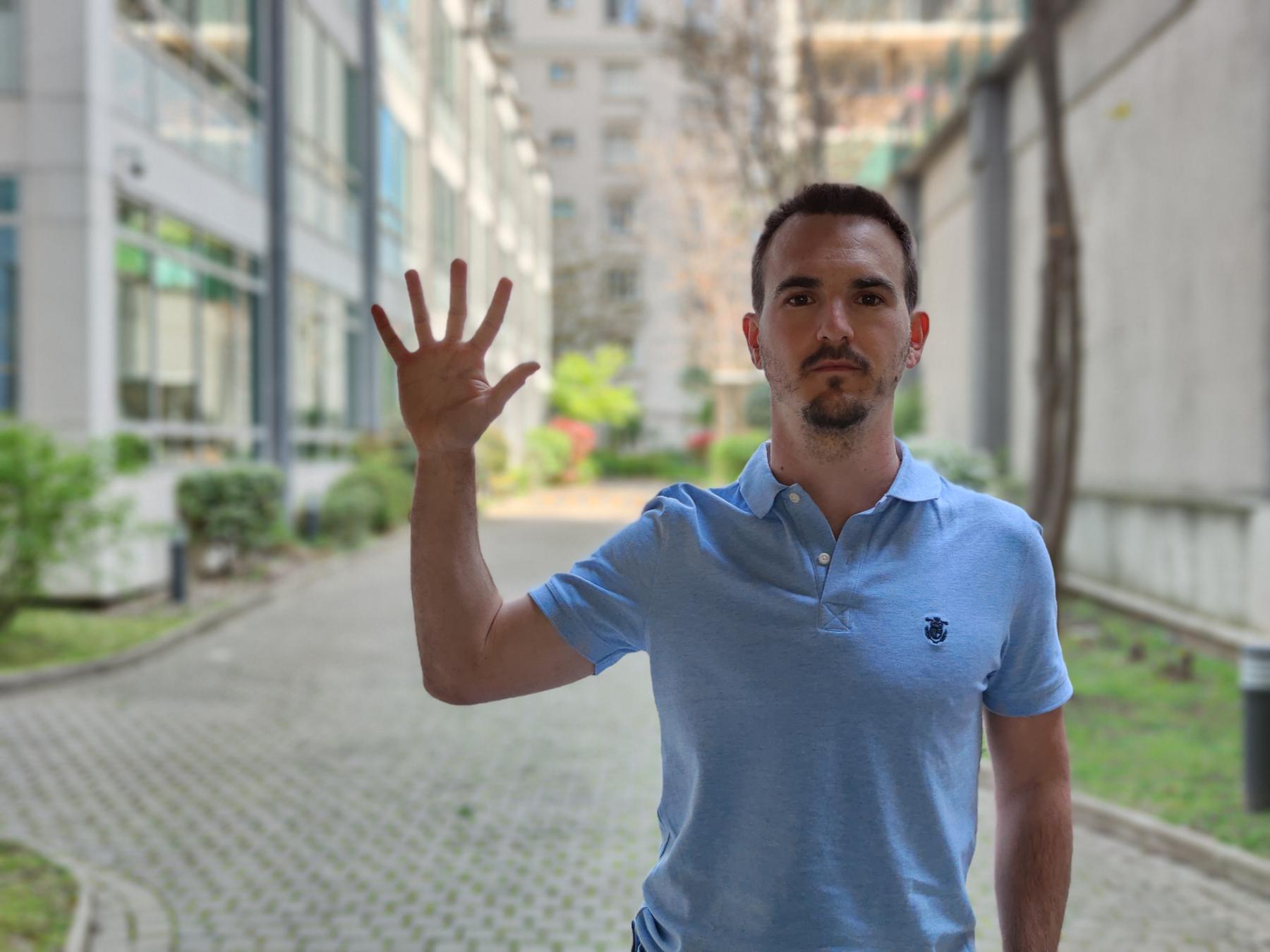} & 
    \includegraphics[width = 0.18 \textwidth]{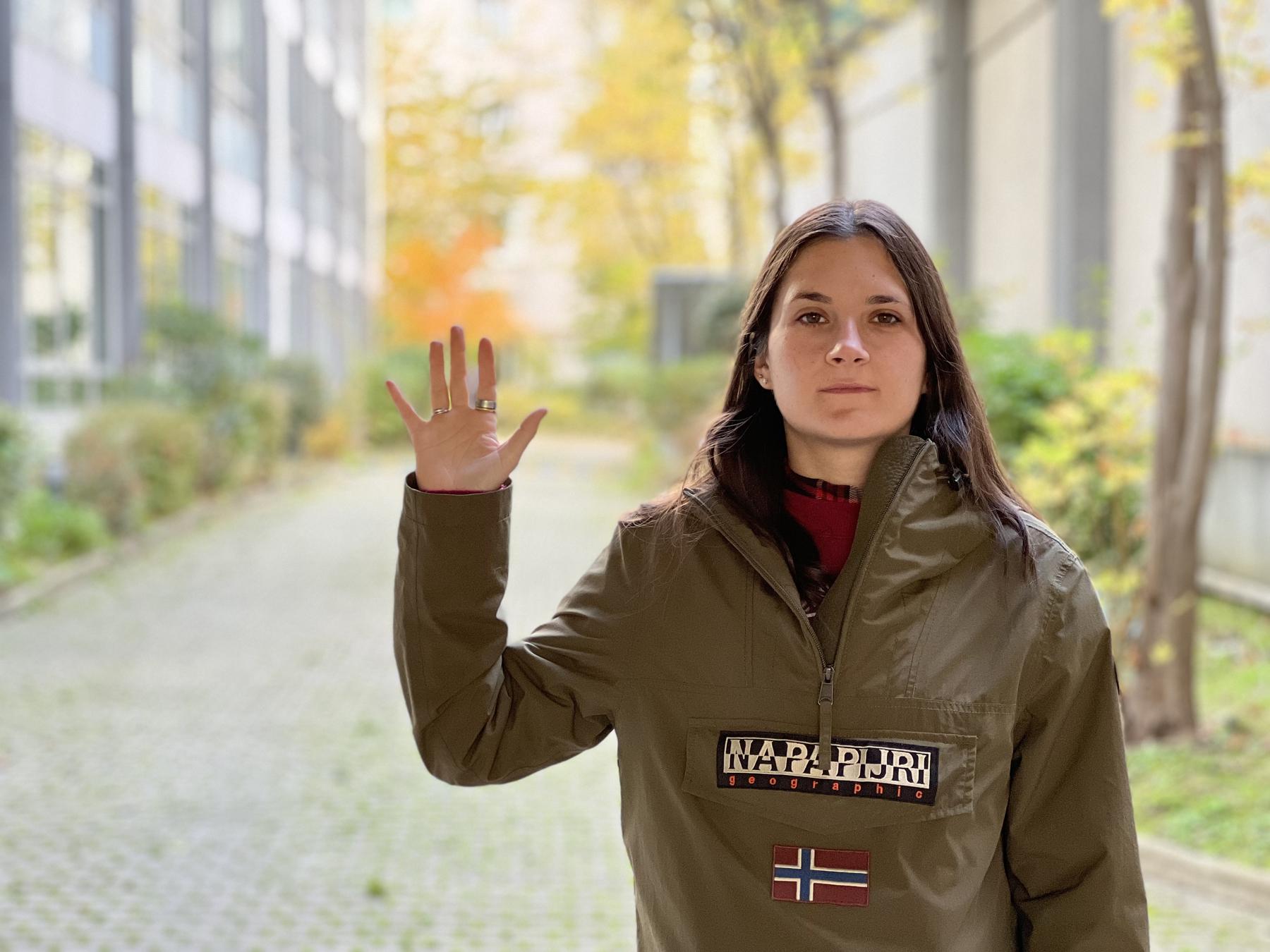}
    \\
    
    \hline[0.5pt,black]
    
    -4.7 & -1.78 & -0.36 & 0.7 & 2.36
    \\
     \hline[0.5pt,black]
\end{tblr}
\caption{Examples of different levels of \textbf{overall image quality} and their corresponding \textit{JOD} scores.}
\end{figure}

 \section{Domain shift}
 \subsection*{Face detail preservation}
 
 \SetTblrInner{rowsep=0pt, colsep=2pt}
\begin{figure}[H]
\centering
\begin{tblr}{cc@{\hskip 20pt}cc}

    \includegraphics[width = 0.2 \textwidth]{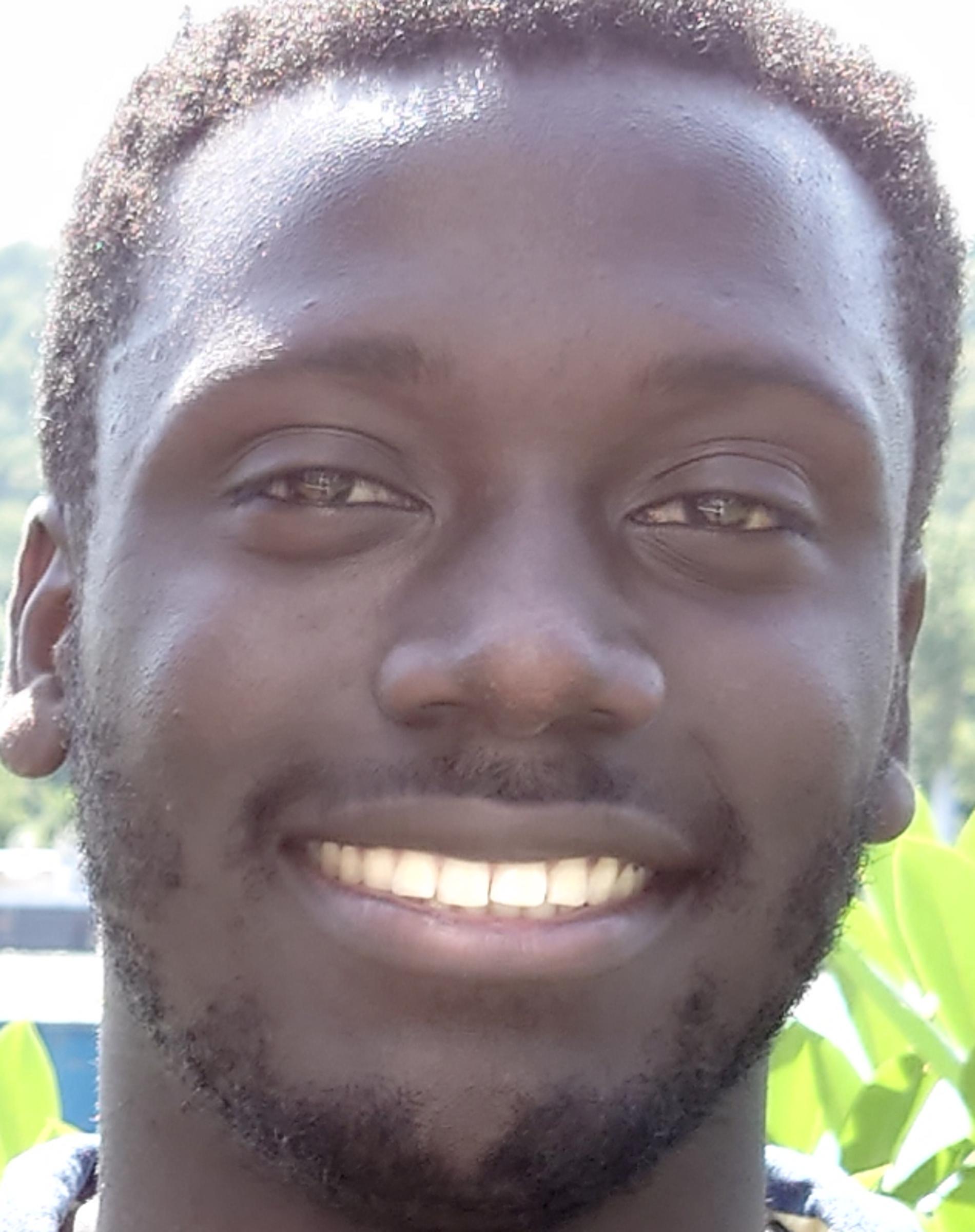} & 
    \includegraphics[width = 0.2 \textwidth]{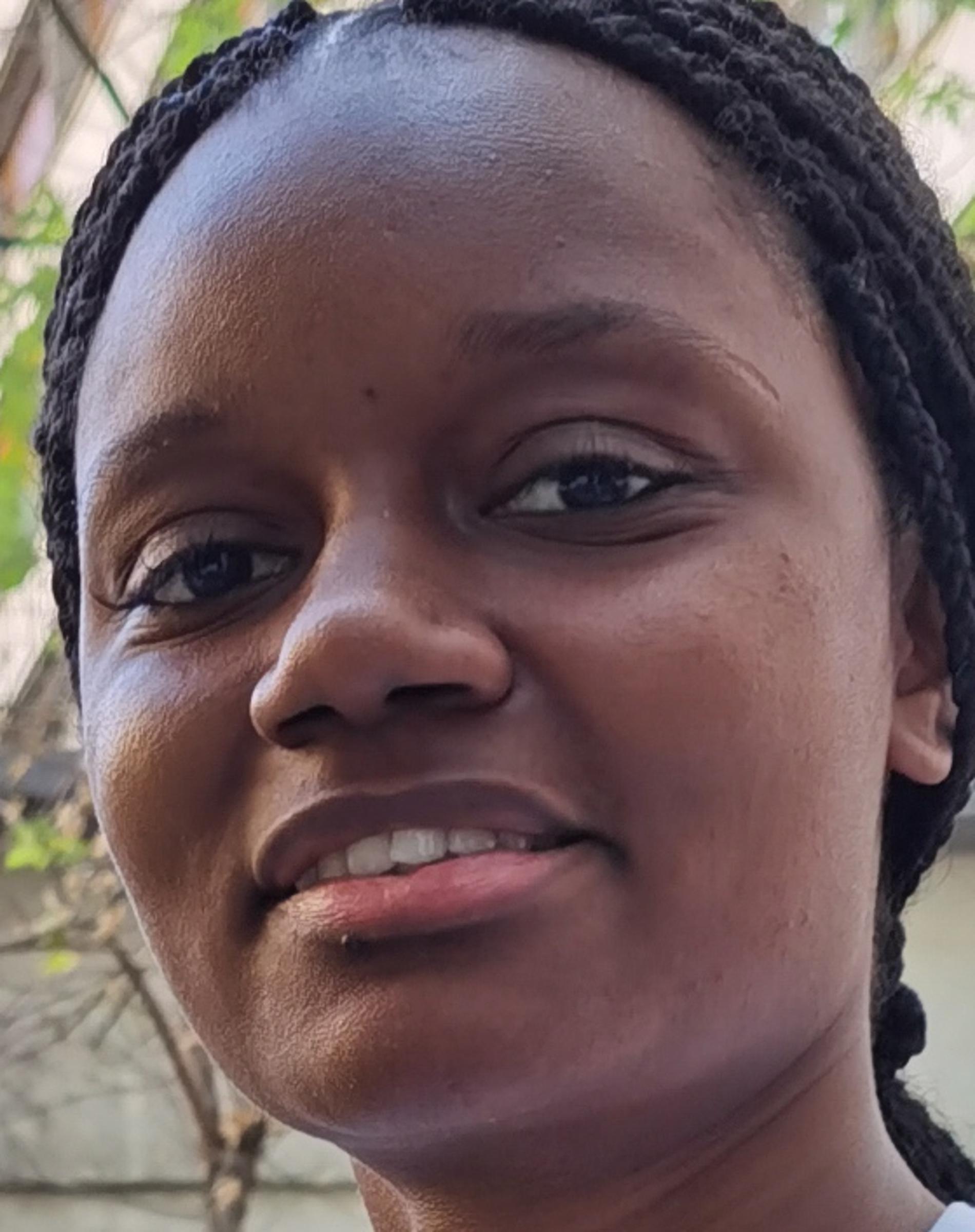} & 
    \includegraphics[width = 0.2 \textwidth]{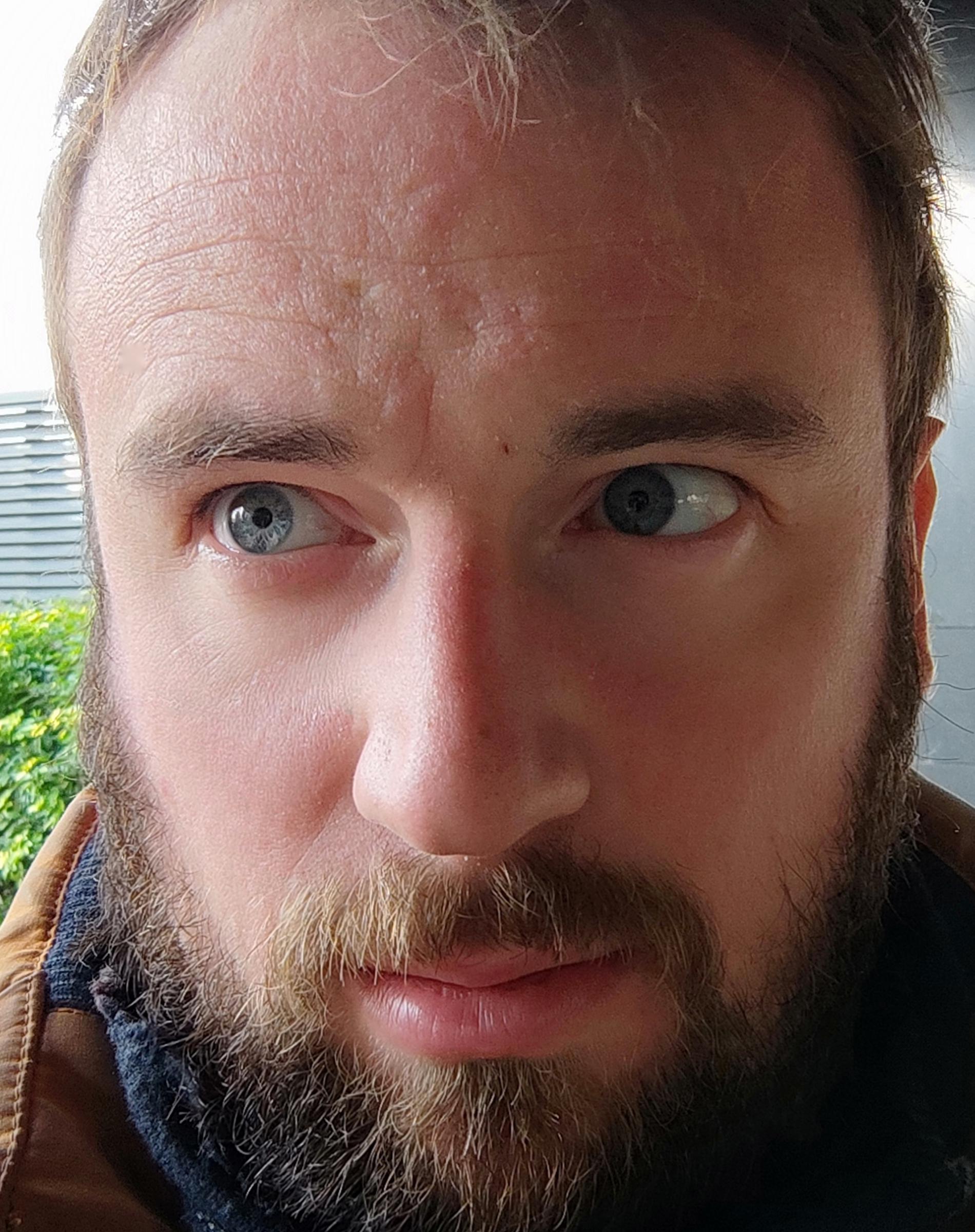} & 
    \includegraphics[width = 0.2 \textwidth]{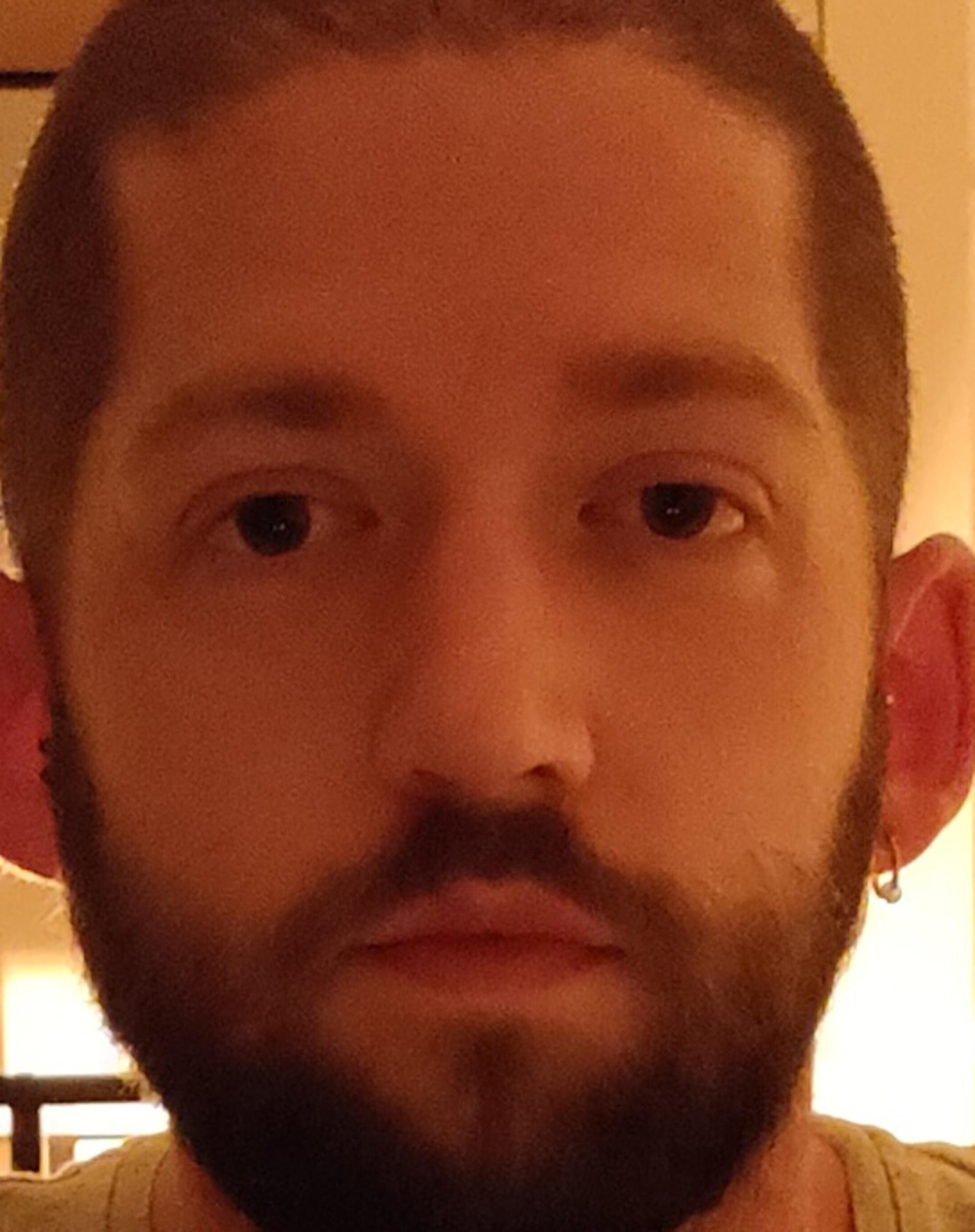} 
    \\

     \hline[0.5pt,black]
       \SetCell[c=2]{c}  \footnotesize{(a) Outdoor scenes.} 
    & & 
     \SetCell[c=2]{c} \footnotesize{(b) \textbf{Left:} outdoor scene; \textbf{right:} lowlight scene.}
     & 
\end{tblr}
\caption{Examples of domain shift in \textbf{face detail preservation} annotations between different scenes of \textbf{(a)} same lighting condition and \textbf{(b)} different lighting conditions. All images have a \textit{JOD} value close to 0. }
\end{figure}
 
\subsection*{Face target exposure}

\SetTblrInner{rowsep=0pt, colsep=2pt}
\begin{figure}[H]
\centering
\begin{tblr}{ccc}
    \includegraphics[width = 0.2 \textwidth]{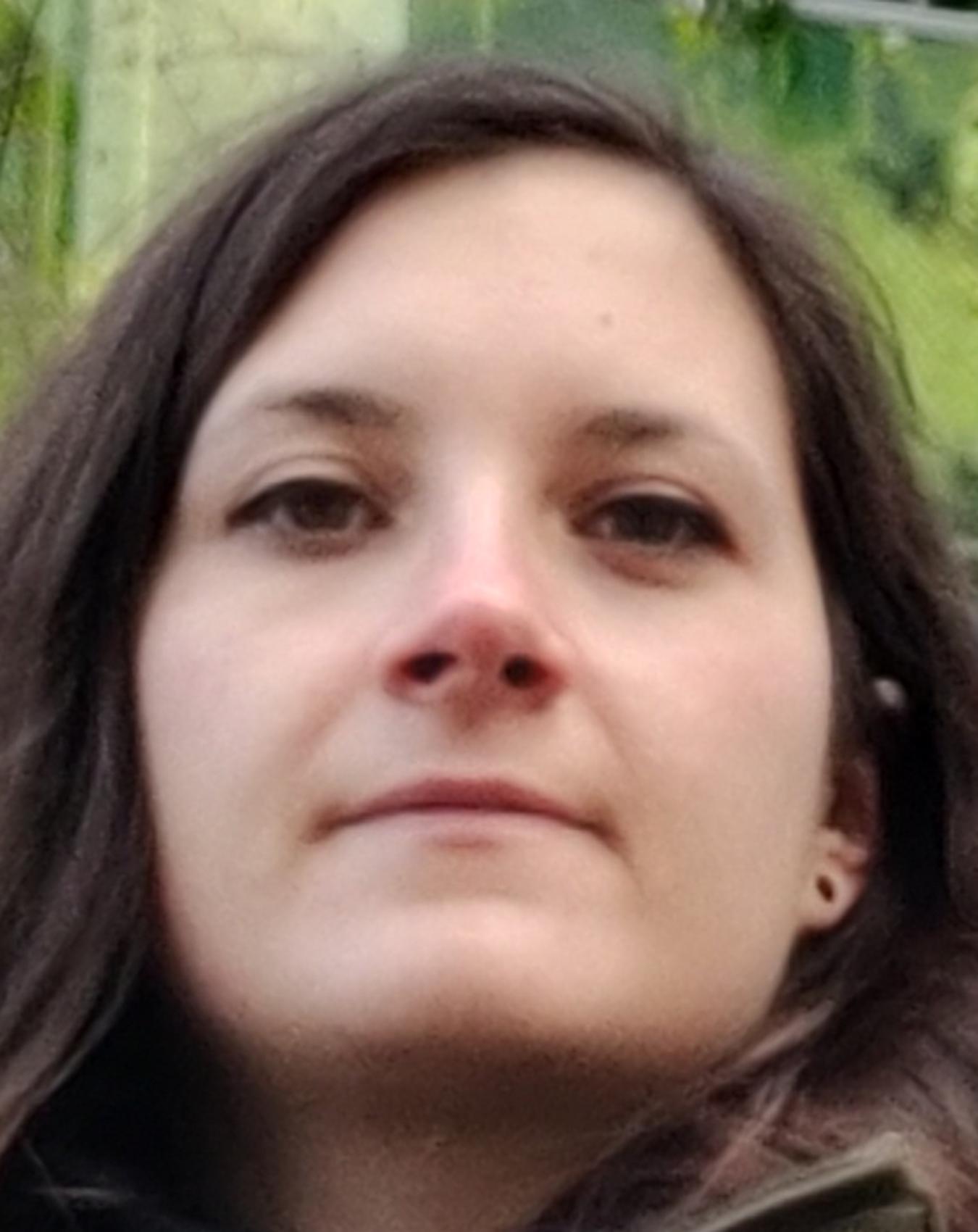} &
    \includegraphics[width = 0.2 \textwidth]{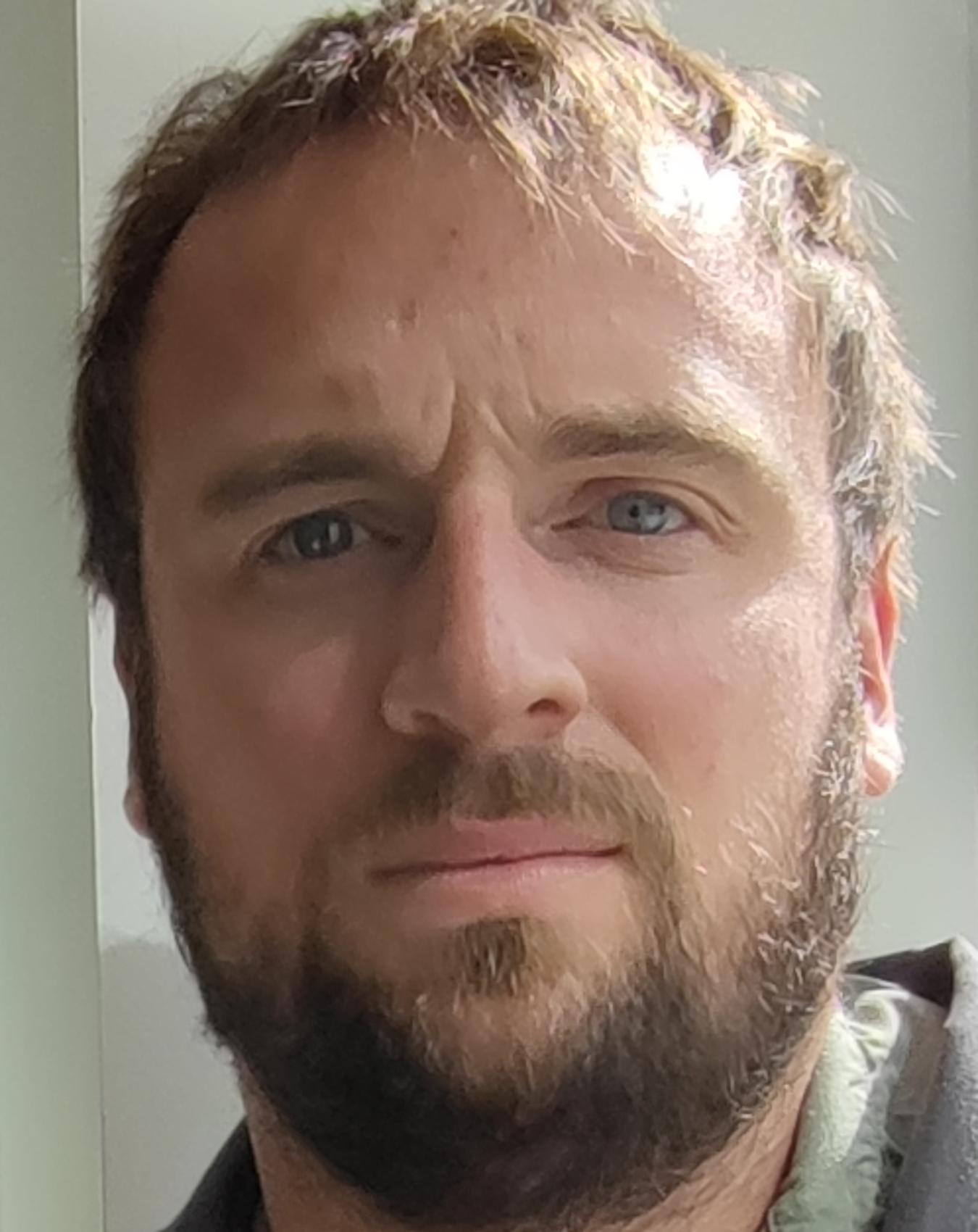} &
    \includegraphics[width = 0.2 \textwidth]{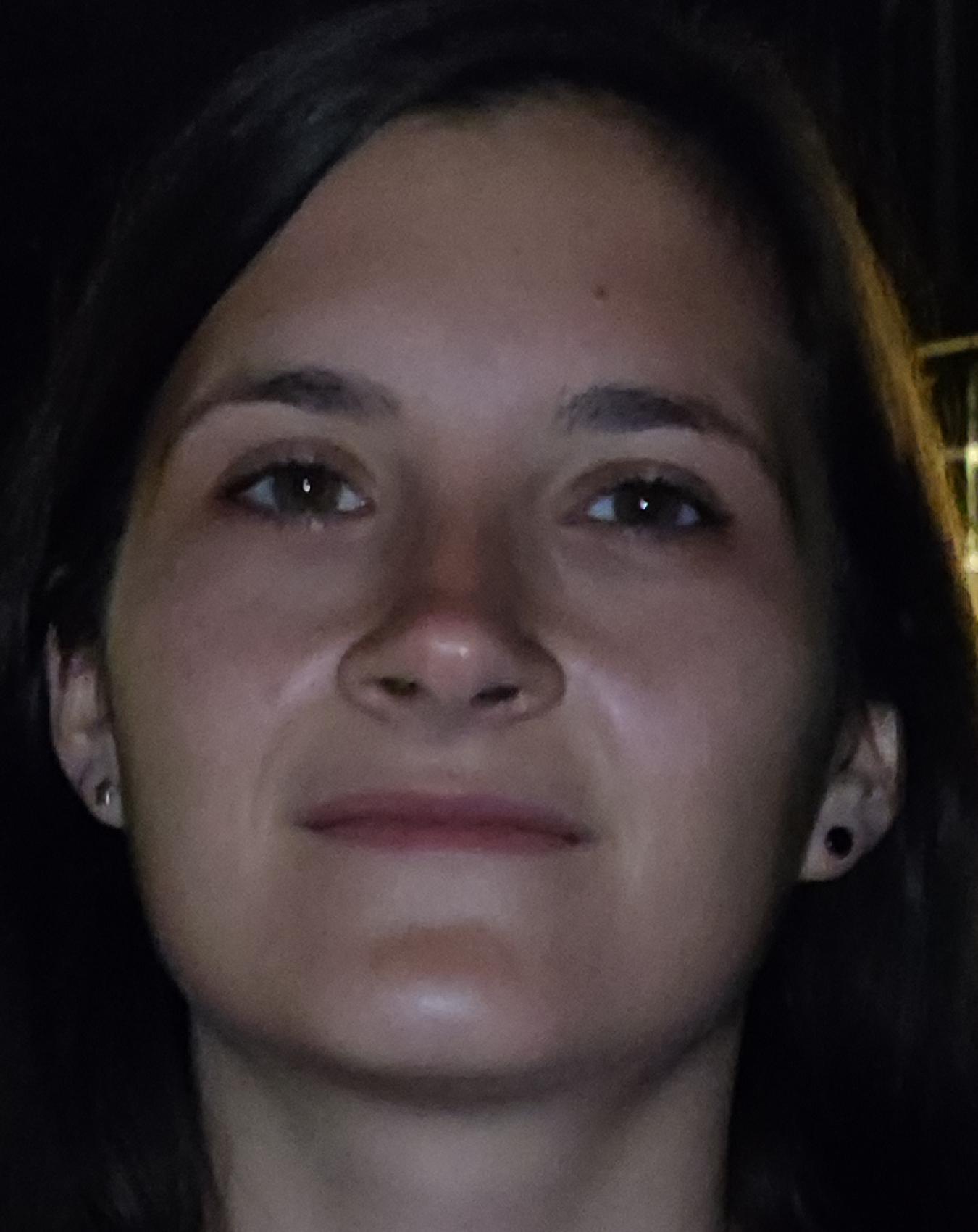}
    \\
     \hline[0.5pt,black]
       \SetCell[c=1]{c}  \footnotesize{(a) Outdoor scene.} 
    &
     \SetCell[c=1]{c} \footnotesize{(b) Indoor  scene.}
     &
     \SetCell[c=1]{c} \footnotesize{(c) Lowlight  scene.}
     &
\end{tblr}
\label{supp:domainShiftExpo}
\caption{Examples of domain shift in \textbf{face target exposure} annotations between different lighting conditions. All images have a \textit{JOD} value close to 0.}
\end{figure}

\subsection*{Global color quality}
\SetTblrInner{rowsep=0pt, colsep=2pt}
\begin{figure}[H]
\centering
\begin{tblr}{ccc}

    \includegraphics[width = 0.3 \textwidth]{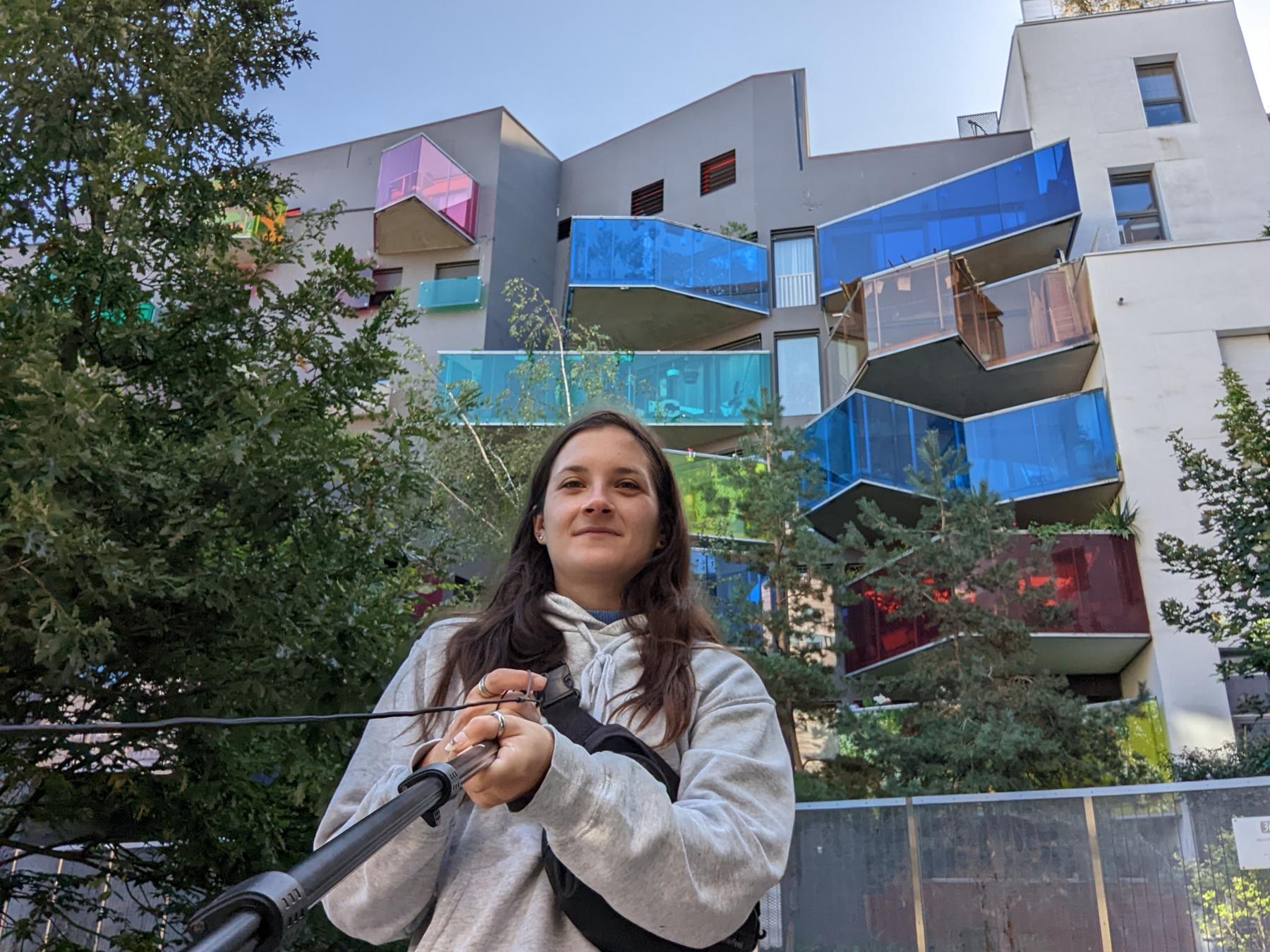} 
    & 
    \includegraphics[width = 0.3 \textwidth]{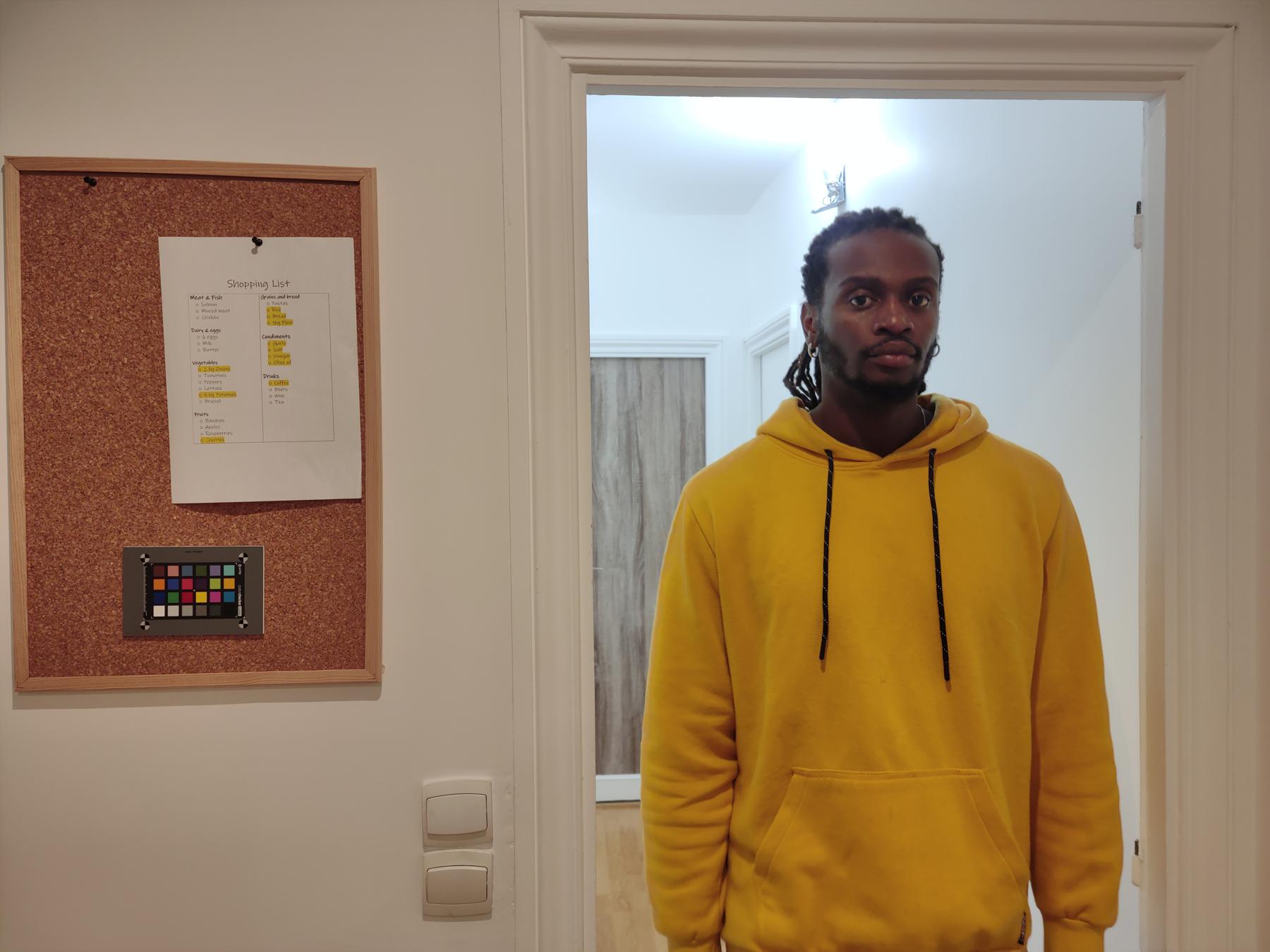}
    & 
    \includegraphics[width = 0.3 \textwidth]{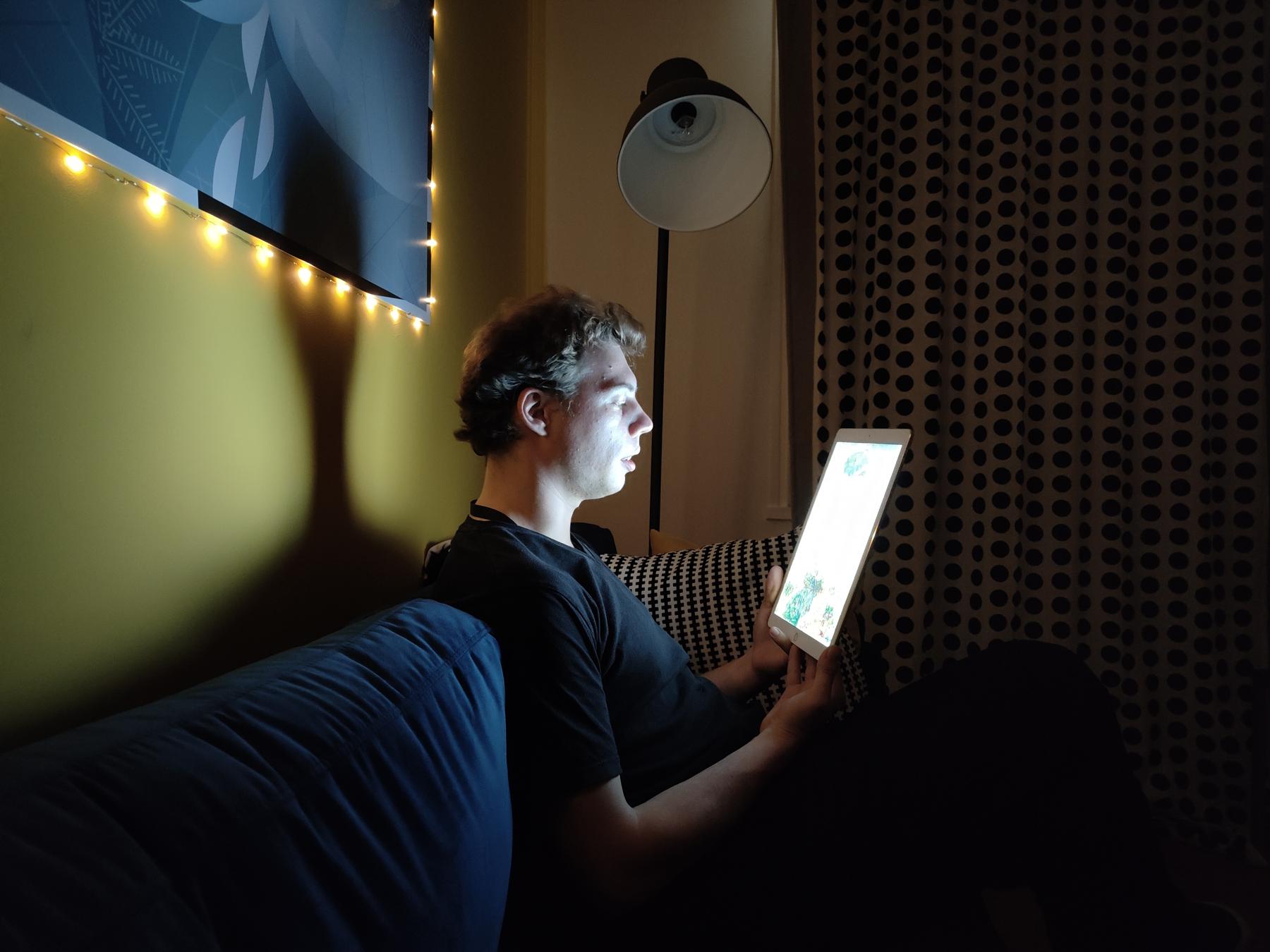} 
    \\

     \hline[0.5pt,black]
       \SetCell[c=1]{c}  \footnotesize{(a) Outdoor scene.} 
    &
     \SetCell[c=1]{c} \footnotesize{(b) Indoor scene.}
     &
      \SetCell[c=1]{c} \footnotesize{(c) Lowlight scene.}
     &
\end{tblr}
\caption{Examples of domain shift in \textbf{global color quality} annotations between different lighting conditions. All images have a \textit{JOD} value close to 0. }
\label{supp:DomainShiftColor}
\end{figure}

\subsection*{Overall image  quality}
\SetTblrInner{rowsep=0pt, colsep=2pt}
\begin{figure}[H]
\centering
\begin{tblr}{ccc}

    \includegraphics[width = 0.3 \textwidth]{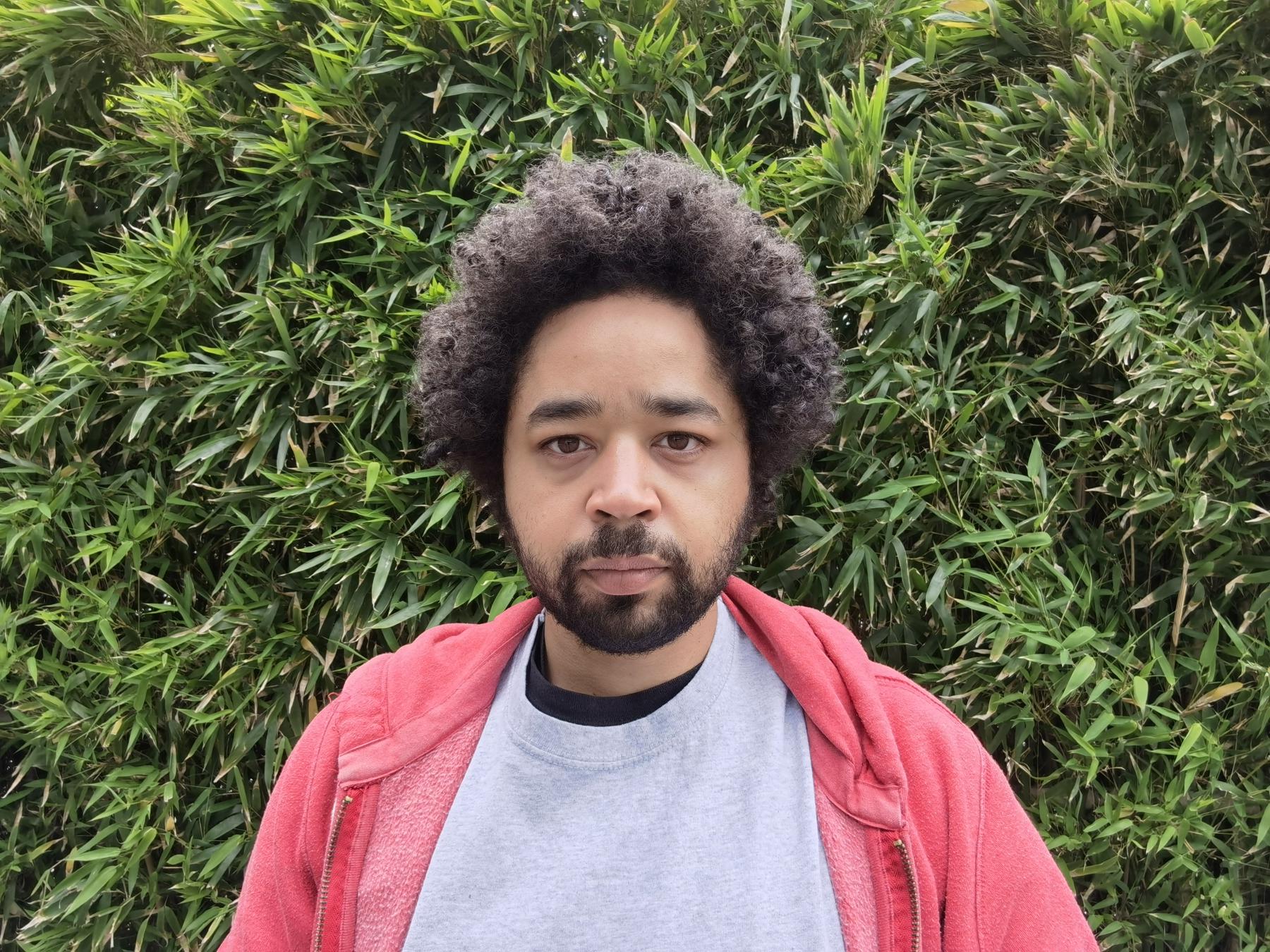} & 
    \includegraphics[width = 0.3 \textwidth]{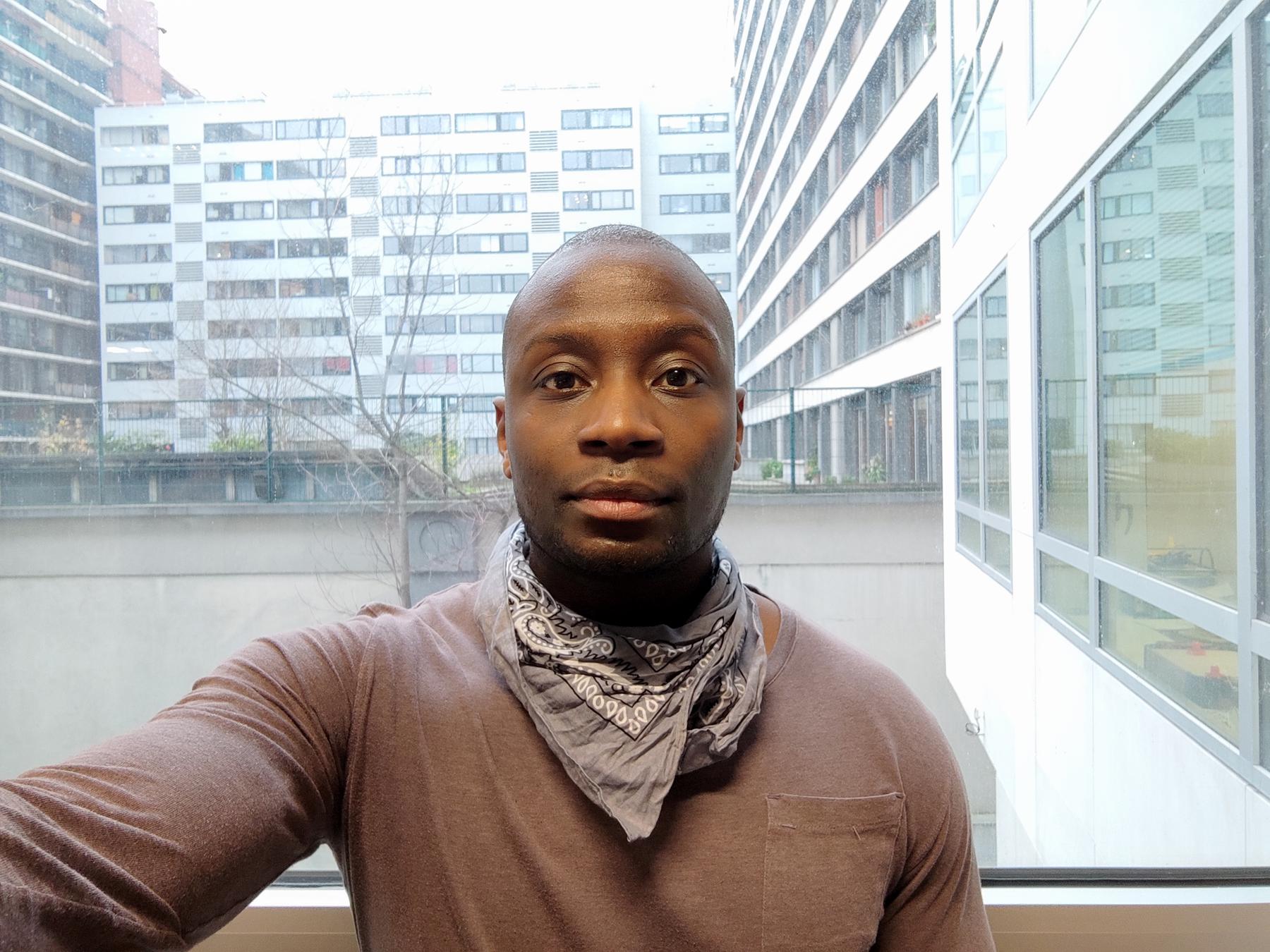} 
    & 
    \includegraphics[width = 0.17 \textwidth]{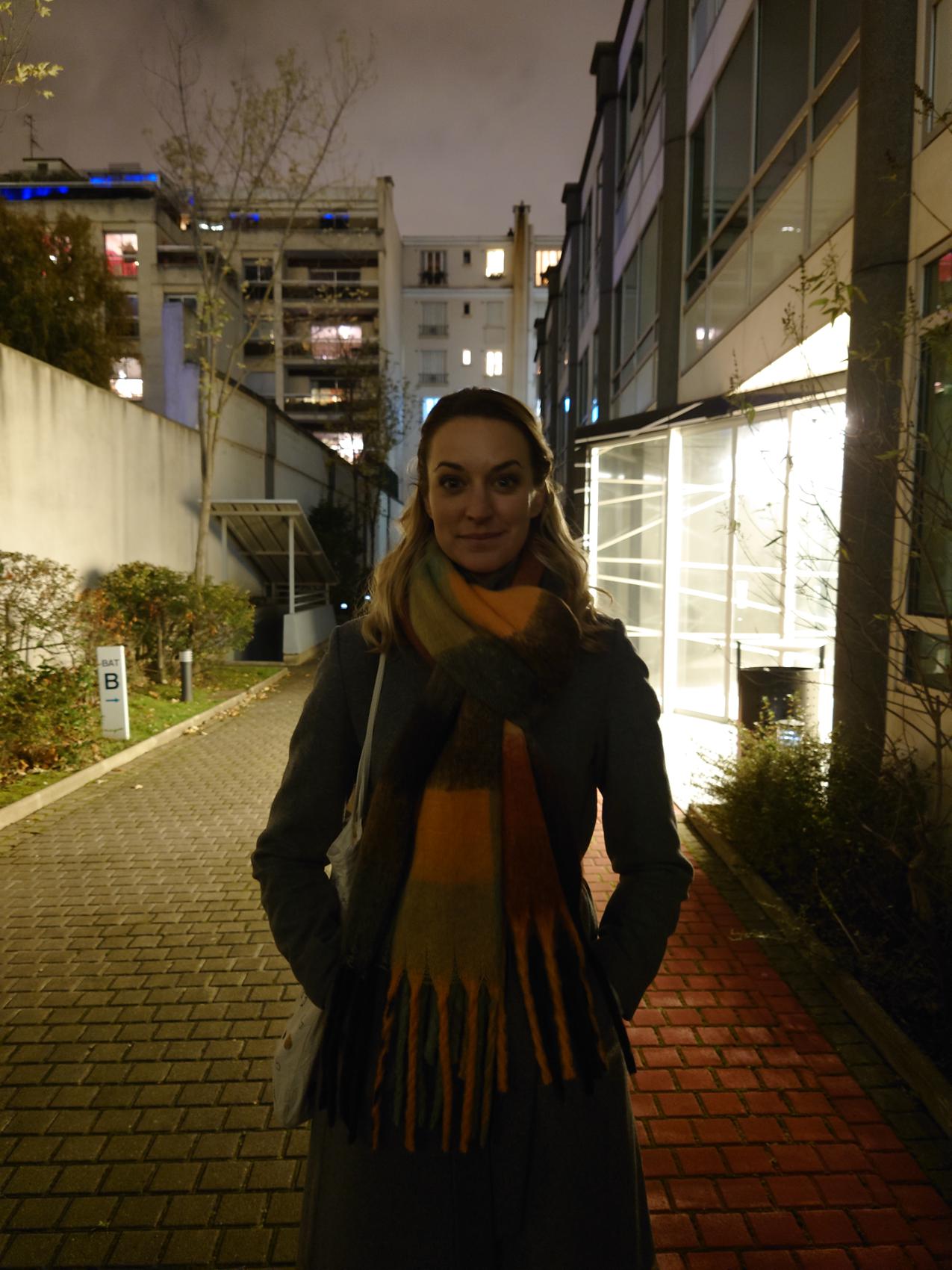}
    \\

     \hline[0.5pt,black]
       \SetCell[c=1]{c}  \footnotesize{(a) Outdoor  scene.} 
    &
     \SetCell[c=1]{c} \footnotesize{(b) Indoor  scene.}
     &
      \SetCell[c=1]{c} \footnotesize{(c) Lowlight  scene.}
     &
\end{tblr}
\caption{Examples of domain shift in \textbf{overall image quality} annotations between different lighting conditions. All images have a \textit{JOD} value close to 0. }
\end{figure}

\section{PIQ23 Characteristics}
\subsection*{Smartphone devices}
\begin{table}[]
\centering
\begin{tabular}{|l|l|}
\hline
\rowcolor[HTML]{DAE8FC} 
\textbf{Brand} & \textbf{Nb devices} \\ \hline
Samsung & 20 \\ \hline
Xiaomi & 14 \\ \hline
Oppo & 14 \\ \hline
Apple & 13 \\ \hline
Vivo & 9 \\ \hline
Huawei & 7 \\ \hline
OnePlus & 6 \\ \hline
Sony & 5 \\ \hline
Google & 4 \\ \hline
Asus & 3 \\ \hline
Realme & 2 \\ \hline
Motorola & 2 \\ \hline
Other & 3 \\ \hline
\end{tabular}
\caption{Smartphone brand distribution in PIQ23}
\label{tab:brands}
\end{table}
PIQ23 was collected with over 100 smartphone devices and 14 brands from the last decade. Additionally, multiple camera modes were included across the scenes. \Cref{tab:brands} shows the distribution of the smartphone brands of PIQ23.

\subsection*{Ethnicities and genders}
We have constructed PIQ23 with extra attention to gender and ethnic biases. We have tried to the best of our capabilities to minimize those biases through bias analysis. It should be noted that PIQ23 is the result of multiple years of engineering and photographic efforts and is not necessarily uniformly distributed through all characteristics. 
We have separated the skin tones, following the Fitzpatrick skin type (FST) ruler \cite{fitzpatrick1988validity}, into four categories: Fair, Asian, Medium, and Deep. We recognize the difficulties encountered by image enhancement algorithms on deep skin tones, hence we have also designed some of the PIQ23 scenes to include uniquely deep skin tones. \Cref{tab:ehtnics} shows a rough estimation of the skin tone distribution in PIQ23. A more detailed analysis could be provided later throughout the life cycle of PIQ23.

\begin{table}[H]
\centering
\begin{tabular}{|l|l|}
\hline
\rowcolor[HTML]{DAE8FC} 
\textbf{Skin tone} & \textbf{Estimated \%} \\ \hline
Fair & 50\% \\ \hline
Deep & 30\%  \\ \hline
Asian & 15\%  \\ \hline
Medium & 5\%  \\ \hline
\end{tabular}
\caption{Skin tone estimated distribution in PIQ23}
\label{tab:ehtnics}
\end{table}

\section{Annotation tool}

\begin{figure}[H]
    \centering
    \includegraphics[width = 0.9 \textwidth]{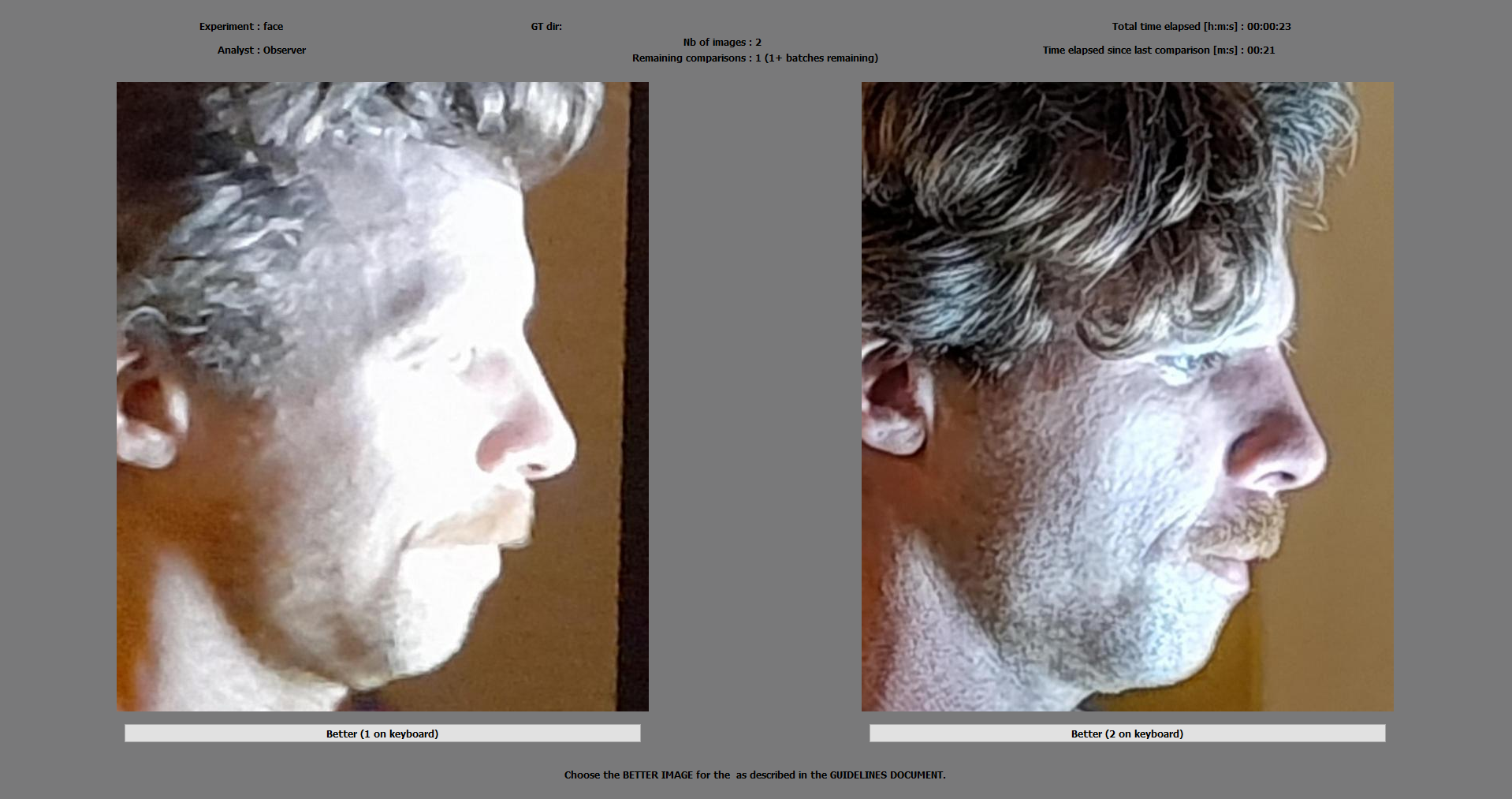}
    \caption{Example of the annotation tool used to acquire pairwise comparisons. }
    \label{fig:my_label}
\end{figure} 

{\small
\bibliographystyle{ieee_fullname}
\bibliography{egbib}
}